\documentclass[12pt]{article}

\usepackage[margin=1in]{geometry}

\usepackage[authoryear]{natbib}
\usepackage{graphicx,caption,subcaption}
\graphicspath{{./figures/}{./figures/fig2/}{../../figures/}{../../figures/fig2/}}

\usepackage{color,hyperref}
\usepackage[dvipsnames]{xcolor}
\hypersetup{colorlinks=true,urlcolor=blue,citecolor=blue,linkcolor=blue}
\usepackage{enumitem}
\usepackage{booktabs}

\usepackage{amsmath,amsthm,amssymb,amsfonts}

\newtheorem{definition}{Definition}
\newtheorem{lemma}{Lemma}
\newtheorem{theorem}{Theorem}

\newtheorem{remark}{Remark}

\usepackage{authblk}

\usepackage[linesnumbered,ruled,vlined]{algorithm2e}
\SetKwInput{KwInput}{Input}     
\SetKwInput{KwOutput}{Output}   

\newcommand{\E}{\mathbb{E}}
\newcommand{\R}{\mathbb{R}}

\newcommand{\Acal}{\mathcal{A}}

\newcommand{\Fcal}{\mathcal{F}}

\newcommand{\Hcal}{\mathcal{H}}

\newcommand{\Pcal}{\mathcal{P}}

\newcommand{\Rcal}{\mathcal{R}}

\newcommand{\bb}[1]{\boldsymbol{#1}}
\newcommand{\bbhat}[1]{\widehat{\boldsymbol{#1}}}

\newcommand{\ind}[1]{\boldsymbol{1}_{#1} }

\newcommand{\trace}[1]{\text{\normalfont Trace}\left( #1 \right)}

\newcommand{\tr}{^{\intercal}}

\newcommand{\var}{\mathrm{Var}}
\newcommand{\prob}{\mathbb{P}}

\DeclareMathOperator*{\argmin}{arg\,min}
\DeclareMathOperator*{\argmax}{arg\,max}

\allowdisplaybreaks

\title{Policy Learning via Active Preference Collection with Applications to AI Prompt Routing}

\date{}
\author[1]{Subhrajyoty Roy}
\author[2]{Zhimei Ren}
\author[1]{Mengxin Yu}

\affil[1]{Washington University in St. Louis}
\affil[2]{University of Pennsylvania}

\begin{document}
\maketitle


\begin{abstract}
Identifying the best AI model for each task from limited human pairwise preference feedback is a key challenge in model evaluation and routing. In this paper, we propose an active learning framework for contextual best-arm identification using pairwise preference feedback. We first design an adaptive pairwise comparison strategy that optimizes data acquisition. Specifically, instead of uniformly sampling candidate pairs, we iteratively sample comparisons according to their relative uncertainties. Building upon the adaptively collected preference data, we then develop a robust decision policy that maintains reliable performance under potential model misspecification. Theoretically, we establish minimax-optimal regret guarantees while requiring only semiparametric convergence rates for the underlying reward estimators. We further validate our theoretical results through extensive simulation studies and experiments on three real-world LLM evaluation benchmark datasets. Our findings demonstrate that the proposed approach identifies the best-performing models more efficiently compared with standard uniform and other benchmark strategies.

\end{abstract}

\section{Introduction}\label{sec:intro}

Large language models (LLMs) have become increasingly prominent across a wide range of applications,
including text generation and comprehension, reasoning, coding, summarization, and information retrieval. 
At the same time, the rapidly expanding landscape of language models has made model selection 
increasingly complex.
No single model consistently outperforms others; instead, models exhibit substantial 
heterogeneity in their relative performance across tasks~\citep{wang2024comprehensive}. 
This challenge is further compounded by the growing 
number of language models (e.g., more than $1.5$ million language models are hosted in the Huggingface\footnote{\url{https://huggingface.co/models}} library~\citep{castano2024analyzing}). Therefore, selecting the most appropriate model for a given user prompt has become an increasingly challenging and pertinent decision-making problem.

The prompt routing framework~\citep{ong2024routellm} addresses this challenge by dynamically routing each input prompt to the model expected to perform best under a chosen evaluation criterion. A key component of this framework is therefore reliable evaluation of model performance across prompts. While many automated evaluation metrics exist~\citep{chang2024survey}, pairwise human (or AI-as-a-judge) preference remains the standard approach to assess how well model responses align with user preferences, especially on open-ended tasks~\citep{chiang2024chatbot,weidinger2025toward}. This is because, pairwise preferences provide a practical source of feedback when absolute numerical evaluations are difficult or costly to define, e.g., subjective evaluation of LLM outputs~\citep{bengs2021preference}. In pairwise-preference evaluation, the experimenter provides the same prompt to two candidate models, and an evaluator then compares the generated responses and indicates which one is preferred. Aggregating such comparisons across prompts provides information about the relative performance of the competing models and can subsequently be used to construct a routing policy. For instance, platforms such as Chatbot Arena~\citep{chiang2024chatbot} collect user preferences by presenting responses generated by two language models and aggregating these pairwise comparisons to construct leaderboards of various models. Other examples also include evaluations of search-ranking systems from user preferences~\citep{yue2009interactively}, personalized recommendation systems~\citep{singla2016actively}, human-robot interactions and autonomous driving~\citep{biyik2018batch}, in social sciences as conjoint surveys~\citep{gosciak2025adaptive}.

However, collecting pairwise preference labels can be costly. When the number of candidate models is large, it becomes infeasible to compare every pair of models for a wide range of prompts. This naturally raises the question of how to allocate a limited comparison budget most efficiently in order to enable decision makers to identify the best-performing model for each task.

Addressing this question involves two complementary tasks: developing an adaptive experimentation framework for efficiently collecting pairwise comparisons and a policy-learning algorithm for identifying the best model for each task. We provide a principled solution to both. In the experimentation stage, we adopt an active sampling framework that adaptively select comparisons based on the current context, past feedback and uncertainty. To be concrete, the experimenter is given a budget of $T$ rounds of pairwise comparisons. At each round,  the experimenter observes a context (e.g., a prompt) and adaptively samples a pair of arms (e.g., two different language models) based on the current context, past observations, and the associated uncertainty. The selected pair is then evaluated by a human annotator or an LLM judge, who provides a preference outcome between the two arms. In the learning stage, the collected data is used to construct a policy that selects the best arm for a future context.

In addition to developing the adaptive sampling and the policy learning strategy within an active learning framework, we also introduce a bias-correction procedure to account for potential deviations from the Bradley-Terry-Luce (BTL) model. Theoretically, our framework achieves the minimax-optimal expected simple regret, and empirically, the proposed adaptive sampling strategy can identify the best-performing models more efficiently and requires significantly fewer comparison samples than uniform sampling.

A similar experimental design has been considered in~\cite{kato2024adaptive}, but our setting differs in that
feedback consists only of binary outcomes from pairwise comparisons. This setting reflects real-world AI evaluation scenarios, where it is often difficult to define or obtain meaningful numerical rewards for subjective tasks. Our framework is also closely related to best-arm identification (BAI), but differs from the commonly studied online bandit setting with the objective of minimizing cumulative regret. In contrast, we consider a fixed-budget policy learning problem, where the goal is to learn the best policy at the end of the experiment rather than to optimize the rewards accumulated during the experiment.

\subsection{Prior Art}\label{sec:literature}
In recent years, substantial efforts have been devoted to developing statistical methodologies for best-arm identification as well as for rank estimation, inference, and decision-making. 

The BAI problem has appeared in a variety of decision-making contexts,  such as the selection of arms in slot machines~\citep{audibert2010best}, the choice of medication and therapeutic interventions~\citep{villar2015multi}, and preference-based recommended movies and songs~\citep{yan2022bandit, ban2024neural}. Depending on the feedback structure and the availability of contextual information, BAI has been studied under various bandit formulations, including multi-arm bandits~\citep{audibert2010best, kaufmann2016complexity}, contextual bandits~\citep{russo2016simple, kato2021role, kato2024adaptive}, and dueling bandits~\citep{bengs2021preference}.

When the feedback takes the form of pairwise preferences, as in our setting, BAI is also closely related to the extensive literature on ranking from comparison data. A large body of work has studied the estimation and identification of top-ranked items under various comparison models, including the Thurstonian and Bradley--Terry--Luce (BTL) models~\citep{negahban2012iterative, chen2015spectral,chen2019spectral,chen2022partial,chen2022optimal}. More recently, several works have incorporated covariate information into ranking models. For example, \cite{fan2024covariate, fan2024uncertainty} studied covariate-adjusted ranking models by decomposing the latent score in the BTL model into an intrinsic score and a covariate effect. Subsequently, \cite{dong2026statistical,luo2026deep} extended these frameworks to dynamic settings with nonparametric latent score functions. In a related direction, \cite{li2025efficient} studied nonparametric latent score models from a semiparametric perspective, developing efficient estimation procedures for expected score functions along with valid statistical inference procedures. A growing body of work has also established asymptotic normality for ranking estimators and developed uncertainty quantification methods for ranking outcomes~\citep{simons1999asymptotics,han2020asymptotic,liu2023lagrangian,gao2023uncertainty,fan2025ranking,fan2026spectral,han2025unified}. With the emergence of reinforcement learning from human feedback (RLHF) as a prominent paradigm for aligning large language models, preference-based learning has received increasing attention as a means of incorporating human preferences into model training~\citep{christiano2017deep,ouyang2022training,casper2023open}. In parallel, pairwise preference feedback has been increasingly used for evaluating and comparing large language models, where human annotators or LLM judges are asked to select the preferred response from a pair of candidate models~\citep{chiang2024chatbot}.

In parallel, a growing body of work has studied decision-making problems in conjunction with active learning and adaptive sampling; see, e.g., \cite{zoghi2014relative,bengs2021preference,saha2021optimal,di2024variance}. Adaptive experimental designs can considerably improve statistical efficiency under a fixed sampling budget~\citep{lai1985asymptotically,burnetas1997optimal,zhan2024policy,zrnic2024active,li2026robust} and, in some settings, are necessary for achieving minimax-optimal guarantees~\citep{komiyama2022minimax}. Among recent studies, \cite{zrnic2024active,li2026robust} consider active labeling strategies for general M-estimators, where samples are selected based on predictive accuracy and uncertainty measures. \cite{kato2021role} develops a sampling rule for identifying top items within a multi-armed bandit framework, with the sampling allocation determined by the estimated variances of the arms. \cite{ao2024prediction}  studies active experimental design guided by machine-learning predictions, optimizing policies that determine whether to collect both covariates and responses. \cite{panda2025adaptive} leverages the shared embedding space between queries and LLM representations to develop an adaptive strategy for routing prompts across models. More recently, \cite{dong2026labels} studies the allocation of a fixed budget between numerical outcome labels and pairwise preference labels, deriving an optimal allocation rule that balances cheaper pairwise preference feedback against more costly ground-truth labels. Another closely related work is \cite{liu2024dual}, where the authors greedily select the pair with the largest information gain during the online exploration phase and subsequently employ a pessimism-based policy to identify the optimal arm.

Our framework differs fundamentally from these existing works in several aspects. First, we do not assume the ability to make fully interactive online decisions throughout the data collection process, making our approach applicable to a broader range of practical settings where data are collected in batches or under limited adaptivity. Second, our framework accommodates a much richer class of reward models, requiring only semiparametric consistency of the score function estimators, rather than the linear modeling assumptions often imposed in existing works. Third, our framework also provides a bias-correction strategy to counter model misspecifications, and hence is more robust under outlying preferences. As a result, our methodology offers greater flexibility while retaining strong theoretical guarantees.



\subsection{Our Contributions}\label{sec:contributions}

In summary, our main contributions are three-fold.

\begin{itemize}
    \item We develop a unified framework for adaptive pairwise comparison experiments and policy learning. Methodologically, we propose an adaptive sampling strategy that sequentially allocates comparisons according to the estimated uncertainty and informativeness of different arm pairs, enabling efficient exploration under limited labeling budgets. 
    Using the adaptively collected samples, we further develop a decision policy for identifying the best-performing arm for each context, with a first-order bias correction that provides robustness against model misspecification.
    \item From a theoretical perspective, we establish minimax-optimal guarantees for our procedure, showing that our method achieves the optimal sample complexity for best-arm identification through our adaptive experiment. These results are established under only a semiparametric convergence rate of the underlying estimator, covering many complex machine learning and neural network reward models. Our analysis also characterizes how adaptive allocation improves statistical efficiency compared with uniform sampling strategies.
    \item Empirically, we evaluate our method through extensive simulation studies and three real-world benchmark datasets with applications to large language model evaluation. The results demonstrate that our adaptive strategy can identify the best-performing models more accurately and with significantly fewer comparison samples than existing uniform or non-adaptive approaches.
\end{itemize}


\subsection{Roadmap}
The remainder of the paper is organized as follows. Section~\ref{sec:notations} introduces the mathematical notations used throughout the manuscript. Section~\ref{sec:math-prelim} formulates the best arm identification problem and reviews the preliminary concepts needed for the subsequent development. Section~\ref{sec:strategy-detail} describes the proposed active learning strategy for the experimenter, whose optimality is established in Section~\ref{sec:theory-guarantee} through matching upper and lower regret bounds. To corroborate the theoretical results, Section~\ref{sec:simulation} provides a series of empirical studies demonstrating the advantages of active sampling over uniform random sampling. Finally, Section~\ref{sec:realdata} presents applications to multiple LLM Arena benchmark datasets, illustrating the practical effectiveness of the proposed method for evaluating and comparing large language models.

\subsection{Notation}\label{sec:notations}

Before diving deep into the mathematical formalism, we briefly discuss the notations used throughout the paper to aid the reader. The shorthand notation $[K]$ indicates the set $\{ 1, 2, \dots, K\}$, and $[K]_2$ indicates the set of pairs $\{ (i,j): 1 \leq i < j \leq K \}$. We use the standard Landau big-$O$ and small-$o$ notations along with their probabilistic versions $O_{\prob}$ and $o_\prob$. For a sequence of scalars $\{ a_i \}_{i \in [n]}$, let $\text{vec}(\{ a_i \}_{i \in [n]})$ denote the $n$-length column vector with elements $a_1, \dots, a_n$. For a sequence of matrix (scalar) $\{ \bb{A}_i \}_{i \in [n]}$, let $\text{Diag}(\{ \bb{A}_i \}_{i \in [n]})$ denotes the block-diagonal matrix with diagonal blocks $\bb{A}_1, \dots, \bb{A}_n$. Let $\nabla_{\bb{x}} f(\bb{x})$ denote the gradient vector (matrix) of a differentiable univariate (multivariate) function $f(\bb{x})$, with respect to $\bb{x}$. The notation $\ind{A}$ is used to denote the indicator function of event (or set) $A$ such that it takes the value $1$ if the argument $x \in A$, otherwise it takes the value $0$.

\section{Preliminaries}\label{sec:math-prelim}

\subsection{Problem Setup}\label{sec:setup}

Let $\chi$ be the space of contextual information, and let $\bb{A} = \{ A_1, \dots, A_K \}$ be a finite collection of arms, which we indicate by the index $i \in [K]$ for notational convenience. For example, in the AI evaluation framework, $A_i$ can be viewed as a large language model. The experimenter is given a sampling budget of $T$ rounds of comparisons and aims to identify the best arm for any context $x\in\chi$ with the collected comparisons.

More specifically, in each round $t = 1, 2, \dots, T$, a context $X_t \in \chi$ is generated according to a probability distribution $P_X$. Following the potential outcome framework of Neyman-Rubin~\citep{neyman1923applications, rubin1974estimating}, we associate each context $X_t$  with a collection of latent preference scores $\bb{\theta}(X_t) = (\theta^1(X_t), \dots, \theta^K(X_t))$. Each of these potential outcomes $\theta^a(x)$ can be regarded as a deterministic yet unknown function of the context. In AI evaluation or prompt routing, $\theta^i(X_t)$ can be interpreted as the latent preference score characterizing model $i$'s performance on prompt $X_t$. These latent scores are typically estimated from observed preference data collected from human annotators. Next, we describe the generation process of the human-annotated preference data.

For each round $t\in [T]$,  the experimenter leverages past observations and contextual information to actively select a pair of indices $a_t, b_t \in [K]$. Given this selection, the decision-maker observes a human-annotated binary outcome variable $Y_t$, taking value $1$ if the model $a_t$ performs better than model $b_t$ on the given context $X_t$. At the end of these $T$ rounds, equipped with the data history $\Hcal_T = \{ (X_s, a_s, b_s, Y_s) \}_{s = 1}^T$, the goal is to identify the best arm given the context, i.e., to find
\begin{equation}
    a^{\ast}(x) =  \argmax_{ a \in [K] }~ \theta^{a}(x). 
    \label{eqn:best-arm-defn}
\end{equation}

Commonly, a Bradley--Terry--Luce (BTL) specification is used to model the generative mechanism of the preference outcome $Y_t$. For example, denoting $\tilde{Y}_t^{ab}$ as the potential outcomes and $Y_t$ as the actual observed outcomes, they can be represented as
\begin{equation}
    \tilde{Y}_t^{ab} \mid X_t \sim \text{Ber}(\sigma(\theta^{a}(X_t) - \theta^{b}(X_t))),
    \ 
    Y_{t} = \sum_{a, b \in [K]} \ind{\{ a_t = a, b_t = b \}} \tilde{Y}_{t}^{ab}.
    \label{eqn:Yt-distn}
\end{equation}
\noindent Here, $\sigma(x) = (1 + e^{-x})^{-1}$ is the sigmoid function. This stochastic formulation captures the inherent variability in human judgment as well as the randomness in responses generated by the competing language models. However, under this BTL model formulation, $\bb{\theta}$s are only identifiable up to an additive constant. Specifically, replacing each $\theta^a$ by $\theta^a + c$ for some arbitrary constant $c$ leaves the BTL model in Eq.~\eqref{eqn:Yt-distn} unchanged. To enforce identifiability, we let $\theta^{a_0}(x) = 0$ for a fixed reference arm $a_0 \in [K]$, and without loss of generality, take the reference arm $a_0 = 1$~\citep{li2025efficient}. 

The BTL model may be misspecified in some realistic preference-learning settings. For example, when pairwise preferences exhibit Condorcet cycles or other forms of intransitivity, they cannot be represented by a common set of latent scores under the BTL model~\citep{sun2024rethinking,liu2026statistical}. To accommodate model misspecification, a natural approach is to define the latent preference scores through a minimum-divergence projection, see \cite{parr1980minimum, jones2001comparison, basu2011statistical, hunter2004mm, roy2024robust} and the references therein. Formally, let 
\begin{align}
    m^{ab}(x) & := \E(Y_t \mid a_t = a, b_t = b, X_t = x), \label{eqn:m-a-b-defn}\\
    m_{\bb{\theta}}^{ab}(x) & := \sigma(\theta^a(x) - \theta^b(x)), \ a, b \in [K].
    \label{eqn:m-ab-theta-defn}
\end{align}
\noindent Here, $m^{ab}(x)$ denotes the true conditional preference probability that an arm $a$ is preferred over another arm $b$, whereas $m_{\bb{\theta}}^{ab}(x)$ denotes the preference probability under the BTL model, as illustrated in~\eqref{eqn:Yt-distn}. 

However, under model misspecifications, there may not exist a common set of latent scores that exactly represents all pairwise preference probabilities using the BTL model. We therefore define the target latent preference scores $\bb{\theta}^\ast(x) = (\theta^{a,\ast}(x))_{a \in [K]}$ through the pointwise KL projection 
\begin{equation}
    \bb{\theta}^{\ast}(x) := \argmin_{\theta \in \Theta} ~ d_{\mathrm{KL}}(\bb{m}(x), \bb{m}_{\bb{\theta}}(x)), \ 
    \text{subject to } \theta^{1}(x) = 0,
    \label{eqn:pointwise-divergence}
\end{equation}
\noindent for each $x \in \chi$, provided that the minimizer exists and is uniquely defined under the imposed identifiability constraint. The detailed conditions ensuring existence, uniqueness, and identifiability are discussed in detail in Section~\ref{sec:bias-correction}. 
Here, the quantities $\bb{m}(x)$ and $\bb{m}_{\bb{\theta}}(x)$ are $K(K-1)/2$-dimensional vectors collecting the pairwise preference probabilities $m^{ab}(x)$ and their BTL counterparts $m_{\bb{\theta}}^{ab}(x)$, respectively. 
While it is possible to consider different choices of divergence (or statistical distance) measures in~\eqref{eqn:pointwise-divergence}, we focus on the KL-divergence projection, which corresponds to the maximum-likelihood principle under the BTL working model. This projection approach has also been studied in \cite{li2025efficient}.

\subsection{Experimentation and Decision-making Strategy}\label{sec:strategy-defn}

To efficiently allocate the sampling budget and accurately recommend the best arm, we need to learn two types of policies: a sampling rule that guides the experimenter's data collection process and a recommendation policy that enables the decision-maker to identify the best arm for any given context.

During the data collection process, at each round $t \in [T]$, the experimenter has access to the historical information $\Hcal_{t-1} = \{ (X_s, a_s, b_s, Y_s) \}_{1 \leq s \leq (t-1)}$ as well as the current context $X_t$. A sampling rule $W = (w_1, \dots, w_T)$ is defined as a sequence of conditional sampling distributions, where  $w_t$ denotes the probabilities of a multinomial distribution over the set of unordered arm pairs $\mathcal{P}=\{(a,b): 1 \le a<b\le K\},$ conditional on $\Hcal_{t-1}$ and $X_t$. Given the sampling rule at round $t$, the arm pair $(a_t,b_t)$ is drawn according to the probabilities $w_t$. For notational convenience, let $w_t^{ab} = w_t((a, b))$ for any $1 \leq a < b \leq K$, representing the conditional probability of choosing the pair $(a, b)$ in round $t$. 

At the end of the $T$ rounds of experimentation, the decision-maker learns a recommendation policy, denoted by $\pi(x)$, which is a deterministic function from the context space $\chi$ to the model indices $[K]$. Since the policy is learned from the collected data, $\pi$ is measurable with respect to $\Hcal_T$, the complete dataset obtained after all $T$ rounds of experimentation. The decision-maker then uses the learned policy $\pi \in \Pi \subseteq \{ f: \chi \to [K] \}$ to recommend the arm $\pi(x)$ for any given context $x$. 

\subsection{Regret of a Policy}\label{sec:regret-defn}

Since the ultimate objective is to identify the best arm $a^\ast(x)$ as defined in Eq.~\eqref{eqn:best-arm-defn}, we next introduce a metric to evaluate the performance of the learned recommendation policy.

The value of a recommendation policy $\pi$ is defined as the expected latent preference score of the recommended arm. Specifically, 
\begin{equation}
    Q(\pi; P) = \E_{P_X}\left[ \theta^{\pi(X), \ast}(X) \right] = \E_{P_X}\left[ \sum_{a \in [K]} \ind{\{\pi(X) = a\}}\theta^{a, \ast}(X) \right].
    \label{eqn:Q-func-popn}
\end{equation}
\noindent Here, $P$ denotes the underlying model parameters, including the choice of latent $\theta$s and the marginal distribution of covariates $P_X$. Given the pre-defined policy class $\Pi$, the regret of any policy $\pi\in \Pi$ is given as
\begin{equation}
    R(\pi; P) = Q(\pi^\ast(P); P) - Q(\pi; P),
    \label{eqn:regret-popn}
\end{equation}
\noindent where $\pi^\ast(P) = \argmax_{\pi \in \Pi} Q(\pi; P)$ is the optimal policy in $\Pi$. 

\begin{remark}\label{remark:discrete-optimal-policy}
As an example, consider the special case where $X$ has a discrete distribution. If the policy class $\Pi$ is sufficiently rich so that a policy is allowed to map any $x$ value to any of the $K$ different arms, the optimal policy is trivially given by
\begin{equation*}
    \pi^\ast(P)(x) = a^\ast(x).
\end{equation*}
\noindent To see this, observe that for any policy $\pi\in\Pi$, 
\begin{equation*}
    Q(\pi; P)
    = \sum_{x\in\chi}
    \theta^{\pi(x),\ast}(x) \prob(X=x)
    \leq \sum_{x \in \chi} \theta^{a^\ast(x), \ast}(x) \prob(X = x) = Q(\pi^\ast(P); P),
\end{equation*}
where the inequality follows from the definition of $a^\ast(x)$.
\end{remark}

\section{Optimal Active Preference Learning Strategy}\label{sec:strategy-detail}

Building on the foundations laid out in Section~\ref{sec:math-prelim}, 
we first introduce the experimenter's adaptive sampling strategy 
and then detail how the decision-maker uses the collected data to identify the best arm for each context.

\subsection{Uncertainty-weighted Adaptive Sampling}\label{sec:sampling-strategy-method}

Intuitively, an experimenter would allocate more samples to pairs $(a,b)\in[K]_2$ that exhibit greater uncertainty. When a pair is easy to distinguish, relatively few comparisons may be sufficient. If $m^{ab}(x)=0.9$, meaning that arm $a$ is preferred to arm $b$ with probability $0.9$, then the pair $(a,b)$ can typically be distinguished with only a few comparisons. Therefore, under a fixed sampling budget, the experimenter would allocate fewer comparisons to such a pair, since an additional observation from $(a,b)$ is expected to provide relatively little new information. In contrast, if another pair $(c,d)$ has a win probability $m^{cd}(x)$ close to $0.5$, then the two arms are more difficult to distinguish, and additional comparisons are needed.

This intuition motivates a sampling rule that assigns weights proportional to the conditional variance of $Y_t$ given the selected pair, i.e.,
\begin{equation*}
    w_t((a, b)) \propto \var(Y_t \mid a_t = a, b_t = b, X_t = x) = m^{ab}(x) (1 - m^{ab}(x)) := v^{ab}(x).
\end{equation*}
\noindent Here, we use the fact that $Y_t$ is a Bernoulli random variable (cf.~\eqref{eqn:Yt-distn}). Indeed, when a pair is easy to distinguish, its win probability is closer to either $0$ or $1$, resulting in a smaller conditional variance. Conversely, when the two arms within a pair are difficult to distinguish, so that the corresponding win probability is close to $0.5$, the pair has larger conditional variance and should receive more comparisons.

To implement the above procedure empirically, at round $t$, given the available data $\Hcal_{t-1}$, the experimenter obtains an estimate of $m^{ab}(x)$ denoted by $\widehat{m}_t^{ab}(x)$ for each $(a,b) \in [K]_2$ and $x\in\chi$. Additionally, we require the estimator to satisfy the symmetry conditions
\begin{equation*}
    \widehat{m}_t^{aa}(x) = 0.5, \ 
    \widehat{m}_t^{ab}(x) = 1 - \widehat{m}_t^{ba}(x)
\end{equation*}
\noindent for any $a, b \in [K]$. Further properties of this estimator will be discussed later in Section~\ref{sec:theory-guarantee}. Using these estimates, at round $t$, the experimenter adopts the corresponding plug-in sampling rule with weights $\widehat{w}_t^{ab}(X_t)$ where,
\begin{equation}
    \widehat{w}_t^{ab}(x) = \dfrac{ \widehat{v}_t^{ab}(x) }{ \sum_{(a', b') \in [K]_2} \widehat{v}_t^{a'b'}(x) }, \ \text{ for any } 1 \leq a < b \leq K.
    \label{eqn:weights-emp-defn}
\end{equation}
\noindent where $\widehat{v}_t^{ab}(x) = \widehat{m}_t^{ab}(x)(1 - \widehat{m}_t^{ab}(x))$. Note that the normalization in the denominator is taken only over the unordered pairs in $[K]_2$ (i.e., $1 \leq a < b \leq K$)
since each pair of distinct arms corresponds to a single comparison, regardless of ordering. Self-comparisons are excluded, i.e., $\widehat w_t^{aa}(x)=0$ for all $a\in[K]$.

\subsection{Robust Recommendation via Bias-Corrected Projection}\label{sec:bias-correction}

Next, we describe the decision-making procedure based on the data collected with the proposed sampling algorithm.

Recall that the goal of the decision-maker is to construct a recommendation policy that identifies the best arm given any context $x$. This recommendation policy is typically obtained by maximizing an empirical version of the value function $Q(\pi, P)$ in \eqref{eqn:Q-func-popn}, based on the full dataset $\Hcal_T$ collected at the end of round $T$. From the expression of $Q(\pi,P)$, a key ingredient of this value function estimation is to obtain a reasonable estimator of the latent score $\theta^a(x)$ for each arm $a\in [K]$. Due to the arguments described in Section~\ref{sec:setup}, we obtain these estimates $\hat{\theta}^a(x)$ by projecting the estimated preference probabilities $\hat{m}_t^{ab}(x)$ onto the BTL model probabilities $m_{\bb{\theta}}^{ab}(x)$ as in~\eqref{eqn:m-ab-theta-defn}; The technical details underlying this projection are described below. Let 
\begin{equation}
    \bb{U}(\bb{\theta}, \bb{m}, x) = \mathrm{vec}\left( \left\{ \sum_{b \in [K] \setminus \{a\} } \rho^{ab}(x) (\sigma({\theta}^a(x) - {\theta}^b(x)) - m^{ab}(x)) \right\}_{a \in [K] \setminus \{1\} } \right).
    \label{eqn:U-defn}
\end{equation}
\noindent be the score function, where $\rho^{ab}(x)$ is an exogenous choice of weights reflecting the relative importance of accurately estimating the preference difference between two arms $a$ and $b$, conditional on the covariate value $x$. A common choice is uniform weighting, i.e., $\rho^{ab}(x) = 2/K(K-1)$. We assume that $\sum_{(a, b) \in [K]_2} \rho^{ab}(x) = 1$. Generalizing Eq.~\eqref{eqn:pointwise-divergence}, the target parameter $\bb{\theta}^\ast(x)\in \R^{K-1}$, representing the projected latent preference scores under the BTL model, is defined as a solution to the population score equation 
\begin{equation}
    \bb{U}(\bb{\theta}^\ast(x), \bb{m}, x) = \bb{0}.
    \label{eqn:theta-ast-defn}
\end{equation}
\noindent Thus, $\bb{\theta}^\ast(x)$ is the population-level target that only depends on the oracle pairwise comparison probabilities $\{m^{ab}(x)\}$ and exogenously specified weights $\rho^{ab}(x)$. 

Given the estimated pairwise comparison probabilities $\widehat{m}_t^{ab}(x)$ at round $t$, 
we estimate the target parameter $\bb{\theta}^\ast(x)$ by solving the empirical counterpart of~\eqref{eqn:U-defn}.
In particular, for any fixed $x\in 
\mathcal{X}$, the estimator $\bb{\widehat\theta}_t(x)$ is obtained through the following equations:
\begin{equation}
    \sum_{b \in [K] \setminus \{a\} } \rho^{ab}(x) (\sigma(\theta^a(x) - \theta^b(x)) - \widehat{m}_t^{ab}(x)) = 0, \ \text{ for all } a \in [K] \setminus \{1 \}.
    \label{eqn:estimating-eqn}
\end{equation}
\noindent  Equivalently, $\bb{U}(\bbhat{\theta}_t, \bbhat{m}_t, x) = \bb{0}$. This procedure can be interpreted as projecting the observed (or estimated) pairwise comparison probabilities $\widehat{m}_t^{ab}(x)$ onto the class of BTL models.
\begin{remark}\label{remark:btl-exact}
    When the true data-generating model coincides with the Bradley-Terry-Luce model class, then we have 
    \begin{equation*}
        \sigma(\theta^{a,\ast}(x) - \theta^{b, \ast}(x)) = m^{ab}(x),
    \end{equation*}
    \noindent for all $a, b \in [K]_2$, by definition. In this correctly-specified case, the population counterpart of the estimating equation in Eq.~\eqref{eqn:estimating-eqn} holds with $\bb{\theta}^\ast(x) = (\theta^{a,\ast}(x))_{a \in [K]},$ regardless of the choice of the cost weights $\rho_t^{ab}(x)$. 
\end{remark}

In general, the existence and uniqueness of solutions to the population and empirical estimating equations require some regularity conditions. At the population level, if the pairwise comparison probabilities $m^{ab}(x)$ are generated from a Bradley--Terry--Luce (BTL) model, then Eq.~\eqref{eqn:estimating-eqn} admits the corresponding BTL latent scores as its unique solution, up to the imposed identifiability constraint, for any choice of positive weights $\rho^{ab}(x)$'s. 
Even when the BTL is misspecified, the existence and uniqueness of $\bb{\theta}^\ast(x)$, 
up to a constant shift, are guaranteed provided $m^{ab}(x)$ is bounded away from $0$, 
a condition formalized later in Assumption~\ref{assum:positivity}.
This result follows from a classical result of~\cite{ford1957solution}, 
who establishes that the BTL estimation equations admit a unique solution, 
up to an additive constant, if and only if the underlying directed comparison graph is strongly connected.


At the empirical level, existence and uniqueness of the solution to~\eqref{eqn:estimating-eqn} similarly depend on sufficient connectivity of the observed comparison graph. Strong connectivity requires that, for every nonempty proper subset $I \subset [K]$, there exists at least one observed comparison in which an arm in $I$ is preferred over an arm in $I^c$. Under standard regularity conditions ensuring a sufficiently well-connected comparison graph, this property holds asymptotically with high probability~\citep{han2023general, fang2026recent}. In practice, this well-connectedness assumption can be satisfied by uniformly sampling across all pairs of arms during an initial short exploration phase; see Section~\ref{sec:simulation} for implementation details. A similar strategy is adopted by~\cite{kato2024adaptive}. These conditions ensure that the empirical estimating equation admits a unique solution with high probability.

Based on the estimated latent preference scores 
$\{\widehat{\theta}_t^a(X_t)\}_{a \in [K]}$, a plug-in approximation to the value of a policy $\pi$ can be constructed as
\begin{equation*}
    \widehat{Q}_T^{\mathrm{plug}}(\pi)
    = \frac{1}{T} \sum_{t=1}^T
    \sum_{a \in [K]}
    \ind{\{ \pi(X_t) = a \}} \widehat{\theta}_t^a(X_t).
\end{equation*}
\noindent However, under model misspecification, the projected BTL scores may not recover the underlying preference structure, and consequently this plug-in estimator can be biased for the true policy value. It is, therefore, possible to construct a first-order bias-corrected estimator of the policy value by augmenting the plug-in estimator with a correction term based on the influence function of $\widehat{\theta}_t^a$.  Leveraging the form of the influence function for pairwise comparison data, derived in Proposition 3.1 of~\cite{li2025efficient}, we define the bias-corrected empirical value function as 
\begin{equation}
    \widehat{Q}_T(\pi) = \dfrac{1}{T}\sum_{t = 1}^T \sum_{a \in [K]} \ind{\{ \pi(X_t) = a \}} \widehat{\Gamma}_t^a,
    \label{eqn:empirical-Q}
\end{equation}
\noindent where $\widehat{\Gamma}_t^1 = 0$ and $\bbhat{\Gamma}_t = (\widehat{\Gamma}_t^2, \dots, \widehat{\Gamma}_t^K)\tr$ is expressed as 
\begin{equation}
    \bbhat{\Gamma}_t = \bbhat{\theta}_t(X_t) - \left[ \nabla_{\bb{\theta}} \bb{U}\vert_{\bb{\theta} = \bbhat{\theta}_t(X_t), \bb{m} = \bbhat{m}_t, x = X_t} \right]^{-1} \left[ \nabla_{\bb{m}}\bb{U}\vert_{\bb{\theta} = \bbhat{\theta}_t(X_t), \bb{m} = \bbhat{m}_t, x = X_t} \right] \bbhat{e}_t(X_t).
    \label{eqn:gamma-defn}
\end{equation}
\noindent Here, $\bbhat{\theta}_t(X_t)$ is the $(K-1)$-length vector obtained by solving the system of estimating equations~\eqref{eqn:estimating-eqn} for $x = X_t$, 
\begin{align}
    \bbhat{e}_t(X_t) 
    & = \text{vec}\left( \left\{\dfrac{\bbhat{1}_t^{ab}}{\widehat{w}_t^{ab}(X_t)} (Y^{ab}_t - \widehat{m}_t^{ab}(X_t)) \right\}_{a \in [K] \setminus \{1\}, b \in [K] \setminus \{a\} } \right), \label{eqn:ehat-theta}\\
    \left[ \nabla_{\bb{\theta}}\bb{U} \right]_{ij}
    & = \begin{cases}
        \displaystyle\sum_{b \in [K] \setminus \{ i\} } \rho^{ib}(x) \sigma({\theta}^i(x) - {\theta}^b(x)) (1 - \sigma({\theta}^i(x) - {\theta}^b(x))), & i = j\\
        -  \rho^{ij}(x)\sigma({\theta}^i(x) - {\theta}^j(x)) (1 - \sigma({\theta}^i(x) - {\theta}^j(x))), & i \neq j,
    \end{cases}
    \label{eqn:del-uhat-theta} \\
    \text{ and, } 
    \left[ \nabla_{\bb{m}}\bb{U} \right]
    & = \text{Diag}( \{ \partial_{\bb{m},i}\bb{U} \}_{i \in [K] \setminus \{1\}} ),    
    \label{eqn:del-u-m}
\end{align}
\noindent with
\begin{equation*}
    \nabla_{\bb{m},i}\bb{U}(x)
    = [- \rho^{i1}(x), \dots, -\rho^{i,(i-1)}(x), -\rho^{i,(i+1)}(x) \dots, -\rho^{iK}(x)], \ i \in [K] \setminus \{1\}.
\end{equation*}
\noindent Also, the indicator variable $\bbhat{1}_t^{ab}$ is shorthand for
$\ind{\{(a_t,b_t)\in \{(a,b),(b,a)\}\}}$, indicating that the set of 
sampled arms $\{a_t,b_t\}$ coincides with the pair $\{a,b\}$. Under the 
sampling scheme based on the estimated weights $\widehat{w}_t^{ab}(X_t)$, 
each sampled pair contributes to both the $(a,b)$ and $(b,a)$ entries. As a result, $\bbhat{e}_t(x)$ is a sparse vector of length $(K-1)^2$ with two nonzero entries. The quantity $Y_t^{ab}$ is defined for both entry $(a,b)$ and $(b,a)$ as follows:
{\begin{equation*}
    Y_t^{ab} = \begin{cases}
        Y_t & \text{ if } a<b\\
        1 - Y_t & \text{ if } a>b \\
        0 & \text{ otherwise},
    \end{cases}
    \qquad 
    \text{ for all } a, b \in [K].
\end{equation*}}
\noindent As noted above, we only sample pairs of arms indexed by  $1\leq a<b\leq K$. Therefore, for the reverse entry $(b,a)$,  we assign the complementary outcome $1-Y_t$ and recall that we also have $\widehat m_t^{ba}(x)=1-\widehat m^{ab}(x)$.

Once the empirical value function $\widehat{Q}_T(\pi)$ is computed at the end of $T$ rounds according to Eq.~\eqref{eqn:empirical-Q}, the decision-maker selects a policy from a prespecified policy class $\Pi$ by maximizing this empirical objective, i.e.,
\begin{equation}
    \widehat{\pi}_T := \argmax_{\pi \in \Pi} \widehat{Q}_T(\pi).
    \label{eqn:recommender-policy}
\end{equation}
\noindent The entire process is summarized in Algorithm~\ref{alg:main-algo}.

\begin{algorithm}[th]
    \DontPrintSemicolon
    \KwInput{Sampling budget $T$, number of arms $K$, policy class $\Pi$, judge function $J : \chi \times [K]^2 \to \{0, 1\}$.}
    Initialize historical dataset $\Hcal \gets \emptyset$\;
    \For{$t = 1$ \KwTo $T_0 (< T)$}{
        Generate $X_t \sim \Pcal$\;
        Sample $a_t \sim \mathrm{Discrete Uniform}([K])$, $b_t \sim \mathrm{Discrete Uniform}([K] \setminus \{a_t\} )$\; 
        $Y_t \gets J(X_t, a_t, b_t)$\;
        $\Hcal \gets \Hcal \cup \{ (X_t, a_t, b_t, Y_t) \}$
    }
    \For{$t = T_0 + 1$ \KwTo $T$}{
        Estimate $\widehat{m}_t^{ab}$ using $\Hcal$\;
        Set $\widehat{m}_t^{ab} \gets \max (\delta, \min (1-\delta, \widehat{m}_t^{ab}))$\;
        Generate $X_t \sim \Pcal$\;
        \For{$a \gets [K]$, $b \gets [K] \setminus [a] $}{
            $v_t^{ab} \gets \widehat{m}_t^{ab}(X_t)(1 - \widehat{m}_t^{ab}(X_t))$\;
        }
        Sample $(a_t, b_t)$ according to $\prob(a_t = a, b_t = b) \propto v_t^{ab}$\;
        $Y_t \gets J(X_t, a_t, b_t)$\;
        $\Hcal \gets \Hcal \cup \{ (X_t, a_t, b_t, Y_t) \}$\;
        Solve Eq.~\eqref{eqn:estimating-eqn} to obtain $\widehat{\theta}^a(x)$\;
        Use Eq.~\eqref{eqn:gamma-defn} to obtain $\bbhat{\Gamma}_t$\;
    }
    $\widehat{\pi}_T \gets \argmax_{\pi \in \Pi} T^{-1}\sum_{t=1}^T \sum_{a \in [K]} \ind{\{ \pi(X_t) = a \}} \widehat{\Gamma}_t^a$\;
    \caption{Proposed active learning framework for preference comparison data.}
    \label{alg:main-algo}
\end{algorithm}

To solve the optimization problem in Eq.~\eqref{eqn:recommender-policy}, one may employ either closed-form solutions, when available, or standard numerical optimization procedures. For example, when context space is discrete with $\chi = \{ x_1, x_2, \dots, \}$ and the policy class $\Pi$ contains all possible functions from $\chi$ to $[K]$, the regret-minimizing policy admits the following explicit form:
\begin{equation}
    \widehat{\pi}_T(x_i) = \argmax_{a \in [K]} \sum_{t=1}^T \widehat{\Gamma}^a_t \ind{\{ X_t = x_i\} }, \ i = 1, 2, \dots.
    \label{eqn:best-policy-kronecker}
\end{equation}
\noindent This follows directly from the representation of the empirical value function
\begin{equation*}
    \widehat{Q}_T(\pi) 
    = \frac{1}{T}\sum_{t=1}^T \widehat\Gamma_t^{\pi(X_t)} 
    = \sum_{i} \sum_{a \in [K]} \ind{\{ \pi(x_i) = a \} } \left( \frac{1}{T}\sum_{t=1}^T \widehat{\Gamma}_t^a \ind{ \{ X_t = x_i \} }\right).
\end{equation*}
\noindent In the case of continuous covariates, one may instead consider a parameterized class of soft policies. For example, one can define the conditional distribution
\begin{equation*}
    \pi'_{\bb{\beta}}(a \mid x) = \frac{e^{ g_{\bb{\beta}}(x, a) }}{\sum_{b \in [K] } e^{ g_{\bb{\beta}}(x, b)}}, \  x \in \chi,
\end{equation*}
\noindent where $g_{\bb{\beta}} : \chi \times [K] \to \R$ is a sufficiently smooth function with $g_{\bb{\beta}}(\cdot, 1) = 0$ and $\bb{\beta}$ is the unknown parameter to be estimated. Then, the optimal soft policy within the specified policy class can be obtained by maximizing
\begin{equation*}
    \widehat{Q}_T(\pi'_{\bb{\beta}}) = \frac{1}{T}\sum_{t=1}^T \sum_{a \in [K]} \pi'_{\bb{\beta}}(a \mid X_t) \hat{\Gamma}_t^a.
\end{equation*}
\noindent Depending on the specific parameterization of $g_{\bb{\beta}}(\cdot)$, this optimization problem can be solved using standard numerical optimization procedures, such as gradient-based methods. Popular choices for $g_{\bb{\beta}}$ include linear functions of the form $g_{\bb{\beta}}(x, a) = \bb{\beta}_a f(x)$ where $f(x)$ is a prespecified feature map, as well as more flexible function classes such as, decision trees, random forest, neural networks of a specific architecture, etc. After estimating $\bb{\beta}$, a deterministic recommendation policy can be obtained by selecting the action with the largest fitted score. Specifically, 
\begin{equation}
    \widehat{\pi}_T(x) = \argmax_{a \in [K]} g_{\bbhat{\beta}}(x, a), \textrm{ where }\ 
    \bbhat{\beta} = \argmax_{\bb{\beta}} \widehat{Q}_T(\pi'_{\bb{\beta}}).
    \label{eqn:best-policy-linear}
\end{equation}

\section{Theoretical Guarantees}\label{sec:theory-guarantee}

In this section, we study the theoretical properties of our proposed algorithm, which consists of an adaptive sampling component and a recommendation component, by analyzing the regret of the resulting policy $\hat \pi_T$.
Specifically, we establish a lower bound of order $T^{-1/2}$ of the expected simple regret for any algorithm producing a recommendation policy in a given exogenous class $\Pi$, and show that the proposed recommendation policy $\widehat\pi_T$ in Eq.~\eqref{eqn:recommender-policy} attains a matching upper bound of order $T^{-1/2}$. This is shown under a set of mild assumptions and requires only semiparametric consistency of the pairwise comparison probabilities $\widehat{m}^{ab}(x)$. Before presenting the formal mathematical statements of these results, we first introduce the key assumptions underlying our analysis.

\begin{enumerate}[label=(A\arabic*)]
    \item\label{assum:unconfound} (\textbf{Unconfoundedness}) For every pair of arms $(a,b)$, the selection indicator $\bb{1}_t^{ab}$ is conditionally independent of the potential feedback $\tilde{Y}_t^{ab}$ given the context $X_t$ and historical observations $\Hcal_{t-1}$, i.e.,
    \begin{equation*}
        \bb{1}_t^{ab} \perp\!\!\!\perp \tilde{Y}_t^{ab} \mid \mathfrak{F}(\{ X_t \} \cup \Hcal_{t-1}),
        \qquad \text{for all } (a,b),   
    \end{equation*}
    \noindent where $\mathfrak{F}(S)$ denotes the $\sigma$-algebra generated by $S$.
    \item\label{assum:positivity}  (\textbf{Positivity}) There exists a $\delta$ such that for any $x \in \chi$, we have 
    \begin{equation*}
        0 < \delta \leq \inf_{(a,b) \in [K]_2} {m}^{ab}(x) \leq \sup_{(a,b) \in [K]_2} {m}^{ab}(x) \leq (1 - \delta) < 1.
    \end{equation*}
    \noindent The same is true for the estimates $\widehat{m}_t^{ab}(x)$ and for the exogenous weights $\rho^{ab}(x)$.
    \item\label{assum:estimate-consistency} (\textbf{Consistency of the pair-comparison probability estimates}): The estimates of the pair-comparison probabilities satisfy a consistency condition in $L^2$-norm with at least a semiparametric efficiency, i.e., for any choice of $(a, b) \in [K]_2$,
    \begin{equation*}
        \E\left[ \sup_{x \in \chi} \vert \widehat{m}_t^{ab}(x) - m^{ab}(x)\vert^2 \right] = \frac{a_t}{t^{1/2}},
    \end{equation*}
    \noindent where $a_t = o(1)$ and does not depend on the context distribution $P$.
\end{enumerate}

Assumptions~\ref{assum:unconfound} and~\ref{assum:positivity} are standard assumptions in the causal inference literature~\citep{rubin2005causal} and are naturally satisfied in our problem setting. Moreover, under Assumption~\ref{assum:unconfound}, when the true preference-generating mechanism follows the BTL model, the terms in~\eqref{eqn:m-a-b-defn}-\eqref{eqn:m-ab-theta-defn} coincide at the true latent preference scores. Assumption~\ref{assum:positivity} also ensures that degeneracy is avoided in the estimation process, i.e., when solving Eq.~\eqref{eqn:estimating-eqn}. The bounds on the $\hat{m}_t^{ab}(x)$ can simply be added in the algorithmic design, by clipping the estimated probability if it is beyond the interval $[\delta, 1- \delta]$. Assumption~\ref{assum:estimate-consistency} imposes a mild semiparametric convergence requirement on the estimated pairwise comparison probabilities and accommodates a broad class of consistent estimators, including nonparametric methods such as machine learning and neural network models that may converge slower than the classical parametric rate. Let $D_t = \sup_{x \in \chi}\vert \widehat{m}_{t}^{ab}(x) - m^{ab}(x)\vert$. To understand condition~\ref{assum:estimate-consistency}, pick any threshold $\epsilon > 0$, and obtain that
\begin{equation}
    \E(D_t^2) = \E(D_t^2 \bb{1}_{ \{ D_t \leq \epsilon t^{-1/4} \} }) + \E(D_t^2 \bb{1}_{ \{ D_t > \epsilon t^{-1/4} \} })
    \leq \epsilon^2 t^{-1/2} + \prob(D_t > \epsilon t^{-1/4}),
    \label{eqn:strong-mhat-convergence}
\end{equation}
\noindent since both $\widehat{m}_t^{ab}(x), m^{ab}(x) \in (0, 1)$. Therefore, as long as $\prob(D_t > \epsilon t^{-1/4}) = o(t^{-1/2})$, i.e., the difference $D_t$ decays stochastically at least at the rate of $t^{-1/4}$, with high probability greater than or equal to $1 - o(t^{-1/2})$, Assumption~\ref{assum:estimate-consistency} follows.

\begin{remark}\label{remark:example-estimator}

To provide a few examples of estimators satisfying Assumption~\ref{assum:estimate-consistency}, beyond parametric methods such as logistic regression, one may use kernel regression or support vector regression to estimate $\widehat{m}^{ab}(x)$. Under suitable regularity conditions, kernel-based estimators achieve an error rate of $O_\prob(\sqrt{\gamma_t/t})$, where $\gamma_t$ denotes the maximum information gain associated with the choice of kernel~\citep{chowdhury2017kernelized}.  For the standard Gaussian kernel with a $d$-dimensional context, $\gamma_t = O((\log t)^{d+1})$ yielding an error of $O(t^{-1/2}(\log t)^{(d+1)/2})=o(t^{-1/4})$. For tree-based estimators, convergence rates of the form $O_\prob(t^{-\beta(d)})$ can be established, where $\beta(d)>0$ depends on the context dimension $d$ and the underlying regularity conditions~\citep{mourtada2017universal,wager2018estimation}. Under suitable restrictions on the context dimension $d$, these rates can also satisfy the required $o_\prob(t^{-1/4})$ convergence condition. Additionally, for multilayer perceptrons and other deep neural network architectures, convergence rates sufficient for semiparametric inference have also been established under appropriate regularity conditions~\citep{farrell2021deep}. While these results establish in-probability convergence, additional regularity conditions may be imposed on the estimator (e.g., as shown in Eq.~\eqref{eqn:strong-mhat-convergence}) to ensure Assumption~\ref{assum:estimate-consistency} holds.
\end{remark}

At this point, before introducing the relevant technical lemmas, we first highlight several facts that simplify the notation used throughout the subsequent analysis. Specifically, the gradient matrix $\nabla_{\bb{\theta}}\bb{U}$ depends only on the preference parameters $\bb{\theta}$, and is independent of $\bb{m}$. Additionally, the gradient matrix $\nabla_{\bb{m}}\bb{U}$ is independent of both and therefore can be viewed as a function of the covariate $x$ alone. Therefore, we write $\nabla_{\bb{\theta}}\bb{U} = \nabla_{\bb{\theta}}\bb{U}(\bb{\theta}, x)$ and $\nabla_{\bb{m}}\bb{U} = \nabla_{\bb{m}}\bb{U}(x)$. We further use the shorthand $\nabla_{\bb{\theta}}\bbhat{U}_{t} = \nabla_{\bb{\theta}}\bb{U}(\bbhat{\theta}_t(X_t), X_t)$ and its population counterpart $\nabla_{\bb{\theta}}\bb{U}^\ast_{t} = \nabla_{\bb{\theta}}\bb{U}(\bb{\theta}^\ast(X_t), X_t)$. 

Equipped with these notations, we now present two direct implications of Assumptions~\ref{assum:positivity}-\ref{assum:estimate-consistency}. First, the sampling weights $\widehat{w}_t^{ab}(x)$ and the debiased values $\bbhat{\Gamma}_t$ remain uniformly bounded over $x \in \chi$. Second, together with the dominated convergence theorem, these assumptions also ensure that the convergence rate of the nonparametric estimates $\widehat{m}^{ab}_t(x)$ propagates to the sampling weights $\widehat{w}_t^{ab}(x)$, the estimated preference parameters $\bb{\theta}_t(x)$, and the gradients of the $U$-function, yielding corresponding convergence rates for these quantities. These results are formalized in the following two lemmas.

\begin{lemma}\label{lem:gamma-est-bound}
    Under Assumption~\ref{assum:positivity}, for all $(a, b) \in [K]_2$, the quantities $\widehat{w}_t^{ab}(x)$, $\widehat{\theta}_t^a(x)$, and $\widehat{\Gamma}_t^a$ are uniformly bounded over all $x \in \chi$ and $t \geq 1$. The population preference parameter $\theta^{a,\ast}(x)$ is also uniformly bounded over $x \in \chi$ and $a \in [K]$.
\end{lemma}

\begin{lemma}\label{lem:consistency-bounds}
    Under Assumptions~\ref{assum:positivity}-\ref{assum:estimate-consistency}, the following bounds hold;
    \begin{align*}
        \E\left[ \sup_{x \in \chi} \sup_{a \in [K]} \vert \widehat{\theta}^a_t(x) - \theta^{a,\ast}(x) \vert^2 \right] & = o(t^{-1/2}),\\
        \E\left[ \sup_{x \in \chi} \sup_{(a,b) \in [K]_2} \vert \widehat{v}^{ab}_t(x) - v^{ab}(x) \vert^2 \right] & = o(t^{-1/2}),\\
        \E\left[ \sup_{x \in \chi} \sup_{(a,b) \in [K]_2} \vert \widehat{w}^{ab}_t(x) - w^{ab}(x) \vert^2 \right] & = o(t^{-1/2}),\\
        \E\left[ \sup_{x \in \chi} \Vert [\nabla_{\bb{\theta}}\bb{U}(\bbhat{\theta}_t(x), x)] - [\nabla_{\bb{\theta}}\bb{U}(\bb{\theta}^\ast(x), x)] \Vert_F^2 \right] & = o(t^{-1/2}).
    \end{align*}
    \noindent Here, $\bb{\theta}^\ast(x) = (\theta^{2,\ast}(x), \dots, \theta^{K,\ast}(x))\tr$ as in Eq.~\eqref{eqn:theta-ast-defn} and $\Vert \cdot\Vert_F$ indicates Frobenius norm. Each of the above $o(\cdot)$-bound is independent of $P$, the distribution of the context.
\end{lemma}
\noindent The proofs of these lemmas are deferred to Appendices~\ref{appendix:proof-lem-gamma-bound} and~\ref{appendix:proof-lem-consistency-bounds}. Notably, the probabilistic guarantees presented in Lemma~\ref{lem:consistency-bounds} inherit the high-probability guarantee imposed in Assumption~\ref{assum:estimate-consistency}. 
In particular, when Assumption~\ref{assum:estimate-consistency} holds with an exponentially decaying tail probability, the same results also hold in Lemma~\ref{lem:consistency-bounds}.

\subsection{Regret Lower Bound}\label{sec:lower-bound}
Let $\Acal$ denote any algorithm that combines a sampling rule and  a recommendation rule to output $\hat \pi_{\Acal,T}\in \Pi$ after $T$ rounds  of sampling. Our proposed Algorithm~\ref{alg:main-algo} is one such algorithm. In this section, we establish a minimax lower bound on the expected regret $R(\hat{\pi}_{\Acal, T}; P)$ defined in Eq.~\eqref{eqn:regret-popn} that holds for any such algorithm $\Acal$.

Before introducing the main results, we introduce a measure of the complexity of the policy class $\Pi$, which plays a key role in characterizing the regret lower bound. For a policy class $\Pi$, its Natarajan dimension $d_N(\Pi)$ is defined as the maximal cardinality of a set shattered by the policy class $\Pi$~\citep{natarajan1989learning}. It is a generalization of the VC-dimension~\citep{vapnik1971uniform}. A formal definition of Natarajan dimension is as follows.
\begin{definition}\label{defn:natarajan-dim}
    Given a $K$-action policy class $\Pi$, i.e., $\Pi \subseteq \{ \pi: \chi \to [K] \}$ is said to shatter a set $\{ x_1, \dots, x_m\}$ of $m$ points if there exists 
    \begin{enumerate}
        \item[(a)] $u_i \neq v_i \in [K]$ for all $i = 1, \dots, m$.
        \item[(b)] For any binary vector $\bb{w} = (w_1, \dots, w_m) \in \{0, 1\}^m$, there exists a policy $\pi^{\bb{w}} \in \Pi$ such that
        \begin{equation*}
            \pi^{\bb{w}}(x_i) = \begin{cases}
                u_i & \text{ if } w_i = 1,\\
                v_i & \text{ if } w_i = 0.
            \end{cases}, 
            \ i = 1, 2, \dots, m.
        \end{equation*}  
    \end{enumerate}
    \noindent The Natarajan dimension $d_N(\Pi)$ of a policy class $\Pi$ is defined as the size of the largest set of points which can be shattered by $\Pi$.
\end{definition}
\noindent We now present the resulting regret lower bound in terms of the Natarajan 
dimension. 

\begin{theorem}\label{thm:regret-lower-bound}
    Let $\hat{\pi}_{\Acal, T} \in \Pi$ be the recommendation policy produced by an algorithm $\Acal$, and $\Pi$ be an exogenous policy class. Then, the expected simple regret satisfies a minimax lower bound
    \begin{equation}
        \inf_{\Acal} \sup_{P} \dfrac{\E(R( \hat{\pi}_{\Acal, T}, P )) }{\sqrt{\E_{P} \left[ \sum_{a \in [K]} (\theta^a(X) - \bar{\theta}(X))^2 + \trace{\bb{\Sigma}_A(X)} \sum_{(a, b) \in [K]_2} v^{ab}(X) \right]}} \geq c\dfrac{\sqrt{d_N(\Pi)}}{\sqrt{T}}.
        \label{eqn:regret-lower-bound}
    \end{equation}
    \noindent for all sufficiently large $T$, where 
    \begin{align}
        \bar{\theta}(X) & = K^{-1}\textstyle\sum_{a \in [K]} \theta^a(X), \nonumber \\
        \bb{A}(x) & = [\nabla_{\bb{\theta}}\bb{U}(\bb{\theta}, x)]^{-1} [\nabla_{\bb{m}}\bb{U}(x)], \label{eqn:A-defn}\\
        \bb{\Sigma}_A(X) & = \bb{A}(x) (\bb{A}(x))\tr, \label{eqn:sigma-A-defn}
    \end{align}
    \noindent and $c$ is a constant depending only on $K$ and $\delta$, given by
    \begin{equation*}
        c = \frac{\delta^{(K+1)}(1 - \delta)^{K}}{2K\sqrt{K-1}(\delta^K + (1- \delta)^K)^2}.
    \end{equation*}
\end{theorem}

\noindent The key idea in the proof of Theorem~\ref{thm:regret-lower-bound} is to carefully construct a collection of model parameters associated with each binary vector of length $d_N(\Pi)$. They are chosen in a way such that the average regret of any policy $\hat{\pi}_{\Acal, T} \in \Pi$ obtained from any strategy $\Acal$ over all possible $2^{d_N(\Pi)}$ binary vectors is not too small.
This, in turn, implies that there exists at least one choice of the model parameter under which $\hat{\pi}_{\Acal,T}$ incurs a non-negligible regret.

\vspace*{1em}
\noindent\textbf{Step 1: Choice of the model parameters.} Let $d_N(\Pi) = m$. Then, by the definition of the Natarajan dimension, there exist tuples $\{(x_i, u_i, v_i)\}_{i=1}^m,$ where $x_i \in \chi, u_i \in [K], v_i \in [K],$ such that for any binary vector $\bb{w} = (w_1, \dots, w_m) \in \{0, 1\}^m$, there exists a corresponding policy $\pi^{\bb{w}} \in \Pi$ satisfying  $\pi^{\bb{w}}(x_i)=w_i u_i + (1-w_i)v_i,i\in[m]$. 

For each binary vector $\bb{w} \in \{0,1\}^m$, we define a corresponding data-generating distribution, denoted by $P_{\bb{w},\Delta}$, as follows.
\begin{enumerate}
    \item Generate the context $X \sim \text{Uniform}\{ x_1, \dots, x_m \}$.
    \item Define 
    \begin{equation*}
        \alpha_{\bb{w}, i} = \begin{cases}
            u_i & \text{ if } w_i = 1,\\
            v_i & \text{ if } w_i = 0.
        \end{cases}
    \end{equation*}
    \item Given the context $x_i$, the latent parameters are chosen such that 
    \begin{equation*}
        \theta^{a}(x_i) = \eta^a(x_i) - \eta^1(x_i), \ \textrm{ where }
        \eta^{a}(x_i) = \begin{cases}
            \Delta & \text{ if } a = \alpha_{\bb{w}, i},\\
            0 & \text{ otherwise}.
        \end{cases}
    \end{equation*}
\end{enumerate}
\noindent Intuitively, under the model $P_{\bb{w},\Delta}$, all arms have the same latent preference score except for a single arm $\alpha_{\bb{w}, i} \in [K]$, whose score is higher by $\Delta > 0$ and is therefore the unique best arm.

\vspace*{1em}
\noindent\textbf{Step 2: Average regret under two related models.} For a binary vector $\bb{w}$, let $\bb{w}^{(i)}$ denote the binary vector obtained by flipping its $i$-th bit. It turns out that the Kullback-Leibler (KL) divergence between the distribution of the entire dataset $\Hcal_T$ under these two different models $P_{\bb{w},\Delta}$ and $P_{\bb{w}^{(i)},\Delta}$ is bounded above by $T\Delta^2/(2d_N(\Pi))$, as formalized in the following Lemma. Furthermore, for any policy $\hat{\pi}_{\Acal,T}$ produced by an algorithm $\Acal$ after $T$ rounds of sampling, its error probabilities are related to this KL-divergence, yielding a difference in regret under two related models.

\begin{lemma}\label{lem:KL-div-gap}
    Let $P_{\bb{w},\Delta}$ denote the model described in Step 1, let $\bb{w}^{(i)}$ be the binary vector obtained from $\bb{w} \in \{0, 1\}^{d_N(\Pi)}$ by flipping its $i$-th bit. Then, the KL-divergence of the distribution of the data history $\Hcal_T$ under the two models $P_{\bb{w},\Delta}$ and $P_{\bb{w}^{(i)},\Delta}$ satisfy
    \begin{equation*}
        d_{\mathrm{KL}}(P_{\bb{w},\Delta} \Vert P_{\bb{w}^{(i)},\Delta}) \leq \frac{T\Delta^2}{2d_N(\Pi)}
    \end{equation*}
\end{lemma}

\noindent With the choice $\Delta=\sqrt{d_N(\Pi)/T}$, the sum of the error probabilities of any policy $\hat{\pi}_{\Acal,T}$ under the two neighboring models $P_{\bb{w},\Delta}$ and $P_{\bb{w}^{(i)},\Delta}$ can be bounded below by a positive constant. Summing these error probabilities over all binary vectors $\bb{w}\in\{0,1\}^{d_N(\Pi)}$ then yields a corresponding lower bound on the average error probability over the entire collection of models. Consequently, there exists at least one binary vector $\bb{w}$ for which the policy $\hat{\pi}_{\Acal,T}$ incurs a constant-order error probability under $P_{\bb{w},\Delta}$. Since each incorrect decision incurs a regret of order $\Delta$, the expected regret under this model is therefore of order $\Omega(\Delta)$, yielding the desired lower bound of order $\sqrt{d_N(\Pi)/T}$. The detailed proof of Theorem~\ref{thm:regret-lower-bound} is provided in Appendix~\ref{appendix:proof-regret-lower}.

\subsection{Regret Upper Bound}\label{sec:regret-upper-bound}

In this section, we show that the regret upper bound of the proposed strategy matches the lower bound established in Theorem~\ref{thm:regret-lower-bound}, with order
\begin{equation}
\dfrac{\sqrt{d_N(\Pi)}}{\sqrt{T}}
\sqrt{\E_{X}\left[
\sum_{a \in [K]} (\theta^a(X) - \bar{\theta}(X))^2
+ \trace{\bb{\Sigma}_A(X)}
\sum_{(a, b) \in [K]_2} v^{ab}(X)
\right]},
\label{eqn:upper-bound-order}
\end{equation}
\noindent up to logarithmic factors. Instead of using the Natarajan dimension $d_N(\Pi)$, we can also characterize the complexity of the policy class using the entropy integral, which provides an equivalent measure of complexity to the Natarajan dimension $d_N(\Pi)$.
To this end, we first introduce the entropy integral below.

\begin{definition}\label{def:entropy-integral}
    Let $\chi$ be the feature domain, $\Pi$ be any policy class and let $\chi_n \subseteq \chi$ be a discrete subset with $\vert \chi_n\vert = n$. We define the quantities:
    \begin{enumerate}
        \item Given two policies $\pi_1$ and $\pi_2$ in $\Pi$, their Hamming distance over $\chi_n$  is defined as the number of points in $\chi_n$ at which the two policies differ, i.e., 
        \begin{equation*}
            d_H(\pi_1, \pi_2) 
            = \frac{1}{\vert \chi_n\vert}\sum_{x \in \chi_n} \ind{\{ \pi_1(x) \neq \pi_2(x) \}}
        \end{equation*}
        \item For any $\eta > 0$, define $N_H(\eta, \Pi, \chi_n)$ as the $\eta$-covering number of $\Pi$ with respect to the Hamming distance over $\chi_n$. Specifically, $N_H(\eta, \Pi, \chi_n)$ is the smallest number $m$ such that there exists policies $\{ \pi_1, \dots, \pi_m\}$ satisfying the following: for each policy $\pi \in \Pi$, there exists $j \in \{1, \dots, m\}$ such that $d_H(\pi_j,\pi) < \eta$.
        \item For any $\eta > 0$, define $N_H(\eta, \Pi)$ as the supremum value of $N_H(\eta, \Pi, \chi_n)$ over discrete subsets $\chi_n$ in $\chi$ with arbitrary size.
        \item The entropy integral of a policy class $\Pi$ is defined as 
        \begin{equation}
            \kappa(\Pi) = \int_0^1 \sqrt{\log (N_H(\eta^2, \Pi))}d\eta.
            \label{eqn:entropy-integral}
        \end{equation}
    \end{enumerate}
\end{definition}
\noindent Both the entropy integral $\kappa(\Pi)$ and the Natarajan dimension $d_N(\Pi)$ quantify the complexities of a policy class $\Pi$, and these two complexity measures are closely related. In particular, it is known that $\kappa(\Pi) \leq C \sqrt{\log(d) d_N(\Pi)},$ where $d$ denotes the dimension of the covariate space and $C$ is a universal constant~\citep{jin2023upper, kato2024adaptive}.

We next present our main result on the upper bound of the expected regret for the proposed policy with the adaptive sampling strategy.

\begin{theorem}\label{thm:regret-upper-bound}
    Suppose Assumptions~\ref{assum:unconfound}-\ref{assum:estimate-consistency} hold. Additionally, suppose that the policy class $\Pi$ satisfies the inequality $N_H(\eta, \Pi) < C e^{D\eta^{-w}}$ for any $\eta \in (0, 1)$, where $C, D$ and $w \in (0, 0.5)$ are some constants independent of $\eta$. Then, for all $t \geq 1$, the expected simple regret of the recommended policy $\widehat{\pi}_T$ as given by Eq.~\eqref{eqn:recommender-policy} satisfies
    \begin{equation}
        \E(\sqrt{T}R_T(\hat{\pi}_T; P))
        \leq 108.8(\kappa(\Pi) + 8)\sqrt{V^\ast} + C_1\left(192\sqrt{2}\kappa(\Pi) +  \sqrt{\frac{\pi}{2}}\right) + o(1),
        \label{eqn:expected-regret-upper-bound}
    \end{equation}
    \noindent where
    \begin{align}
        C_1 & = 2K\left[ \sqrt{K} \log\left( \frac{1}{\delta} - 1 \right) + \frac{(K-1)(\delta^K + (1-\delta)^K)^2}{4\sqrt{2}\delta^{K+2}(1-\delta)^{K}} \right],\label{eqn:C1-bound-defn} \\
        V^\ast & := \E_{X}\left[ \sum_{a \in [K]} (\theta^{a,\ast}(X) - \bar{\theta}^\ast(X))^2 + \trace{\bb{\Sigma}_A^\ast(X)} \sum_{(a,b) \in [K]_2} v^{ab}(X) \right], \label{eqn:V-star-defn}
    \end{align}
    \noindent $\bar{\theta}^\ast(X) = K^{-1}\sum_{a = 1}^{K} \theta^{a,\ast}(X)$,
    $\bb{\theta}^\ast(X)$ is the ``best fitting'' parameter obtained as solution to Eq.~\eqref{eqn:theta-ast-defn}, and $\bb{\Sigma}^\ast_A(X)$ is as in Eq.~\eqref{eqn:sigma-A-defn} with $\bb{\theta}(X)$ substituted by $\bb{\theta}^\ast(X)$.
\end{theorem}

Before moving on to the proof of Theorem~\ref{thm:regret-upper-bound}, we indicate a few examples of the policy class for which the log-covering number satisfies the bound $\log N_H(\eta, \Pi) \lesssim D\eta^{-w}$. By Haussler's theorem~\citep{haussler1995sphere}, any policy class with a bounded VC-dimension satisfies this restriction, e.g., linear policies, bounded-depth decision trees~\citep{athey2021policy}, fixed-architecture neural networks~\citep{jin2023upper}, etc. In addition, certain nonparametric estimators using $\beta$-H\"{o}lder smooth classes also satisfy this bound on the covering number.

We next provide the proof sketch of Theorem~\ref{thm:regret-upper-bound}.

\vspace*{1em}
\noindent\textbf{Step 1: Construction of proxy value function.} The first step in the proof of Theorem~\ref{thm:regret-upper-bound} is to consider a proxy value function 
\begin{equation}
    Q_T^\ast(\pi) = \dfrac{1}{T}\sum_{t=1}^T \sum_{a \in [K]} \bb{1}_{\{ \pi(X_t) = a \}} \Gamma_t^{a,\ast},
    \label{eqn:Q-function-proxy}
\end{equation}
\noindent where $\Gamma_t^{1,\ast} = 0$ and $\bb{\Gamma}_t^\ast = (\Gamma_t^{2,\ast}, \dots, \Gamma_t^{K,\ast})\tr$ is expressed as
\begin{equation*}
    \bb{\Gamma}_t^\ast = \bb{\theta}^\ast(X_t) - [\nabla_{\bb{\theta}} \bb{U}(\bb{\theta}^\ast, \bb{m}, X_t) ]^{-1} [\nabla_{\bb{m}} \bb{U}(X_t) ] \bb{e}_t^\ast(X_t),
\end{equation*}
\noindent and,
\begin{equation*}
    \bb{e}_t^\ast(X) = \text{vec}\left( \left\{\dfrac{\bb{1}_t^{ab}}{w^{ab}(X_t)} (Y^{ab}_t - m^{ab}(X_t)) \right\}_{a \in [K] \setminus \{1\}, b \in [K] \setminus \{a\} } \right).
\end{equation*}
\noindent Here, the indicator variable $\bb{1}_t^{ab}$ is a shorthand for $\ind{\{(a_t,b_t)\in \{ (a,b),(b,a)\}\}}$, under the sampling scheme based on the oracle weights $w^{ab}(X_t)$.

\vspace*{1em}
\noindent\textbf{Step 2: Decomposition of the regret.} Next, we decompose the regret upper bound into two parts: The first part captures the additional regret arising from the estimation process by quantifying the discrepancy between $\widehat{Q}_T(\pi)$ and $Q_T^\ast(\pi)$. The second part captures the regret between the proxy value function with oracle information $Q_T^\ast(\pi)$ and the population value function $Q(P;\pi)$.

To this end, we first introduce some notations
\begin{align*}
    \widehat{\Delta}_T(\pi_1, \pi_2) & := \widehat{Q}_T(\pi_1) - \widehat{Q}_T(\pi_2),\\
    \Delta^\ast_T(\pi_1, \pi_2) & := Q^\ast_T(\pi_1) - Q^\ast_T(\pi_2),\\
    \Delta(\pi_1, \pi_2) & := Q(\pi_1; P) - Q(\pi_2; P),
\end{align*}
\noindent where the definitions of various value functions are introduced in Eq.~\eqref{eqn:empirical-Q}, \eqref{eqn:Q-function-proxy} and \eqref{eqn:Q-func-popn}, respectively. In addition, we let $\pi_0 := \pi^\ast(P) = \argmax_{\pi \in \Pi} Q(\pi; P)$. It holds that
\begin{align}
    R_T(P; \widehat{\pi}_T)
    & = Q(\pi_0; P) - Q(\widehat{\pi}_T; P)\nonumber\\
    & = \Delta(\pi_0, \widehat{\pi}_T)\nonumber\\
    & = \widehat{Q}_T(\pi_0) - \widehat{Q}_T(\widehat{\pi}_T) + \Delta(\pi_0, \widehat{\pi}_T) - \widehat{\Delta}_T(\pi_0, \widehat{\pi}_T)\nonumber\\
    & \leq \sup_{\pi_1, \pi_2 \in \Pi} \vert \Delta(\pi_1, \pi_2) - \widehat{\Delta}_T(\pi_1, \pi_2) \vert, \ \text{ as } \widehat{\pi}_T \text{ maximizes } \widehat{Q}_T(\pi), \nonumber \\
    & \leq \sup_{\pi_1, \pi_2 \in \Pi} \vert \Delta(\pi_1, \pi_2) - {\Delta}_T^\ast(\pi_1, \pi_2) \vert + \sup_{\pi_1, \pi_2 \in \Pi} \vert \Delta_T^\ast(\pi_1, \pi_2) - \widehat{\Delta}_T(\pi_1, \pi_2) \vert \nonumber \\
    & \leq \sup_{\pi_1, \pi_2 \in \Pi} \left\vert \Delta(\pi_1, \pi_2) - {\Delta}^\ast_T(\pi_1, \pi_2) \right\vert + \sup_{\pi_1, \pi_2 \in \Pi} \left\vert D_T(\pi_1) - D_T(\pi_2) \right\vert,
    \label{eqn:risk-bound-decomp}
\end{align}
\noindent where $D_T(\pi) = \widehat{Q}_T(\pi) - Q_T^\ast(\pi)$. Next, we control the two terms in Eq.~\eqref{eqn:risk-bound-decomp} in Steps 3 and 4, respectively.

\vspace*{1em}
\noindent\textbf{Step 3: Bounding the difference in oracle policy.} Note that, by construction, the proxy value function $Q_T^\ast(\pi)$ is based on a sequence of i.i.d. observations. Consequently, the first term can be expressed as the difference between two empirical processes indexed by the policies $\pi_1$ and $\pi_2$. Uniform convergence of these empirical processes can then be established using standard concentration inequalities, with the complexity of the policy class $\Pi$ characterized by its Rademacher complexity. We refer interested readers to Appendix~\ref{appendix:policy-diff-Rademacher} for further details. This argument leads to the following lemma, which provides a bound on the expectation of the first term in Eq.~\eqref{eqn:risk-bound-decomp}.

\begin{lemma}\label{lem:Q-function-diff-step1}
    Under Assumptions~\ref{assum:unconfound}-\ref{assum:positivity}, 
    \begin{equation*}
        \E\left[ \sup_{\pi_1, \pi_2 \in \Pi} \vert \Delta(\pi_1, \pi_2) - {\Delta}^\ast_T(\pi_1,\pi_2) \vert \right] \leq \dfrac{108.8 (\kappa(\Pi) + 8)}{\sqrt{T}} \sqrt{V^\ast} + \sqrt{\frac{\pi}{2T}}C_1 + o(T^{-1/2}).
    \end{equation*}
    \noindent Here $V^\ast$ is as defined in Eq.~\eqref{eqn:V-star-defn}, $C_1$ is as in Eq.~\eqref{eqn:C1-bound-defn}.
\end{lemma}
\noindent Detailed proof of Lemma~\ref{lem:Q-function-diff-step1} is provided in Appendix~\ref{appendix:proof-lem-Q-function-diff-step1}.

\vspace*{1em}
\noindent \textbf{Step 4: Bounding the difference in proxy and empirical value functions.} In order to obtain an upper bound of the second term in the decomposition~\eqref{eqn:risk-bound-decomp}, we establish uniform convergence between the empirical value function $\widehat{Q}_T(\pi)$ and the proxy value function $Q_T^\ast(\pi)$. A standard approach is to exploit the martingale structure of their difference and apply martingale concentration inequalities. To this end, we define the filtration $\{ \Fcal_t \}_{t = 1}^T$, where $\Fcal_t$ denotes the sigma-algebra generated from $\Hcal_{t-1}$ and the covariate $X_t$. By construction, $\widehat{m}^{ab}(X_t), (a,b)\in [K]_2$ is $\Fcal_t$ measurable, and $\widehat{\theta}^a_t(X_t), a\in [K]$ is also $\Fcal_t$ measurable. 

Therefore, it follows that 
\begin{equation}
    \bb{Z}_{1t} = \bbhat{\Gamma}_t - \E(\bbhat{\Gamma}_t \mid \Fcal_t)
    \label{eqn:Z1t-defn}
\end{equation}
\noindent is a martingale difference sequence with respect to the filtration $\{\Fcal_t\})_{t \ge 1}$ and 
\begin{equation}
    \bb{Z}_{2t} = \bb{\Gamma}_t^\ast - \E(\bb{\Gamma}_t^\ast \mid \Fcal_t)
    \label{eqn:Z2t-defn}
\end{equation}
\noindent is an i.i.d. mean-zero random variable. The difference between the conditional expectations $\E(\bbhat{\Gamma}_t \mid \Fcal_t)$ and $\E(\bb{\Gamma}_t^\ast \mid \Fcal_t)$ can be controlled using the consistency results established under Assumption~\ref{assum:estimate-consistency} and Lemma~\ref{lem:consistency-bounds}. We apply these arguments over the policy difference of value function errors, i.e., on $D_T(\pi_1) - D_T(\pi_2)$. As a result, we obtain the following result.

\begin{lemma}\label{lem:value-function-diff}
    Let $\Pi$ be any policy class. Assume that Assumptions~\ref{assum:unconfound}-\ref{assum:estimate-consistency} hold. Then, for any $T \geq 1$, the difference between the value function errors $D_T(\pi_1) - D_T(\pi_2)$ satisfies 
    \begin{equation*}
        \E\left[ \sqrt{T} \sup_{\pi_1, \pi_2 \in \Pi} \left\vert D_T(\pi_1) - D_T(\pi_2) \right\vert \right]
        \leq 192\sqrt{2}C_1 \kappa(\Pi) + o(1).
    \end{equation*}
    \noindent Here, $D_T(\pi) = \widehat{Q}_T(\pi) - Q_T^\ast(\pi)$.
\end{lemma}

\noindent The proof of Lemma~\ref{lem:value-function-diff} is outlined in Appendix~\ref{appendix:proof-value-function-diff}. Equipped with Lemmas~\ref{lem:Q-function-diff-step1} and~\ref{lem:value-function-diff}, one applies the triangle inequality to the decomposition~\eqref{eqn:risk-bound-decomp}, which recovers Theorem~\ref{thm:regret-upper-bound}. The technical details are provided in Appendix~\ref{appendix:proof-regret-ub}.

Eq.~\eqref{eqn:expected-regret-upper-bound} may be compared with the lower bound obtained in Theorem~\ref{thm:regret-lower-bound}, which matches up to the dominating term. Since the entropy integral $\kappa(\Pi)$ is bounded above by the Natarajan dimension $d_N(\Pi)$, asymptotic equivalence between the lower and upper bounds of the regret establishes the optimality of the proposed strategy.

\section{Simulation Studies}\label{sec:simulation}

\subsection{Simulation Settings}\label{sec:sim-settings}

We now illustrate the benefits of the proposed adaptive sampling procedure through a series of numerical experiments. We consider four scenarios that represent varying degrees of model complexity and misspecification. Specifically, the first three scenarios are generated under the BTL model, whereas the fourth scenario is generated under a misspecified BTL model.

\begin{enumerate}[label=(S\arabic*)]
    \item\label{sim:discrete} In the first setting, we assume the context $X_t$ follows a discrete uniform distribution on $[K]$. We define the value of arm $a \in [K]$ as $\theta^{a,\ast}(x) = (x-1)^2 -(x - a)^2$. Note that, $\theta^1(x) = 0$ and $\theta^{a,\ast}(x)$ is maximized at $a = x$. In our experiments, we let $K = 5$.
    \item\label{sim:li_and_li_correct_discrete} This setting is inspired by Setting II of~\cite{li2025efficient}. We set $K = 5$, and generate the context $\bb{X} = (U, V)\tr$ with the first coordinate $U$ following a uniform distribution over $\{ -3, -2, \dots, 2, 3\}$ and the second coordinate $V$ independently following a Bernoulli distribution with success probability $1/2$. The values of the arms are defined as
    \begin{align*}
        \theta^{1,\ast}((u, v)) & = 0, \
        \theta^{2,\ast}((u, v)) = uv, \
        \theta^{3,\ast}((u, v)) = u^2 + v,\\
        \theta^{4,\ast}((u, v)) & = -0.5u + v,\
        \theta^{5,\ast}((u, v)) = 0.5u + v,
    \end{align*}
    \noindent for any $u \in \R$ and $v \in \{ 0, 1\}$.
    \item\label{sim:li_and_li_miss1_discrete} This setup introduces nonlinearity in the value function for the fifth arm. Specifically, this setup is the same as Setup~\ref{sim:li_and_li_correct_discrete}, except that the value function of the fifth arm is replaced by
    \begin{equation*}
        \theta^{5,\ast}((u, v)) = \sin(1.5(u + 0.5v)).
    \end{equation*}
    \item\label{sim:li_and_li_miss2_discrete} This setup represents a scenario in which the Bradley--Terry model is misspecified, similar to Setting I of~\cite{li2025efficient}. Specifically, we let $K = 3$, and use the same context distribution as in Setup~\ref{sim:li_and_li_correct_discrete}, and define the pairwise comparison probabilities as
    \begin{align*}
        m^{1,2}((u, v)) & = 0.5 + 0.2 \sin(1.5(u + v)),\\
        m^{1,3}((u, v)) & = \sigma(0.3 u(v - 1)),\\
        m^{2,3}((u, v)) & = \sigma(0.2u^2 - 0.5),
    \end{align*}
    \noindent where $\sigma(\cdot)$ is the sigmoid function.
\end{enumerate}

In addition to these settings, we also consider their continuous-context variants (CS2), (CS3) and (CS4), analogous to \ref{sim:li_and_li_correct_discrete}, \ref{sim:li_and_li_miss1_discrete} and~\ref{sim:li_and_li_miss2_discrete} respectively. Here we assume the context $X = (U, V)$ is generated such that $U$ follows a normal distribution with mean $0$ and variance $0.5$, and $V$, independently of $U$, follows a Bernoulli distribution with success probability equal to $1/2$. 

\subsection{Inference and Sampling methods}\label{sec:sim-sampling}

For each scenario described above, we simulate $100$ independent trajectories, each consisting of $T = 1000$ rounds. We consider two approaches for estimating the latent preferences $\theta(x)$s. 

\begin{enumerate}
    \item The first approach follows our proposed estimation procedure. Specifically, for each pair $(a, b)$, we fit a logistic regression model to obtain an estimate of $\widehat{m}_{ab}(x)$, and then solve Eq.~\eqref{eqn:estimating-eqn} to estimate the latent $\theta$'s. 
    \item In the second approach, we parametrize $\theta^a(x) = \bb{w}_{a}\tr f(x)$, where $f(x)$ is a suitable transformation of the covariate and $\bb{w}\tr = (\bb{w}_1\tr, \dots, \bb{w}_K\tr)$ is the unknown parameter. Then, we fit a joint Bradley-Terry model directly to the observed pairwise comparison data by maximizing the following log-likelihood
    \begin{equation}
        \ell(\bb{w}) = \sum_{t=1}^T \left[ Y_t \log(\sigma( (\bb{w}_{a_t} - \bb{w}_{b_t})\tr f(X_t) )) + (1-Y_t) \log\left( 1 - \sigma( (\bb{w}_{a_t} - \bb{w}_{b_t})\tr f(X_t)) \right) \right],
        \label{eqn:log-likelihood-bt-model}
    \end{equation}
    \noindent subject to an $L_2$-regularizer. Given the estimated latent scores, we obtain the pairwise comparison probabilities using the following plug-in estimator
    \begin{equation*}
        \widehat{m}^{ab}(x) = \sigma(\widehat{\theta}^a(x) - \widehat{\theta}^b(x)) = \sigma((\bbhat{w}_a - \bbhat{w}_b)\tr f(x)).
    \end{equation*}
    \noindent By construction, the resulting pairwise comparison probabilities satisfy the estimating equations in Eq.~\eqref{eqn:estimating-eqn}.
\end{enumerate}

We compare the proposed adaptive sampling strategy with several existing benchmark methods. The benchmark sampling methods include \emph{uniform sampling}, which selects each pair of arms with equal probability, maximum entropy sampling(\emph{MaxEnt})~\citep{elwood2023maximum}, maximum informative pair sampling (\emph{MaxIntP})~\citep{saha2021optimal}, and Double Thompson Sampling (\emph{DTS})~\citep{li2024feel}. To ensure a fair comparison across sampling strategies, all methods use a common uniform warm-up phase during the first $100$ rounds to obtain stable initial estimates of the latent preference scores. For the remaining $900$ rounds, each sampling strategy is used to select pairs of arms and collect the corresponding preference data. At the end of $T = 1000$ rounds, we construct the recommendation policy using different procedures for discrete and continuous contexts, as illustrated in Eq.~\eqref{eqn:best-policy-kronecker}-\eqref{eqn:best-policy-linear}. All the relevant source code has been published as a \href{https://github.com/subroy13/active-policy-learning}{GitHub repository}\footnote{\url{https://github.com/subroy13/active-policy-learning}} to ensure reproducibility and broader reuse.

\subsection{Results}\label{sec:sim-results}

\begin{figure}[htbp]
    \centering
    \includegraphics[width=0.7\linewidth]{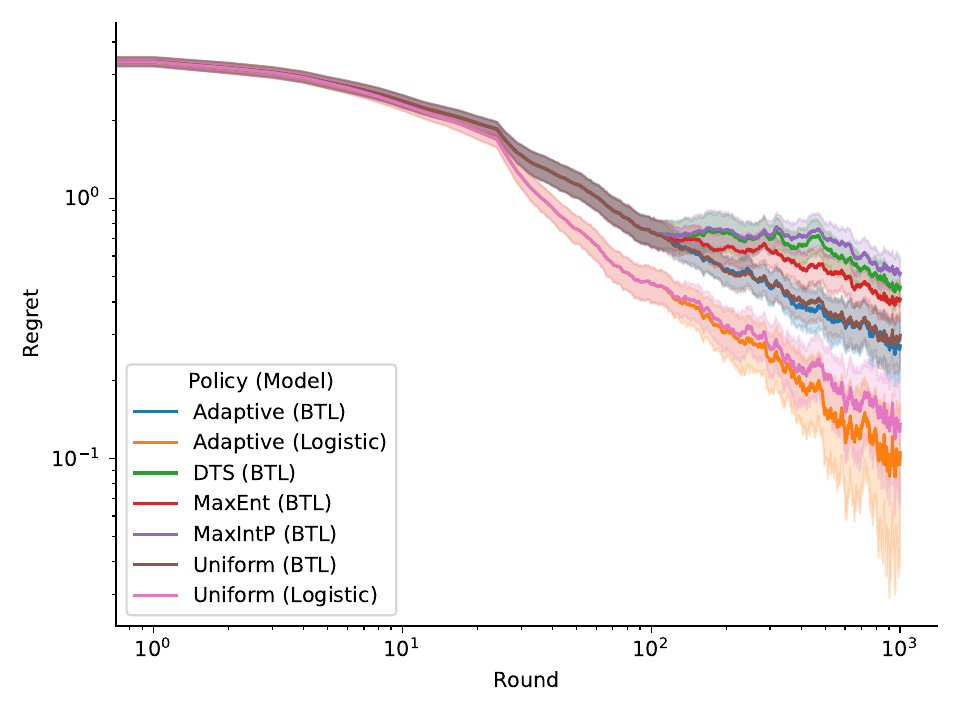}
    \caption{The regret curve for various sampling schemes under the active learning framework for scenario~\ref{sim:discrete}. Both axes are on a logarithmic scale.}
    \label{fig:discrete}
\end{figure}

\begin{figure}[htbp]
    \centering
    \begin{subfigure}[t]{0.48\textwidth}
        \centering
        \includegraphics[width=\linewidth]{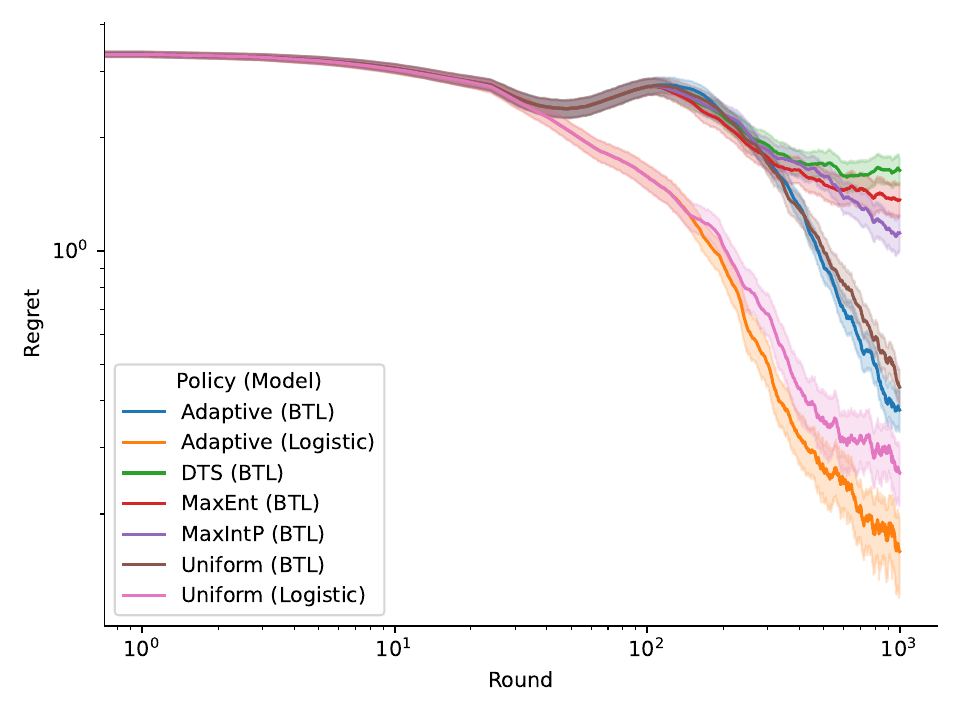}
        \caption{}
    \end{subfigure}
    \begin{subfigure}[t]{0.48\textwidth}
        \centering
        \includegraphics[width=\linewidth]{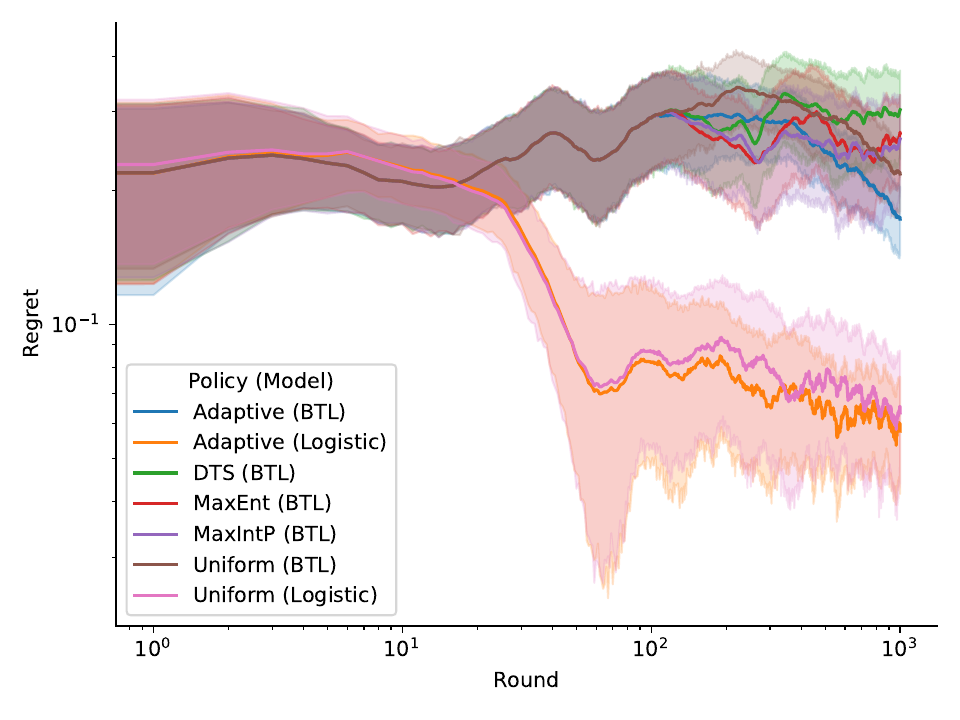}
        \caption{}
    \end{subfigure}
    \caption{The regret curve for various sampling schemes under the active learning framework for scenario~\ref{sim:li_and_li_correct_discrete} (Subfig. a) and (CS2) (Subfig. b). Both axes are on a logarithmic scale.}
    \label{fig:li_li_correct}
\end{figure}

\begin{figure}[htbp]
    \centering
    \begin{subfigure}[t]{0.48\textwidth}
        \centering
        \includegraphics[width=\linewidth]{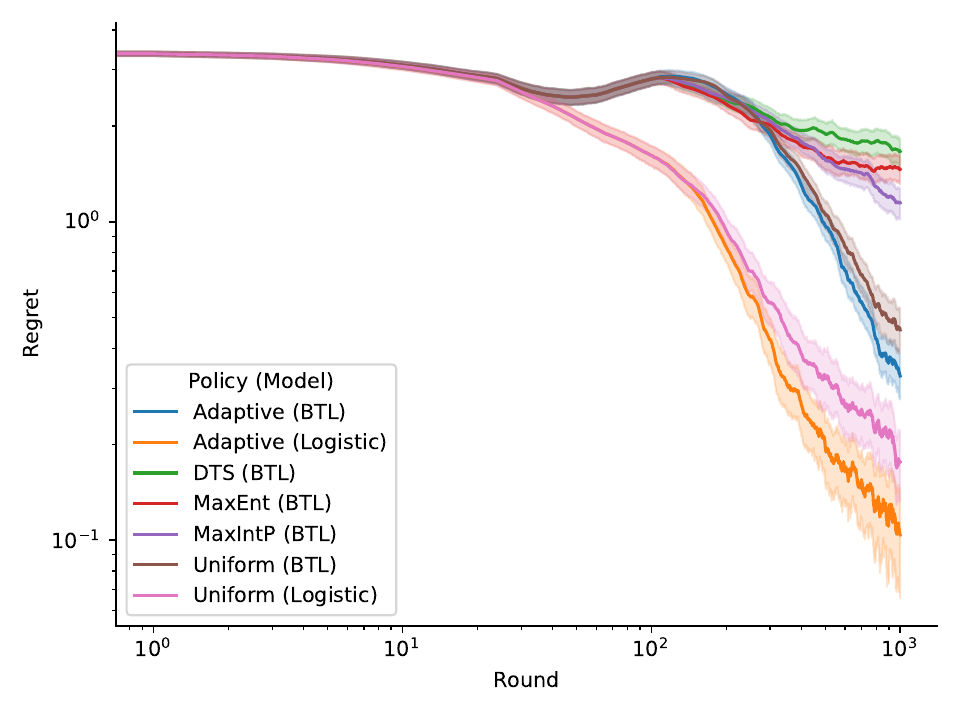}
        \caption{}
    \end{subfigure}
    \begin{subfigure}[t]{0.48\textwidth}
        \centering
        \includegraphics[width=\linewidth]{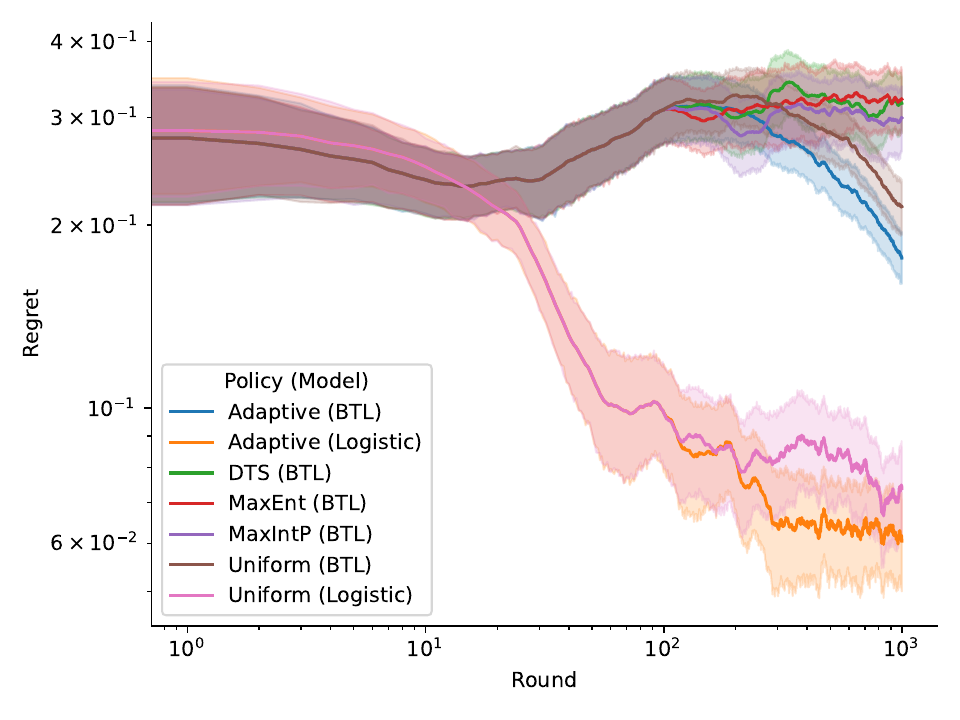}
        \caption{}
    \end{subfigure}
    \caption{The regret curve for various sampling schemes under the active learning framework for scenario~\ref{sim:li_and_li_miss1_discrete} (Subfig. a) and (CS3) (Subfig. b). Both axes are on a logarithmic scale.}
    \label{fig:li_li_miss1}
\end{figure}

\begin{figure}[htbp]
    \centering
    \begin{subfigure}[t]{0.48\textwidth}
        \centering
        \includegraphics[width=\linewidth]{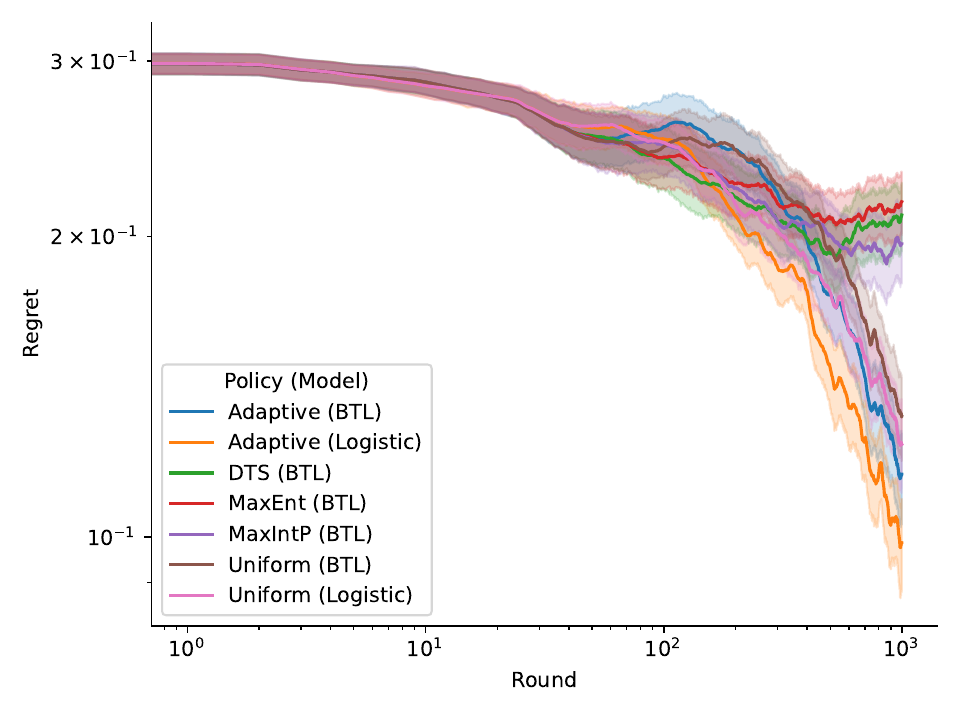}
        \caption{}
    \end{subfigure}
    \begin{subfigure}[t]{0.48\textwidth}
        \centering
        \includegraphics[width=\linewidth]{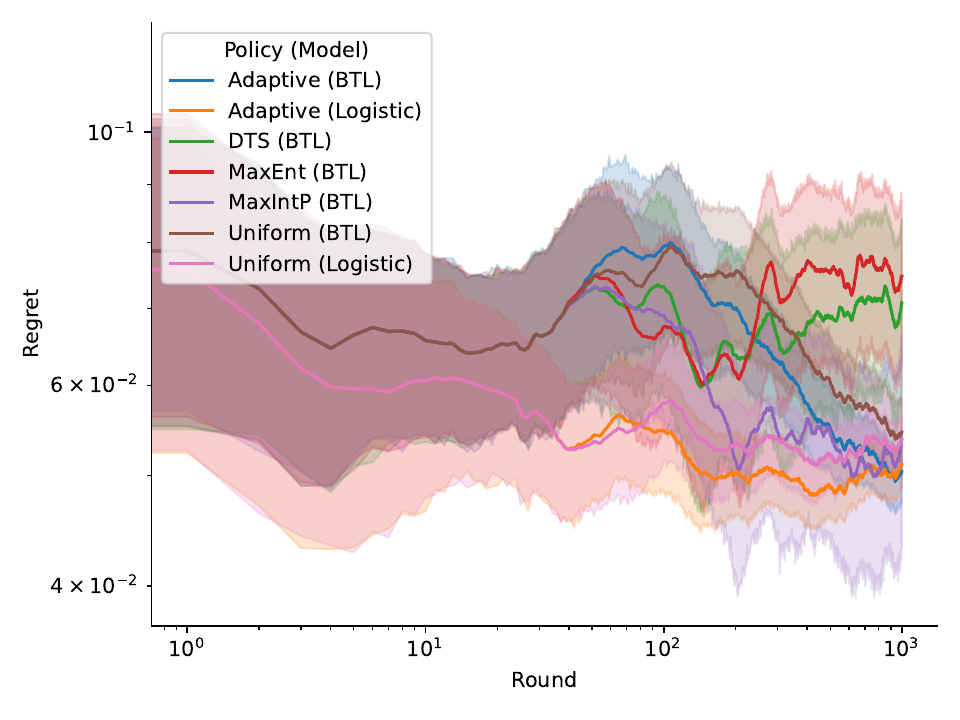}
        \caption{}
    \end{subfigure}
    \caption{The regret curve for various sampling schemes under the active learning framework for scenario~\ref{sim:li_and_li_miss2_discrete} (Subfig. a) and (CS4) (Subfig. b). Both axes are on a logarithmic scale.}
    \label{fig:li_li_miss2}
\end{figure}

Figures~\ref{fig:discrete}--\ref{fig:li_li_miss2} plot the regret trajectories, against the number of rounds $T$, for scenarios~\ref{sim:discrete}--\ref{sim:li_and_li_miss2_discrete} and their continuous variants (CS2)--(CS4), under the two inference methods (pairwise logistic regression and joint Bradley--Terry MLE) and the five sampling schemes described in Section~\ref{sec:sim-sampling}. In each panel, all policy--model combinations coincide for approximately the first $4K^2$ rounds ($K$ is the number of arms), reflecting the shared uniform warm-up phase used to stabilize the initial estimates; the curves then separate as each sampling rule begins to adapt to its own accumulated information.

Several conclusions emerge from the simulation results. First, paired with the same underlying estimator, the proposed adaptive sampling strategy often performs substantially better than uniform sampling. The improvement is most pronounced when adaptive sampling is combined with the pairwise logistic regression estimator, achieving a substantial reduction in regret in all scenarios relative to uniform sampling. This is consistent with our theoretical results, namely, under correct specification, the logistic estimator converges at the parametric rate, enabling accurate estimation of the adaptive weights~\eqref{eqn:weights-emp-defn} early in the sampling process. When adaptive sampling is instead paired with the joint Bradley--Terry--Luce MLE, it continues to outperform uniform sampling, although the improvement is more modest in the discrete, exact-policy settings (Figures~\ref{fig:discrete} and~\ref{fig:li_li_correct}a, \ref{fig:li_li_miss1}a) and considerably larger in the continuous-context settings, where the recommendation policy is restricted to the linear class (Figures~\ref{fig:li_li_correct}b, \ref{fig:li_li_miss1}b).

Second, the existing informativeness-based sampling schemes, namely \emph{DTS}, \emph{MaxEnt}, and \emph{MaxIntP}, consistently underperform in the discrete settings when implemented under the Bradley-Terry--Luce model. They sometimes perform worse than the uniform sampling; see Figures~\ref{fig:discrete}, \ref{fig:li_li_correct}a, \ref{fig:li_li_miss1}a. This suggests that heuristics designed for cumulative-regret or non-contextual bandit objectives do not necessarily translate into strong fixed-budget performance for contextual best-arm identification from pairwise comparisons. This also underscores the benefit of the sampling rule derived from the estimation-variance objective in Eq.~\eqref{eqn:weights-emp-defn}.

Finally, comparisons across different degrees of model mis-specification illustrate how the relative performance of the estimators changes as misspecification becomes more severe. Under the correctly specified setting~\ref{sim:li_and_li_correct_discrete} and its mildly misspecified variant~\ref{sim:li_and_li_miss1_discrete}, the results based on the Bradley--Terry--Luce MLE and our logistic regression-based estimator remain clearly separated, with the latter retaining a sizable advantage. However, under the more severely misspecified setting~\ref{sim:li_and_li_miss2_discrete} (Figure~\ref{fig:li_li_miss2}), this separation largely disappears with many procedures exhibiting a comparable regret by the end of the horizon ($T = 1000$). In other words, misspecification diminishes the advantage of one estimator over another, but does not eliminate the advantage of adaptive sampling over uniform sampling within a fixed estimator. These results provide some empirical evidence for the robustness of the proposed procedure in Section~\ref{sec:strategy-detail} to model misspecification.

\section{Application to Large Language Models Evaluation}\label{sec:realdata}

Recent advancements in large language models have shown remarkable capabilities across a wide range of artificial intelligence tasks~\citep{zhao2023survey}. As a result, a growing number of models have been developed, differing in architecture, training data, and tokenization schemes, and exhibiting varying capabilities across tasks such as text comprehension and summarization, code generation, puzzle solving, and mathematical reasoning. Alongside these developments, a growing body of literature has emerged on methods for evaluating across different tasks; see~\cite{chang2024survey} for a comprehensive overview. One important class of evaluation methods focuses on ranking a given set of language models based on evaluations of their performance across various tasks by human or LLM judges. These rankings are typically constructed from pairwise comparisons between the responses generated by different language models to the same inputs. In this section, we apply the proposed active learning framework with adaptive sampling to efficiently construct such preference rankings from pairwise comparison data.

Before presenting the detailed analyses, we first note that these datasets are static and were collected offline, whereas our framework assumes adaptive data collection. To adapt to our framework, we randomly split each dataset into training and testing sets with $75\%$ of the samples in the training set. We use the former to simulate sequential data collection and the latter for out-of-sample evaluation. At each round, a context $X_t$ is sampled from the covariates in the training data, after which a pair of models $(a_t,b_t)$ is selected according to the sampling policy. If the training set contains at least one observation corresponding to the selected context-pair combination $(X_t,a_t,b_t)$, the corresponding response $Y_t$ is recorded. For multiple matching observations, one is selected uniformly at random. If no such observation exists, the sampled context is rejected, and a new context is drawn, and this procedure is repeated until the selected model pair has an observed comparison. We evaluate the learned soft-policy $\widehat{\pi}_{\mathrm{soft}}$ using the following alignment accuracy, computed separately on the training and testing sets:
\begin{equation}
    \mathrm{Acc}(\widehat{\pi}_{\mathrm{soft}}) = \frac{1}{n}\sum_{i=1}^n \left[ Y^i \ind{\{\widehat{\pi}_{\mathrm{soft}}(a^i \mid X^i) > \widehat{\pi}_{\mathrm{soft}}(b^i \mid X^i)\}} + (1 - Y^i) \ind{\{ \widehat{\pi}_{\mathrm{soft}}(b^i \mid X^i) > \widehat{\pi}_{\mathrm{soft}}(a^i \mid X^i) \} } \right].
    \label{eqn:alignment-accuracy}
\end{equation}
\noindent Here, the superscript $i$ indexes the $i$-th observation in a dataset. This metric measures the proportion of observations for which the observed preference $Y^i$ agrees with the pairwise ranking induced by the learned policy $\widehat{\pi}$ between models $(a^i,b^i)$. The alignment accuracy averages the agreement over the entire training or test dataset of size $n$ on the arm selection probabilities for any learned policy.

\subsection{MT-Bench and Chatbot Arena}\label{sec:realdata-chatbot-arena}

We first demonstrate our method on two benchmark datasets introduced by~\cite{zheng2023judging}: Multi-Turn Benchmark (MT-Bench) and Chatbot Arena. The first dataset, MT-Bench, consists of 80 carefully designed multi-turn questions spanning eight categories, namely tasks involving \texttt{Writing}, \texttt{Roleplay}, \texttt{Reasoning}, \texttt{Knowledge Extraction}, \texttt{Scientific Analysis (STEM)} and questions from \texttt{Mathematics}, \texttt{Coding}, and \texttt{Humanities}. We use the one-hot encoded vector of these categories as the covariates for our study. The responses were generated by six large language models and evaluated through pairwise comparisons. The model responses were evaluated both by human judges and by GPT-4 serving as an LLM judge, yielding $3,355$ and $2,400$ pairwise comparisons, respectively. The train-test split on the MT-Bench dataset produced $n = 3,281$ observations and $1,094$ observations in the training and testing sets, respectively. 

\begin{figure}[ht]
    \centering
    \begin{subfigure}[t]{0.3\textwidth}
        \centering
        \includegraphics[width=\linewidth]{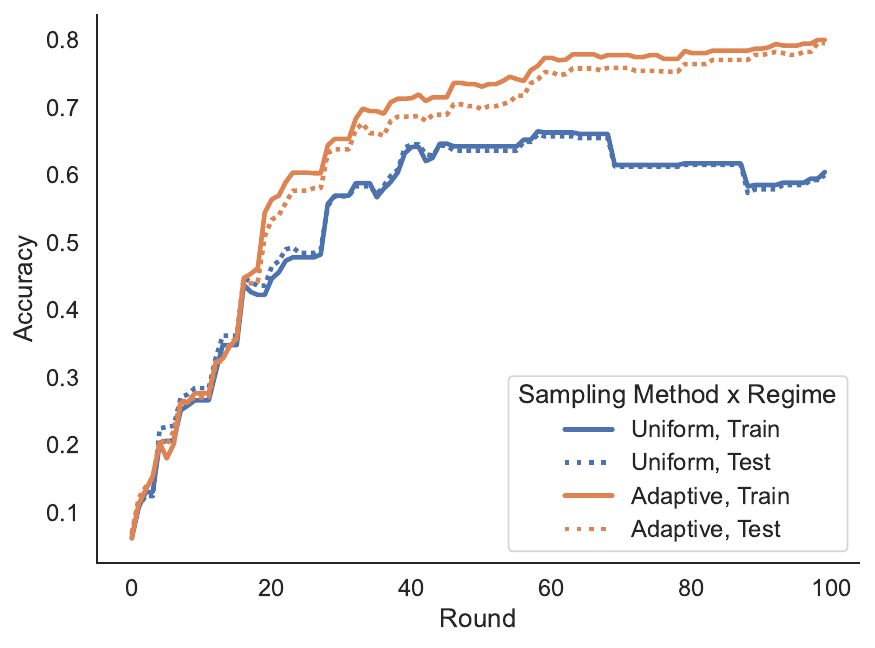}
        \caption{}
    \end{subfigure}
    \begin{subfigure}[t]{0.3\textwidth}
        \centering
        \includegraphics[width=\linewidth]{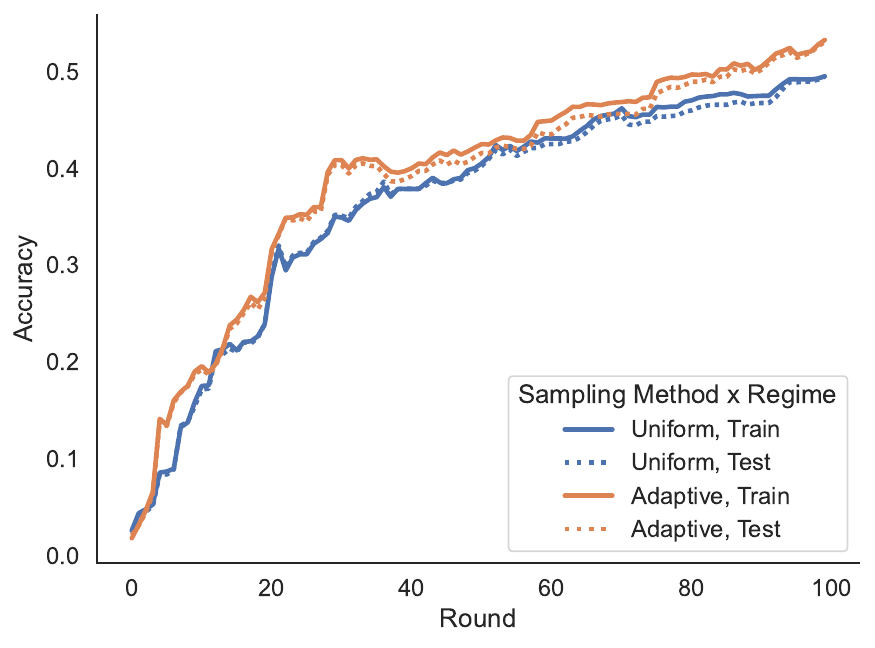}
        \caption{}
    \end{subfigure}
    \begin{subfigure}[t]{0.3\textwidth}
        \centering
        \includegraphics[width=\linewidth]{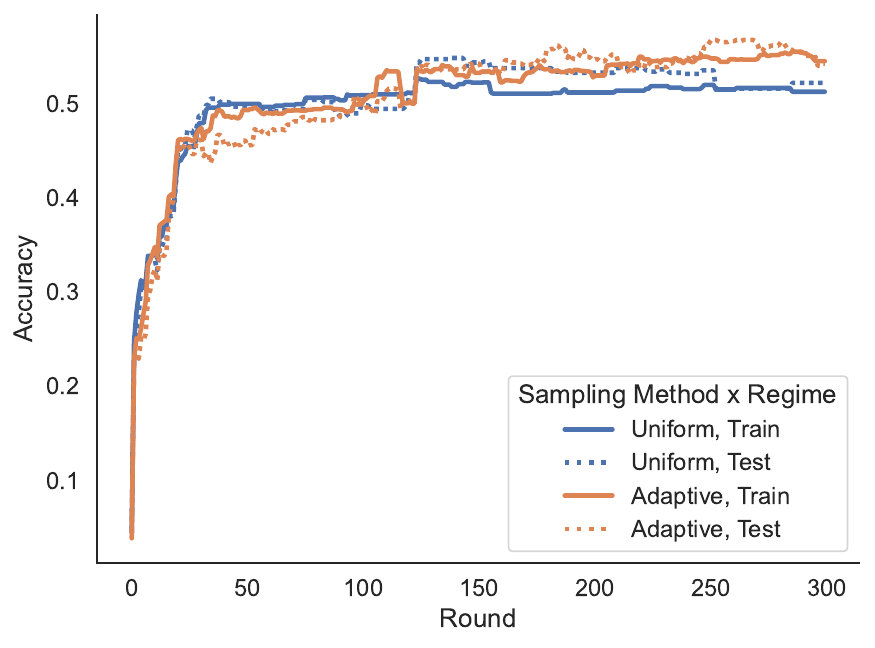}
        \caption{}
    \end{subfigure}
    \caption{The alignment accuracy $\mathrm{Acc}(\widehat{\pi}_t)$ of the recommendation policy for the training and test sets of MT-bench dataset (Subfig a), Chatbot arena dataset (Subfig b), LMArena-140K dataset (Subfig c)  under uniform and adaptive sampling.}
    \label{fig:llm-mt-bench}
\end{figure}

\begin{figure}[ht]
    \centering
    \begin{subfigure}[t]{\textwidth}
        \centering
        \includegraphics[width=\linewidth]{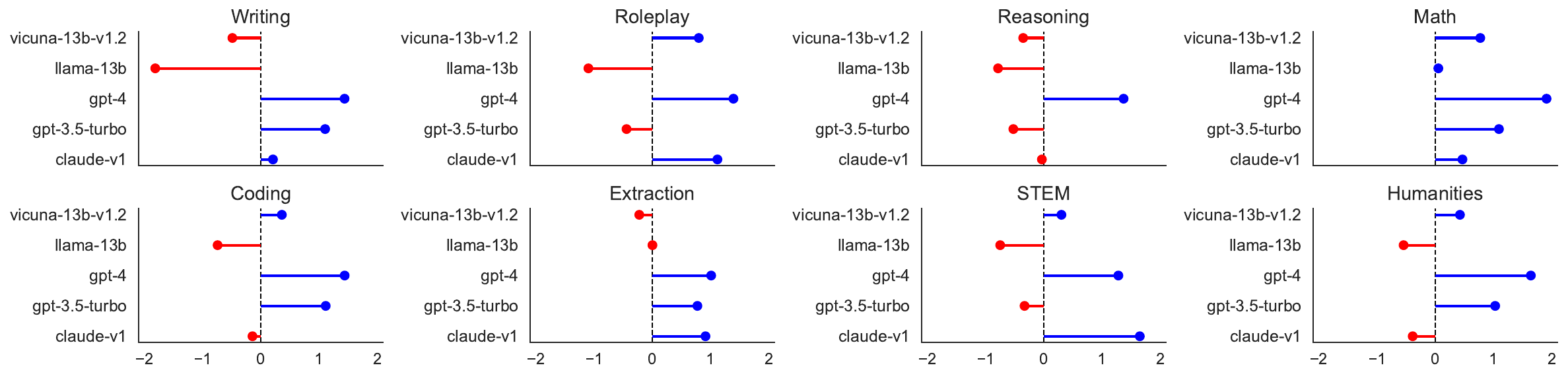}
        \caption{}
    \end{subfigure}
    \begin{subfigure}[t]{\textwidth}
        \centering
        \includegraphics[width=\linewidth]{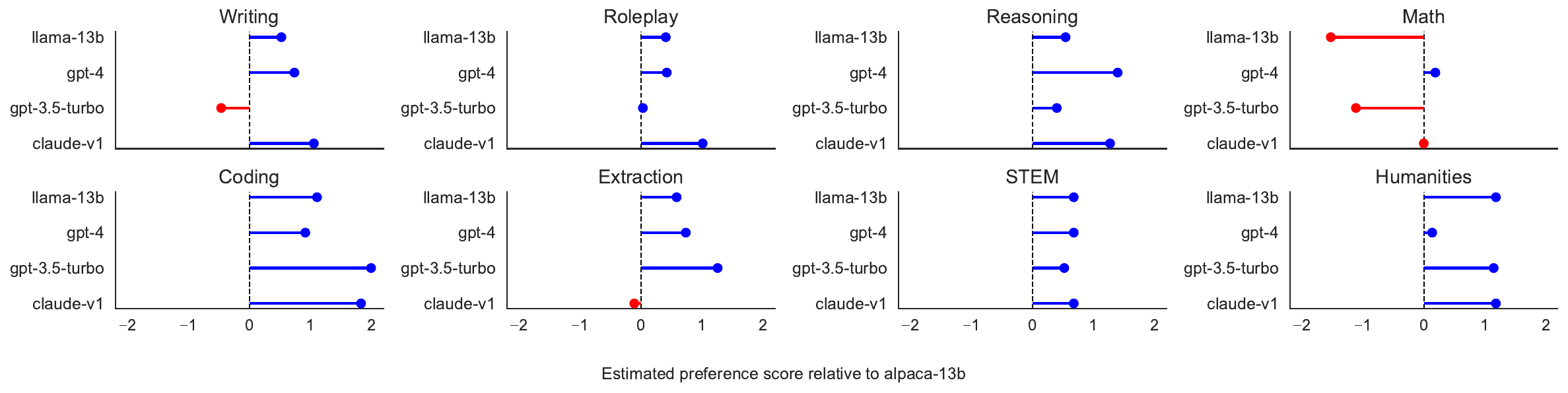}
        \caption{}
    \end{subfigure}
    \begin{subfigure}[t]{\textwidth}
        \centering
        \includegraphics[width=\linewidth]{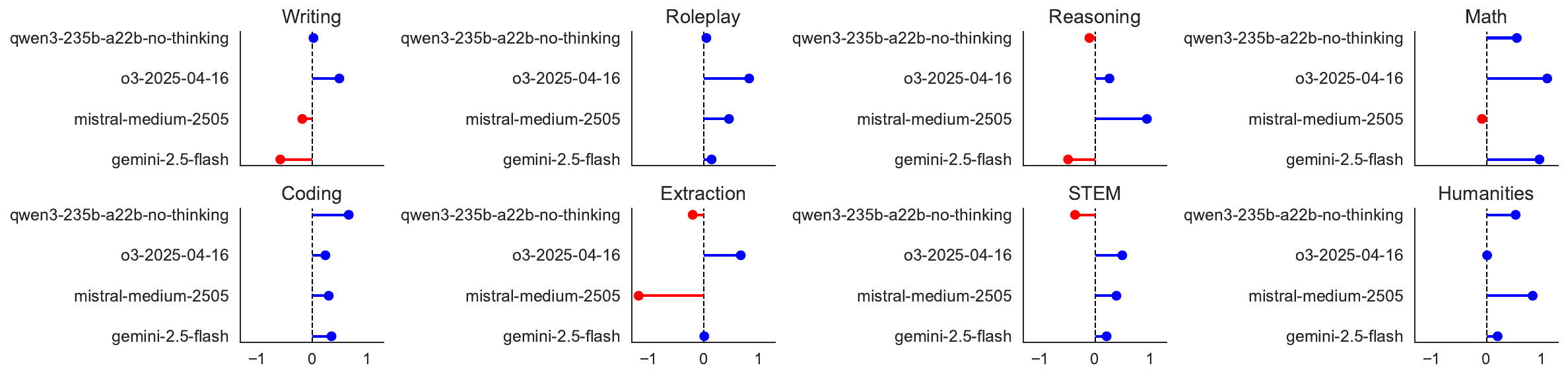}
        \caption{}
    \end{subfigure}
    \caption{Contextwise preference value for each language model for MT-Bench (Subfig. a), Chatbot Arena (Subfig. b), LMArena-140K (Subfig. c) datasets.}
    \label{fig:llm-mt-bench-theta}
\end{figure}

We perform both adaptive sampling and uniform sampling within the active learning framework. Here, we directly estimate ${\theta}^a(x)$ for each $a\in [K]$ using the Bradley-Terry-Luce (BTL) model in Eq.~\eqref{eqn:log-likelihood-bt-model}. This choice reflects the fact that, in realistic settings, some model pairs may have no observed comparisons, making it infeasible to fit a separate pairwise logistic regression to estimate $m^{ab}(x)$ for every pair. For both sampling strategies, we run the procedure for a total of $T=100$ rounds. Let $\widehat{\pi}_t$ be the estimated recommendation policy at the end of round $t$ based on the training samples. The resulting trajectories of the alignment accuracy $\mathrm{Acc}(\widehat{\pi}_t)$ given in~\eqref{eqn:alignment-accuracy}, over both the training ($n = 3,281$) and testing sets, are illustrated in Figure~\ref{fig:llm-mt-bench}(a). Clearly, adaptive sampling achieves a substantially higher alignment accuracy, approaching $80\%$, whereas uniform sampling performs only marginally better than random guessing. The Bradley-Terry-Luce model also generalizes well in this setting, as evidenced by the close agreement between the training and testing accuracies. At the end of round $T = 100$, we compute the $\widehat{\theta}_T^a(x)$ for each context $x$ (i.e., question category) and for each arm $a$ (i.e., language model). Taking \texttt{alpaca-13b} as the baseline model ($\widehat{\theta}_T^1(x)=0$), we report the resulting estimates in Figure~\ref{fig:llm-mt-bench-theta}(a). The model \texttt{llama-13b} generally performs worse than \texttt{alpaca-13b} across nearly all categories except Roleplay, whereas \texttt{gpt-4} and \texttt{gpt-3.5-turbo} tend to outperform the other models across many categories.

We repeat the same analysis on the Chatbot Arena dataset, which consists of approximately $33,000$ pairwise comparisons collected through a crowdsourcing platform. Because category labels for the input prompts are not available in this dataset, we infer them using a support vector machine (SVM) classifier trained on the labeled MT-Bench prompts using TF--IDF features. The train-test split for the Chatbot Arena produced $24,750$ comparisons in training sample and $8,250$ comparisons in the testing sample, showcasing the analysis for a large dataset. The resulting alignment accuracy trajectories and context-specific preference scores are shown in Figures~\ref{fig:llm-mt-bench}(b) and~\ref{fig:llm-mt-bench-theta}(b), respectively. As in MT-Bench, we use \texttt{alpaca-13b} as the baseline model. Figure~\ref{fig:llm-mt-bench}(b) again shows a substantial improvement under adaptive sampling relative to uniform sampling. In contrast to the MT-Bench results, however, the estimated preference scores do not show consistent dominance of the GPT-based models. For example, among the models shown in panel (b), \texttt{claude-v1} has the highest estimated score for Roleplay. For STEM tasks, all four displayed models score above the \texttt{alpaca-13b} baseline. This discrepancy raises the possibility of a self-enhancement bias in the MT-Bench evaluations arising from the use of GPT-4 as an LLM judge, although differences between the two datasets may also contribute to the observed patterns.

\subsection{LMArena 140K Dataset}\label{sec:realdata-lmarena}

Over time, the chatbot arena project has evolved into a newer crowdsourcing platform ``LMArena''\footnote{\url{http://arena.ai/}} for evaluating large language models. A substantial portion of the crowdsourced data has been released publicly, encompassing approximately $140,000$ pairwise comparisons between $52$ language models. For our study, we restrict attention to only the top $5$ models having the most comparison data available, namely \texttt{claude-opus-4}, \texttt{gemini-2.5}, \texttt{mistral-medium}, \texttt{qwen3-235b}, and OpenAI's \texttt{o3}. This brought down the number of samples to $3,325$ comparisons, of which $2,493$ are kept in the training set and the remaining $832$ are kept in the testing set. As before, we use the trained SVM classifier to infer a category for each input prompt and run both uniform and adaptive sampling for $T=300$ rounds.

As shown in Figure~\ref{fig:llm-mt-bench}(c), adaptive sampling performs similarly to uniform sampling, with neither strategy achieving an alignment accuracy substantially above that of random guessing. This limited predictability highlights the difficulty of learning context-dependent preference rankings in this dataset. One possible explanation is that modern large language models have become increasingly competitive, with their performance approaching saturation on many existing evaluation tasks~\citep{akhtar2026ai}. As a result, no single model consistently dominates another within a given category, making broad task categories alone insufficiently informative for distinguishing among top-performing models. Nevertheless, this experiment provides empirical support for an important feature of our adaptive sampling strategy: under extreme uncertainty, the adaptive sampling allocation approaches uniform sampling and therefore gains little relative to the uniform baseline.

\section{Concluding Discussions}\label{sec:conclusion}

Typical data collection strategies for evaluating large language models rely on randomly selecting pairs of models for comparison by human or LLM judges. As we show in this paper, such uniform sampling can be statistically inefficient and lead to suboptimal regret. A more effective strategy is to couple sampling with learning, adaptively allocating observations across model pairs according to their contribution to the estimation uncertainty. In this work, we propose such an adaptive sampling rule, establish its minimax optimality by deriving matching upper and lower bounds on the regret, and demonstrate its empirical effectiveness across three benchmark datasets.

Our analysis also points to several directions for further improving the efficiency of the proposed sampling strategy. One limitation of the current adaptive strategy is that its finite-sample allocation does not explicitly account for the number of comparisons already collected for each pair. For example, consider two pairs of arms, $(a,b)$ and $(c,d)$. Suppose that, at some stage of the experiment, 100 comparisons have been collected for the first pair but only 80 for the second. Even if all arms have equal underlying rewards, so that the asymptotically optimal allocation assigns the same sampling proportion to the two pairs, a finite-sample procedure should preferentially sample the second pair until the two pairs are estimated with comparable precision.

Another possible refinement is motivated by the observation that, as the experiment progresses, certain arms may be identified as clearly suboptimal for particular tasks. Under the current sampling rule, the probability of selecting comparisons involving such arms decreases but never becomes exactly zero. Further efficiency gains may be achieved by eliminating such arms from subsequent pairwise comparisons altogether, thereby concentrating the sampling budget on more informative comparisons. We leave these extensions for future work.

Moreover, our design is also conceptually related to the principle of pessimism in policy learning~\citep{jin2021pessimism}. While pessimistic approaches account for estimation uncertainty directly in the policy optimization objective, our approach addresses such uncertainty through adaptive data collection, allowing the final policy to be learned using standard policy optimization procedures. This avoids the additional computational burden of constructing and optimizing pessimistic objectives, particularly over flexible or nonparametric policy classes.

\bibliography{references}%

@article{akhtar2026ai,
  author  = {Akhtar, Mubashara and Reuel, Anka and Soni, Prajna and Ahuja, Sanchit and Ammanamanchi, Pawan Sasanka and Rawal, Ruchit and Zouhar, Vil{\'e}m and Yadav, Srishti and Whitehouse, Chenxi and Ki, Dayeon and others},
  journal = {arXiv preprint arXiv:2602.16763},
  title   = {{When AI Benchmarks Plateau: A Systematic Study of Benchmark Saturation}},
  year    = {2026}
}

@article{wang2024comprehensive,
  title   = {A comprehensive review of multimodal large language models: Performance and challenges across different tasks},
  author  = {Wang, Jiaqi and Jiang, Hanqi and Liu, Yiheng and Ma, Chong and Zhang, Xu and Pan, Yi and Liu, Mengyuan and Gu, Peiran and Xia, Sichen and Li, Wenjun and others},
  journal = {arXiv preprint arXiv:2408.01319},
  year    = {2024}
}

@inproceedings{biyik2018batch,
  title        = {Batch active preference-based learning of reward functions},
  author       = {Biyik, Erdem and Sadigh, Dorsa},
  booktitle    = {Conference on robot learning},
  pages        = {519--528},
  year         = {2018},
  organization = {PMLR}
}

@inproceedings{singla2016actively,
  title        = {Actively learning hemimetrics with applications to eliciting user preferences},
  author       = {Singla, Adish and Tschiatschek, Sebastian and Krause, Andreas},
  booktitle    = {International Conference on Machine Learning},
  pages        = {412--420},
  year         = {2016},
  organization = {PMLR}
}

@inproceedings{yue2009interactively,
  title     = {Interactively optimizing information retrieval systems as a dueling bandits problem},
  author    = {Yue, Yisong and Joachims, Thorsten},
  booktitle = {Proceedings of the 26th annual international conference on machine learning},
  pages     = {1201--1208},
  year      = {2009}
}

@article{gosciak2025adaptive,
  title     = {Adaptive Randomization in Conjoint Survey Experiments},
  author    = {Gosciak, Jennah and Molitor, Daniel and Lundberg, Ian},
  journal   = {Political Analysis},
  pages     = {1--28},
  year      = {2025},
  publisher = {Cambridge University Press}
}

@inproceedings{audibert2010best,
  author    = {Audibert, Jean-Yves and Bubeck, S{\'e}bastien},
  booktitle = {{COLT-23th Conference on Learning Theory-2010}},
  pages     = {13--p},
  title     = {Best arm identification in multi-armed bandits},
  year      = {2010}
}

@book{basu2011statistical,
  author    = {Basu, Ayanendranath and Shioya, Hiroyuki and Park, Chanseok},
  publisher = {CRC press},
  title     = {{Statistical Inference: The Minimum Distance Approach}},
  year      = {2011}
}

@article{burnetas1997optimal,
  author    = {Burnetas, Apostolos N and Katehakis, Michael N},
  journal   = {Mathematics of operations research},
  number    = {1},
  pages     = {222--255},
  publisher = {INFORMS},
  title     = {{Optimal adaptive policies for Markov decision processes}},
  volume    = {22},
  year      = {1997}
}

@inproceedings{castano2024analyzing,
  author    = {Casta{\~n}o, Joel and Mart{\'\i}nez-Fern{\'a}ndez, Silverio and Franch, Xavier and Bogner, Justus},
  booktitle = {Proceedings of the 21st International Conference on Mining Software Repositories},
  pages     = {607--618},
  title     = {{Analyzing the evolution and maintenance of ML models on Hugging-face}},
  year      = {2024}
}

@article{chang2024survey,
  author    = {Chang, Yupeng and Wang, Xu and Wang, Jindong and Wu, Yuan and Yang, Linyi and Zhu, Kaijie and Chen, Hao and Yi, Xiaoyuan and Wang, Cunxiang and Wang, Yidong and others},
  journal   = {{ACM transactions on Intelligent Systems and Technology}},
  number    = {3},
  pages     = {1--45},
  publisher = {ACM New York, NY},
  title     = {{A survey on evaluation of large language models}},
  volume    = {15},
  year      = {2024}
}

@inproceedings{chowdhury2017kernelized,
  author       = {Chowdhury, Sayak Ray and Gopalan, Aditya},
  booktitle    = {International Conference on Machine Learning},
  organization = {PMLR},
  pages        = {844--853},
  title        = {On kernelized multi-armed bandits},
  year         = {2017}
}

@article{elwood2023maximum,
  author    = {Elwood, Adam and Leonardi, Marco and Mohamed, Ashraf and Rozza, Alessandro},
  journal   = {Entropy},
  number    = {2},
  pages     = {188},
  publisher = {MDPI},
  title     = {Maximum entropy exploration in contextual bandits with neural networks and energy based models},
  volume    = {25},
  year      = {2023}
}

@article{farrell2021deep,
  author    = {Farrell, Max H and Liang, Tengyuan and Misra, Sanjog},
  journal   = {Econometrica},
  number    = {1},
  pages     = {181--213},
  publisher = {Wiley Online Library},
  title     = {Deep neural networks for estimation and inference},
  volume    = {89},
  year      = {2021}
}

@article{ford1957solution,
  author    = {Ford Jr, Lester R},
  journal   = {The American Mathematical Monthly},
  number    = {8P2},
  pages     = {28--33},
  publisher = {Taylor \& Francis},
  title     = {Solution of a ranking problem from binary comparisons},
  volume    = {64},
  year      = {1957}
}

@article{hunter2004mm,
  author    = {Hunter, David R},
  journal   = {The annals of statistics},
  number    = {1},
  pages     = {384--406},
  publisher = {Institute of Mathematical Statistics},
  title     = {MM algorithms for generalized Bradley-Terry models},
  volume    = {32},
  year      = {2004}
}

@article{jones2001comparison,
  author    = {Jones, MC and Hjort, Nils Lid and Harris, Ian R and Basu, Ayanendranath},
  journal   = {Biometrika},
  pages     = {865--873},
  publisher = {JSTOR},
  title     = {A comparison of related density-based minimum divergence estimators},
  year      = {2001}
}

@article{kato2021role,
  author  = {Kato, Masahiro and Ariu, Kaito},
  journal = {arXiv preprint arXiv:2106.14077},
  title   = {The role of contextual information in best arm identification},
  year    = {2021}
}

@article{kato2024adaptive,
  author  = {Kato, Masahiro and Okumura, Kyohei and Ishihara, Takuya and Kitagawa, Toru},
  journal = {arXiv preprint arXiv:2401.03756},
  title   = {Adaptive experimental design for policy learning},
  year    = {2024}
}

@article{kaufmann2016complexity,
  author    = {Kaufmann, Emilie and Capp{\'e}, Olivier and Garivier, Aur{\'e}lien},
  journal   = {The Journal of Machine Learning Research},
  number    = {1},
  pages     = {1--42},
  publisher = {JMLR. org},
  title     = {On the complexity of best-arm identification in multi-armed bandit models},
  volume    = {17},
  year      = {2016}
}

@article{komiyama2022minimax,
  author  = {Komiyama, Junpei and Tsuchiya, Taira and Honda, Junya},
  journal = {Advances in Neural Information Processing Systems},
  pages   = {10393--10404},
  title   = {Minimax optimal algorithms for fixed-budget best arm identification},
  volume  = {35},
  year    = {2022}
}

@article{lai1985asymptotically,
  author    = {Lai, Tze Leung and Robbins, Herbert},
  journal   = {Advances in applied mathematics},
  number    = {1},
  pages     = {4--22},
  publisher = {Academic Press, Inc. Orlando, FL, USA},
  title     = {Asymptotically efficient adaptive allocation rules},
  volume    = {6},
  year      = {1985}
}

@article{li2024feel,
  author  = {Li, Xuheng and Zhao, Heyang and Gu, Quanquan},
  journal = {arXiv preprint arXiv:2404.06013},
  title   = {Feel-good thompson sampling for contextual dueling bandits},
  year    = {2024}
}

@article{li2025efficient,
  author  = {Li, Xiudi and Li, Sijia},
  journal = {arXiv preprint arXiv:2503.18256},
  title   = {Efficient Inference for Covariate-adjusted Bradley-Terry Model with Covariate Shift},
  year    = {2025}
}

@article{mourtada2017universal,
  author  = {Mourtada, Jaouad and Ga{\"\i}ffas, St{\'e}phane and Scornet, Erwan},
  journal = {Advances in neural information processing systems},
  title   = {Universal consistency and minimax rates for online mondrian forests},
  volume  = {30},
  year    = {2017}
}

@article{natarajan1989learning,
  author    = {Natarajan, Balas K},
  journal   = {Machine Learning},
  number    = {1},
  pages     = {67--97},
  publisher = {Springer},
  title     = {On learning sets and functions},
  volume    = {4},
  year      = {1989}
}

@article{neyman1923applications,
  author  = {Neyman, Jersey},
  journal = {Roczniki Nauk Rolniczych},
  number  = {1},
  pages   = {1--51},
  title   = {Sur les applications de la th{\'e}orie des probabilit{\'e}s aux experiences agricoles: Essai des principes},
  volume  = {10},
  year    = {1923}
}

@article{ong2024routellm,
  author  = {Ong, Isaac and Almahairi, Amjad and Wu, Vincent and Chiang, Wei-Lin and Wu, Tianhao and Gonzalez, Joseph E and Kadous, M Waleed and Stoica, Ion},
  journal = {arXiv preprint arXiv:2406.18665},
  title   = {Routellm: Learning to route llms with preference data},
  year    = {2024}
}

@article{ouyang2022training,
  author  = {Ouyang, Long and Wu, Jeffrey and Jiang, Xu and Almeida, Diogo and Wainwright, Carroll and Mishkin, Pamela and Zhang, Chong and Agarwal, Sandhini and Slama, Katarina and Ray, Alex and others},
  journal = {Advances in neural information processing systems},
  pages   = {27730--27744},
  title   = {Training language models to follow instructions with human feedback},
  volume  = {35},
  year    = {2022}
}

@article{parr1980minimum,
  author    = {Parr, William C and Schucany, William R},
  journal   = {Journal of the American Statistical Association},
  number    = {371},
  pages     = {616--624},
  publisher = {Taylor \& Francis},
  title     = {Minimum distance and robust estimation},
  volume    = {75},
  year      = {1980}
}

@article{Rakhlin_Sridharan_Tewari_2015,
  author    = {Rakhlin, Alexander and Sridharan, Karthik and Tewari, Ambuj},
  doi       = {10.1007/s00440-013-0545-5},
  issn      = {0178-8051},
  journal   = {Probability Theory and Related Fields},
  language  = {en},
  month     = feb,
  number    = {1–2},
  pages     = {111–153},
  publisher = {Springer Science and Business Media LLC},
  title     = {Sequential complexities and uniform martingale laws of large numbers},
  volume    = {161},
  year      = {2015}
}

@article{roy2024robust,
  author  = {Roy, Subhrajyoty and Basu, Ayanendranath and Ghosh, Abhik},
  journal = {Journal of Machine Learning Research},
  number  = {324},
  pages   = {1--40},
  title   = {Robust principal component analysis using density power divergence},
  volume  = {25},
  year    = {2024}
}

@article{rubin1974estimating,
  author    = {Rubin, Donald B},
  journal   = {Journal of educational Psychology},
  number    = {5},
  pages     = {688},
  publisher = {American Psychological Association},
  title     = {Estimating causal effects of treatments in randomized and nonrandomized studies.},
  volume    = {66},
  year      = {1974}
}

@article{rubin2005causal,
  author    = {Rubin, Donald B},
  journal   = {Journal of the American statistical Association},
  number    = {469},
  pages     = {322--331},
  publisher = {Taylor \& Francis},
  title     = {Causal inference using potential outcomes: Design, modeling, decisions},
  volume    = {100},
  year      = {2005}
}

@inproceedings{russo2016simple,
  author       = {Russo, Daniel},
  booktitle    = {{Conference on Learning Theory}},
  organization = {PMLR},
  pages        = {1417--1418},
  title        = {{Simple Bayesian Algorithms for Best Arm Identification}},
  year         = {2016}
}

@article{saha2021optimal,
  author  = {Saha, Aadirupa},
  journal = {{Advances in Neural Information Processing Systems}},
  pages   = {30050--30062},
  title   = {{Optimal Algorithms for Stochastic Contextual Preference Bandits}},
  volume  = {34},
  year    = {2021}
}

@article{sun2024rethinking,
  author  = {Sun, Hao and Shen, Yunyi and Ton, Jean-Francois},
  journal = {{arXiv preprint arXiv:2411.04991}},
  title   = {{Rethinking Bradley-Terry Models in Preference-Based Reward Modeling: Foundations, Theory, and Alternatives}},
  year    = {2024}
}

@article{vapnik1971uniform,
  author    = {Vapnik, Vladimir Naumovich and Chervonenkis, Alexey Yakovlevich},
  journal   = {{Probability Theory and Its Applications}},
  number    = {2},
  pages     = {264--279},
  publisher = {Russian Academy of Sciences, Steklov Mathematical Institute of the Russian~…},
  title     = {{On the Uniform Convergence of the Frequencies of Occurrence of Events to Their Probabilities}},
  volume    = {16},
  year      = {1971}
}

@article{wager2018estimation,
  author    = {Wager, Stefan and Athey, Susan},
  journal   = {{Journal of the American Statistical Association}},
  number    = {523},
  pages     = {1228--1242},
  publisher = {Taylor \& Francis},
  title     = {{Estimation and Inference of Heterogeneous Treatment Effects Using Random Forests}},
  volume    = {113},
  year      = {2018}
}

@book{wainwright2019high,
  author    = {Wainwright, Martin J},
  publisher = {Cambridge university press},
  title     = {{High-Dimensional Statistics: A Non-Asymptotic Viewpoint}},
  volume    = {48},
  year      = {2019}
}

@article{weidinger2025toward,
  author  = {Weidinger, Laura and Raji, Inioluwa Deborah and Wallach, Hanna and Mitchell, Margaret and Wang, Angelina and Salaudeen, Olawale and Bommasani, Rishi and Ganguli, Deep and Koyejo, Sanmi and Isaac, William},
  journal = {{arXiv preprint arXiv:2503.05336}},
  title   = {{Toward an Evaluation Science for Generative AI Systems}},
  year    = {2025}
}

@article{zhan2024policy,
  author    = {Zhan, Ruohan and Ren, Zhimei and Athey, Susan and Zhou, Zhengyuan},
  journal   = {{Management Science}},
  number    = {8},
  pages     = {5270--5297},
  publisher = {INFORMS},
  title     = {{Policy Learning with Adaptively Collected Data}},
  volume    = {70},
  year      = {2024}
}

@article{zhao2023survey,
  author  = {Zhao, Wayne Xin and Zhou, Kun and Li, Junyi and Tang, Tianyi and Wang, Xiaolei and Hou, Yupeng and Min, Yingqian and Zhang, Beichen and Zhang, Junjie and Dong, Zican and others},
  journal = {{arXiv preprint arXiv:2303.18223}},
  number  = {2},
  pages   = {1--124},
  title   = {{A Survey of Large Language Models}},
  volume  = {1},
  year    = {2023}
}

@article{zheng2023judging,
  author  = {Zheng, Lianmin and Chiang, Wei-Lin and Sheng, Ying and Zhuang, Siyuan and Wu, Zhanghao and Zhuang, Yonghao and Lin, Zi and Li, Zhuohan and Li, Dacheng and Xing, Eric and others},
  journal = {{Advances in Neural Information Processing Systems}},
  pages   = {46595--46623},
  title   = {{Judging LLM-as-a-Judge with MT-Bench and Chatbot Arena}},
  volume  = {36},
  year    = {2023}
}

@article{villar2015multi,
  title   = {Multi-armed bandit models for the optimal design of clinical trials: benefits and challenges},
  author  = {Villar, Sof{\'\i}a S and Bowden, Jack and Wason, James},
  journal = {Statistical science: a review journal of the Institute of Mathematical Statistics},
  volume  = {30},
  number  = {2},
  pages   = {199},
  year    = {2015}
}

@article{yan2022bandit,
  title   = {Bandits for Recommender Systems},
  author  = {Yan, Ziyou},
  journal = {eugeneyan.com},
  year    = {2022},
  month   = {May},
  url     = {https://eugeneyan.com/writing/bandits/}
}

@inproceedings{ban2024neural,
  title     = {Neural contextual bandits for personalized recommendation},
  author    = {Ban, Yikun and Qi, Yunzhe and He, Jingrui},
  booktitle = {Companion Proceedings of the ACM Web Conference 2024},
  pages     = {1246--1249},
  year      = {2024}
}

@article{zhou2023offline,
  author    = {Zhou, Zhengyuan and Athey, Susan and Wager, Stefan},
  journal   = {{Operations Research}},
  number    = {1},
  pages     = {148--183},
  publisher = {INFORMS},
  title     = {{Offline Multi-Action Policy Learning: Generalization and Optimization}},
  volume    = {71},
  year      = {2023}
}

@article{christiano2017deep,
  title   = {Deep reinforcement learning from human preferences},
  author  = {Christiano, Paul F and Leike, Jan and Brown, Tom and Martic, Miljan and Legg, Shane and Amodei, Dario},
  journal = {Advances in Neural Information Processing Systems},
  volume  = {30},
  year    = {2017}
}

@inproceedings{panda2025adaptive,
  title     = {Adaptive LLM Routing under Budget Constraints.},
  author    = {Panda, Pranoy and Magazine, Raghav and Devaguptapu, Chaitanya and Takemori, Sho and Sharma, Vishal},
  booktitle = {EMNLP (Findings)},
  pages     = {23934--23949},
  year      = {2025}
}

@inproceedings{jin2023upper,
  title        = {Upper bounds on the natarajan dimensions of some function classes},
  author       = {Jin, Ying},
  booktitle    = {2023 IEEE International Symposium on Information Theory (ISIT)},
  pages        = {1020--1025},
  year         = {2023},
  organization = {IEEE}
}

@article{fang2026recent,
  title   = {Recent advances in the Bradley--Terry Model: theory, algorithms, and applications},
  author  = {Fang, Shuxing and Han, Ruijian and Luo, Yuanhang and Xu, Yiming},
  journal = {arXiv preprint arXiv:2601.14727},
  year    = {2026}
}

@inproceedings{chen2015spectral,
  title        = {Spectral mle: Top-k rank aggregation from pairwise comparisons},
  author       = {Chen, Yuxin and Suh, Changho},
  booktitle    = {International conference on machine learning},
  pages        = {371--380},
  year         = {2015},
  organization = {PMLR}
}

@article{negahban2012iterative,
  title   = {Iterative ranking from pair-wise comparisons},
  author  = {Negahban, Sahand and Oh, Sewoong and Shah, Devavrat},
  journal = {Advances in neural information processing systems},
  volume  = {25},
  year    = {2012}
}

@article{chen2019spectral,
  title   = {Spectral method and regularized MLE are both optimal for top-K ranking},
  author  = {Chen, Yuxin and Fan, Jianqing and Ma, Cong and Wang, Kaizheng},
  journal = {Annals of statistics},
  volume  = {47},
  number  = {4},
  pages   = {2204},
  year    = {2019}
}

@article{chen2022partial,
  title     = {PARTIAL RECOVERY FOR TOP-k RANKING},
  author    = {Chen, Pinhan and Gao, Chao and Zhang, Anderson Y},
  journal   = {The Annals of Statistics},
  volume    = {50},
  number    = {3},
  pages     = {1618--1652},
  year      = {2022},
  publisher = {JSTOR}
}

@article{chen2022optimal,
  title     = {Optimal full ranking from pairwise comparisons},
  author    = {Chen, Pinhan and Gao, Chao and Zhang, Anderson Y},
  journal   = {The Annals of Statistics},
  volume    = {50},
  number    = {3},
  pages     = {1775--1805},
  year      = {2022},
  publisher = {JSTOR}
}

@article{simons1999asymptotics,
  title     = {Asymptotics when the number of parameters tends to infinity in the Bradley-Terry model for paired comparisons},
  author    = {Simons, Gordon and Yao, Yi-Ching},
  journal   = {The Annals of Statistics},
  volume    = {27},
  number    = {3},
  pages     = {1041--1060},
  year      = {1999},
  publisher = {Institute of Mathematical Statistics}
}

@article{han2020asymptotic,
  title   = {Asymptotic theory of sparse Bradley--Terry model},
  author  = {Han, Ruijian and Ye, Rougang and Tan, Chunxi and Chen, Kani},
  journal = {The Annals of Applied Probability},
  volume  = {30},
  number  = {5},
  pages   = {2491--2515},
  year    = {2020}
}

@article{liu2023lagrangian,
  title     = {Lagrangian inference for ranking problems},
  author    = {Liu, Yue and Fang, Ethan X and Lu, Junwei},
  journal   = {Operations research},
  volume    = {71},
  number    = {1},
  pages     = {202--223},
  year      = {2023},
  publisher = {INFORMS}
}

@article{gao2023uncertainty,
  title     = {Uncertainty quantification in the Bradley--Terry--Luce model},
  author    = {Gao, Chao and Shen, Yandi and Zhang, Anderson Y},
  journal   = {Information and Inference: A Journal of the IMA},
  volume    = {12},
  number    = {2},
  pages     = {1073--1140},
  year      = {2023},
  publisher = {Oxford University Press}
}

@article{fan2025ranking,
  title     = {Ranking inferences based on the top choice of multiway comparisons},
  author    = {Fan, Jianqing and Lou, Zhipeng and Wang, Weichen and Yu, Mengxin},
  journal   = {Journal of the American Statistical Association},
  volume    = {120},
  number    = {549},
  pages     = {237--250},
  year      = {2025},
  publisher = {Taylor \& Francis}
}

@article{fan2026spectral,
  title     = {Spectral ranking inferences based on general multiway comparisons},
  author    = {Fan, Jianqing and Lou, Zhipeng and Wang, Weichen and Yu, Mengxin},
  journal   = {Operations Research},
  volume    = {74},
  number    = {1},
  pages     = {161--180},
  year      = {2026},
  publisher = {INFORMS}
}

@article{han2025unified,
  title     = {A unified analysis of likelihood-based estimators in the Plackett--Luce model},
  author    = {Han, Ruijian and Xu, Yiming},
  journal   = {The Annals of Statistics},
  volume    = {53},
  number    = {5},
  pages     = {2077--2102},
  year      = {2025},
  publisher = {Institute of Mathematical Statistics}
}

@article{dong2026statistical,
  title     = {Statistical ranking with dynamic covariates},
  author    = {Dong, Pinjun and Han, Ruijian and Jiang, Binyan and Xu, Yiming},
  journal   = {Journal of the Royal Statistical Society Series B: Statistical Methodology},
  volume    = {88},
  number    = {1},
  pages     = {221--238},
  year      = {2026},
  publisher = {Oxford University Press UK}
}

@article{fan2024covariate,
  title   = {Covariate assisted entity ranking with sparse intrinsic scores},
  author  = {Fan, Jianqing and Hou, Jikai and Yu, Mengxin},
  journal = {arXiv preprint arXiv:2407.08814},
  year    = {2024}
}

@article{luo2026deep,
  title   = {Deep Ranking with Heterogeneous Effect},
  author  = {Luo, Yuanhang and Fang, Shuxing and Han, Ruijian and Xu, Yiming},
  journal = {arXiv preprint arXiv:2604.16129},
  year    = {2026}
}

@article{fan2024uncertainty,
  title   = {Uncertainty quantification of MLE for entity ranking with covariates},
  author  = {Fan, Jianqing and Hou, Jikai and Yu, Mengxin},
  journal = {Journal of Machine Learning Research},
  volume  = {25},
  number  = {358},
  pages   = {1--83},
  year    = {2024}
}

@article{bengs2021preference,
  title   = {Preference-based online learning with dueling bandits: A survey},
  author  = {Bengs, Viktor and Busa-Fekete, R{\'o}bert and El Mesaoudi-Paul, Adil and H{\"u}llermeier, Eyke},
  journal = {Journal of Machine Learning Research},
  volume  = {22},
  number  = {7},
  pages   = {1--108},
  year    = {2021}
}

@inproceedings{zoghi2014relative,
  title        = {Relative upper confidence bound for the k-armed dueling bandit problem},
  author       = {Zoghi, Masrour and Whiteson, Shimon and Munos, Remi and Rijke, Maarten},
  booktitle    = {International conference on machine learning},
  pages        = {10--18},
  year         = {2014},
  organization = {PMLR}
}

@article{chiang2024chatbot,
  title   = {Chatbot arena: An open platform for evaluating llms by human preference},
  author  = {Chiang, Wei-Lin and Zheng, Lianmin and Sheng, Ying and Angelopoulos, Anastasios Nikolas and Li, Tianle and Li, Dacheng and Zhang, Hao and Zhu, Banghua and Jordan, Michael and Gonzalez, Joseph E and others},
  journal = {arXiv preprint arXiv:2403.04132},
  year    = {2024}
}

@article{casper2023open,
  title   = {Open problems and fundamental limitations of reinforcement learning from human feedback},
  author  = {Casper, Stephen and Davies, Xander and Shi, Claudia and Gilbert, Thomas Krendl and Scheurer, J{\'e}r{\'e}my and Rando, Javier and Freedman, Rachel and Korbak, Tomasz and Lindner, David and Freire, Pedro and others},
  journal = {arXiv preprint arXiv:2307.15217},
  year    = {2023}
}

@inproceedings{di2024variance,
  title     = {Variance-aware regret bounds for stochastic contextual dueling bandits},
  author    = {Di, Qiwei and Jin, Tao and Wu, Yue and Zhao, Heyang and Farnoud, Farzad and Gu, Quanquan},
  booktitle = {International Conference on Learning Representations},
  volume    = {2024},
  pages     = {36738--36761},
  year      = {2024}
}

@article{zrnic2024active,
  title   = {Active statistical inference},
  author  = {Zrnic, Tijana and Cand{\`e}s, Emmanuel J},
  journal = {arXiv preprint arXiv:2403.03208},
  year    = {2024}
}

@article{li2026robust,
  title   = {Robust sampling for active statistical inference},
  author  = {Li, Puheng and Zrnic, Tijana and Candes, Emmanuel},
  journal = {Advances in Neural Information Processing Systems},
  volume  = {38},
  pages   = {68727--68756},
  year    = {2026}
}

@article{ao2024prediction,
  title   = {Prediction-guided active experiments},
  author  = {Ao, Ruicheng and Chen, Hongyu and Simchi-Levi, David},
  journal = {arXiv preprint arXiv:2411.12036},
  year    = {2024}
}

@article{dong2026labels,
  title   = {{Labels or Preferences? Budget-Constrained Learning with Human Judgments over AI-Generated Outputs}},
  author  = {Dong, Zihan and Hou, Xiaotian and Wu, Ruijia and Zhang, Linjun},
  journal = {arXiv preprint arXiv:2601.13458},
  year    = {2026}
}

@article{liu2024dual,
  title   = {Dual active learning for reinforcement learning from human feedback},
  author  = {Liu, Pangpang and Shi, Chengchun and Sun, Will Wei},
  journal = {arXiv preprint arXiv:2410.02504},
  year    = {2024}
}

@article{han2023general,
  title     = {A general pairwise comparison model for extremely sparse networks},
  author    = {Han, Ruijian and Xu, Yiming and Chen, Kani},
  journal   = {Journal of the American Statistical Association},
  volume    = {118},
  number    = {544},
  pages     = {2422--2432},
  year      = {2023},
  publisher = {Taylor \& Francis}
}

@article{haussler1995sphere,
  title     = {Sphere packing numbers for subsets of the Boolean n-cube with bounded Vapnik-Chervonenkis dimension},
  author    = {Haussler, David},
  journal   = {Journal of Combinatorial Theory, Series A},
  volume    = {69},
  number    = {2},
  pages     = {217--232},
  year      = {1995},
  publisher = {Elsevier}
}

@article{athey2021policy,
  title     = {Policy learning with observational data},
  author    = {Athey, Susan and Wager, Stefan},
  journal   = {Econometrica},
  volume    = {89},
  number    = {1},
  pages     = {133--161},
  year      = {2021},
  publisher = {Wiley Online Library}
}

@article{liu2026statistical,
  title     = {Statistical impossibility and possibility of aligning LLMs with human preferences: From Condorcet paradox to Nash equilibrium},
  author    = {Liu, Kaizhao and Long, Qi and Shi, Zhekun and Su, Weijie J and Xiao, Jiancong},
  journal   = {The Annals of Statistics},
  volume    = {54},
  number    = {4},
  pages     = {2110--2135},
  year      = {2026},
  publisher = {Institute of Mathematical Statistics}
}

@inproceedings{jin2021pessimism,
  title        = {Is pessimism provably efficient for offline rl?},
  author       = {Jin, Ying and Yang, Zhuoran and Wang, Zhaoran},
  booktitle    = {International conference on machine learning},
  pages        = {5084--5096},
  year         = {2021},
  organization = {PMLR}
}

\pagebreak
\appendix

\section{Important Concepts and Definitions}\label{appendix:defn}

\subsection{The Policy Difference class and Rademacher Complexity}\label{appendix:policy-diff-Rademacher}

Given the class of policies $\Pi$, define a policy difference class, 
\begin{equation*}
\Pi^D = \left\{ h: \chi \times \R^K \to \R \mid h(x, \{ \Gamma^a \}_{a \in [K]} ) = \sum_{a \in [K]} (\pi_1(a \mid x) - \pi_2(a \mid x)) \Gamma^a, \ \pi_1, \pi_2 \in \Pi\right\}
\end{equation*}
\noindent which contains all possible values of performance difference between two policies $\pi_1$ and $\pi_2$ from the policy class $\Pi$. The empirical Rademacher complexity of this class is defined as 
\begin{equation*}
    \Rcal_T(\Pi^D \mid \{ X_t, \bb{\Gamma}_t^{\ast} \}_{t = 1}^T ) \\
    = \E_{\varepsilon}\left[ \sup_{\pi_1, \pi_2 \in \Pi} \dfrac{1}{T}\left\vert \sum_{t = 1}^T \varepsilon_t \delta_t^\ast(\pi_1, \pi_2) \right\vert \mid \{ X_t, \bb{\Gamma}_t^{\ast} \}_{t = 1}^T \right]
\end{equation*}
\noindent where
\begin{equation}
    \delta_t^\ast(\pi_1, \pi_2) = \sum_{a \in [K]} (\pi_1(a \mid X_t) - \pi_2(a \mid X_t))\Gamma_t^{a,\ast},
    \label{eqn:small-delta-t-star-defn}
\end{equation}
\noindent and the expectation is taken with respect to the i.i.d. Rademacher random variables $\varepsilon_1, \dots, \varepsilon_T$ satisfying $\prob(\varepsilon_t = -1) = \prob(\varepsilon_t = 1) = 1/2$. The Rademacher complexity of the class $\Pi^D$ is defined as the expectation of this empirical Rademacher complexity, where the expectation is taken over the distribution of context $X_t \in \chi$ and the adjusted value function $\bb{\Gamma}^\ast_t$ based on the observations. In mathematical notations, 
\begin{equation*}
    \Rcal_T(\Pi^D) = \E\left[ \Rcal_T(\Pi^D \mid \{ X_t, \bb{\Gamma}_t^{\ast} \}_{t = 1}^T ) \right].
\end{equation*} 

\subsection{The Dependency Tree and Rademacher Tree Process}\label{appendix:z-tree}

As a generalization of the Rademacher complexity from the i.i.d. sequences to Martingale differences, \cite{Rakhlin_Sridharan_Tewari_2015} introduced the notion of dependency tree and Rademacher tree process. 

\begin{definition}[Definition 3 of~\cite{zhan2024policy}]
    A $\mathcal{Z}$-valued tree $\bb{z}$ of depth $T$ is a rooted binary tree with nodes labeled by elements of $\mathcal{Z}$. We identify the tree $\bb{z}$ with the sequence $(\bb{z}_1, \dots, \bb{z}_T)$ where each $\bb{z}_t: \{ -1, 1\}^{t-1} \mapsto \mathcal{Z}$, is a labelling function.
\end{definition}

Note that, if $\varepsilon_1, \dots, \varepsilon_T$ are i.i.d. Rademacher random variables, then
\begin{equation*}
    \varepsilon_t \bb{z}_t(\varepsilon_1, \dots, \varepsilon_{t-1}),
\end{equation*}
\noindent is a Martingale difference sequence. As a result, if we have two identically distributed copies of Martingale difference sequences $\bb{Z}_t \in \mathcal{Z}$ and $\bb{Z}'_t \in \mathcal{Z}$ with respect to the common filtration $\Fcal_t$, then, by symmetry of $\bb{Z}_t - \bb{Z}'_t$, their dependence structure can be captured by this tree. In fact, conditional on $\varepsilon_1, \dots, \varepsilon_{t-1}$, $\bb{z}_{t+1}(\varepsilon_1, \dots, \varepsilon_{t-1}, +1)$ and $\bb{z}_{t+1}(\varepsilon_1, \dots, \varepsilon_{t-1}, -1)$ are independent of each other; and hence one can carefully choose a labelling function $\bb{z}$ that agrees with $\bb{Z}_t - \bb{Z}'_t$ on one branch and agrees with $\bb{Z}'_t - \bb{Z}_t$ on another.

\begin{definition}[Definition 2 of~\cite{Rakhlin_Sridharan_Tewari_2015}]
    Let $f: \mathcal{Z} \mapsto \R$ be any measurable function, and let $\varepsilon_1, \dots, \varepsilon_T$ be i.i.d. Rademacher random variables. Then, given a $\mathcal{Z}$-valued tree $\bb{z}$ of depth $T$, the stochastic process indexed by the function $f$, 
    \begin{equation*}
        f \mapsto \frac{1}{T}\sum_{t=1}^T \varepsilon_t f(\bb{z}_t(\varepsilon_1, \dots, \varepsilon_{t-1})),
    \end{equation*}
    \noindent is called a Rademacher tree process.
\end{definition}

\section{Proofs and Technical Details}\label{appendix:proof}

\subsection{Proof of Lemma~\ref{lem:gamma-est-bound}}\label{appendix:proof-lem-gamma-bound}

\begin{proof}
    \noindent Due to Assumption~\ref{assum:positivity}, we establish the boundedness for the estimated quantities, $\hat{v}_t^{ab}(x)$, $\hat{w}_t^{ab}(x)$, etc. The analogous bound holds for the corresponding population terms $v^{ab}(x)$, $w^{ab}(x)$, etc. as well.

    \paragraph{Boundedness of weights.} Because of Assumption~\ref{assum:positivity}, we have for any $x \in \chi$,
    \begin{equation*}
        \widehat{v}_t^{ab}(x) = \widehat{m}_t^{ab}(x) (1 - \widehat{m}_t^{ab}(x)) \in [\delta(1- \delta), 1/4].
    \end{equation*}
    \noindent Consequently, 
    \begin{equation*}
        \widehat{w}_t^{ab}(x) = \frac{\widehat{v}_t^{ab}(x)}{\sum_{a, b \in [K]_2} \widehat{v}_t^{ab}(x)} \in \left[ \frac{8\delta(1-\delta)}{K(K-1)}, \frac{1}{2K(K-1)\delta(1-\delta)} \right].
    \end{equation*}

    \paragraph{Boundedness of preference parameters.} Suppose for the sake of contradiction, $\widehat{\theta}_t^{a}(x)$'s are not uniformly bounded. Hence, there exists an $x_0 \in \chi$ such that 
    \begin{equation*}
        \max_{a \in [K]} {\widehat{\theta}}_t^{a}(x_0) - \min_{a \in [K]} {\widehat{\theta}}_t^{a}(x_0) > C,
    \end{equation*}
    \noindent for $t \geq 1$ and any arbitrarily large constant $C$. 
    If we order the estimates ${\widehat{\theta}}_t^{1}(x_0) \geq {\widehat{\theta}}_t^{2}(x_0) \geq \dots \geq {\widehat{\theta}}_t^{K}(x_0)$, then, there exists a $j \in [K]$ such that 
    \begin{equation*}
        {\widehat{\theta}}_t^{j}(x_0) - {\widehat{\theta}}_t^{j+1}(x_0) \geq C/K.
    \end{equation*}
    \noindent We now partition the arms into two sets, $S_1 = [j]$ and $S_2 = [K] \setminus [j]$. Combining all the estimating equations leads to
    \begin{equation}
        \sum_{a \in S_1}\sum_{b \in [K] \setminus \{a\} } \rho^{ab}(x_0) (\sigma({\widehat{\theta}}_t^{a}(x_0) - {\widehat{\theta}}_t^{b}(x_0)) - \widehat{m}_t^{ab}(x_0) ) = 0.
        \label{eqn:all-est-equation}
    \end{equation}
    \noindent Since $\rho^{ab}(x) = \rho^{ba}(x)$ and $\widehat{m}_t^{ab}(x) + \widehat{m}_t^{ba}(x) = 1$ for any $a, b \in [K]$, it follows that the summation in~\eqref{eqn:all-est-equation} corresponding to $a, b \in S_1$ is zero, as for any $a, b \in S_1$, 
    \begin{equation*}
        \rho^{ab}(x_0) (\sigma({\widehat{\theta}}_t^{a}(x_0) - {\widehat{\theta}}_t^{b}(x_0)) - \widehat{m}_t^{ab}(x_0) ) + \rho^{ba}(x_0)(\sigma({\widehat{\theta}}_t^{b}(x_0) - {\widehat{\theta}}_t^{a}(x_0)) - \widehat{m}_t^{ba}(x_0) )
        = 0.
    \end{equation*}
    \noindent Similarly, the summation corresponding to $a, b \in S_2$ also equates to $0$. Therefore, 
    \begin{equation}
        \sum_{a \in S_1}\sum_{b \in S_2 \setminus \{a\} } \rho^{ab}(x_0) (\sigma({\widehat{\theta}}_t^{a}(x_0) - {\widehat{\theta}}_t^{b}(x_0)) - \widehat{m}_t^{ab}(x_0) ) = 0.
        \label{eqn:all-est-equation-2}
    \end{equation}
    \noindent However, for any $a \in S_1$ and $b \in S_2$, the difference ${\widehat{\theta}}_t^{a}(x_0) - {\widehat{\theta}}_t^{b}(x_0) > C/K$.
    Choose $C = K \log(1/\delta - 1)$. Then, $\sigma({\widehat{\theta}}_t^{a}(x_0) - {\widehat{\theta}}_t^{b}(x_0)) > \sigma(C/K) = 1 - \delta$. However, $\widehat{m}_t^{ab}(x_0) \leq 1 - \delta$. Therefore, each of the summands in~\eqref{eqn:all-est-equation-2} must be positive, leading to a contradiction.

    Therefore, we obtain that uniformly for any $x \in \chi$, 
    \begin{equation*}
        \max_{a \in [K]} {\widehat{\theta}}_t^{a}(x) - \min_{a \in [K]} {\widehat{\theta}}_t^{a}(x) \leq K\log(1/\delta - 1).
    \end{equation*}
    \noindent However, since $\widehat{\theta}_t^{1}(x) = 0$ by definition, it implies that $\vert {\widehat{\theta}}_t^{a}(x)\vert \leq K\log(1/\delta - 1)$ for all $a \in [K]$ and $x \in \chi$.

    \paragraph{Boundedness of $\Gamma_t^a$.} As the previous derivation shows, $\widehat{\theta}_t^{a}(x)$s are uniformly bounded by $K\log(1/\delta - 1)$. Invoking the Gershgorin disc theorem, we obtain that the smallest eigenvalue of $\nabla_{\bb{\theta}}\bb{U}(\bb{\theta}, x)$ must satisfy 
    \begin{align}
        \lambda_{K-1}(\nabla_{\bb{\theta}}\bb{U}(\bbhat{\theta}, x))
        & \geq \min_{i} \left[ [\nabla_{\bb{\theta}}\bb{U}(\bbhat{\theta}_t, x)]_{ii} - \sum_{j \neq i} \vert  [\nabla_{\bb{\theta}}\bb{U}(\bbhat{\theta}_t, x)]_{ij} \vert \right] \nonumber \\
        & = \rho_{i1}(x) \sigma(\widehat{\theta}_t^i(x)) (1 - \sigma(\widehat{\theta}^i_t(x))), \ \text{ since } \widehat{\theta}_t^1(x) = 0 \nonumber \\
        & \geq \delta \sigma(K\log(1/\delta - 1)) (1 - \sigma(K\log(1/\delta - 1))) 
        = \frac{\delta^{K+1}(1-\delta)^K}{(\delta^K + (1-\delta)^K)^2} > 0.
        \label{eqn:Utheta-diff-eigen-lower-bound}
    \end{align}
    \noindent Also, 
    \begin{equation*}
        \left\Vert \nabla_{\bb{m}}\bb{U}(X_t) \right\Vert_F^2
        = \sum_{(a, b) \in [K]_2} (\rho^{ab}(x))^2
        \leq \sum_{(a, b) \in [K]_2} \rho^{ab}(x) = 1.
    \end{equation*}
    \noindent The inequality follows from the fact that each $\rho^{ab}(x) \in (0, 1)$. As a result, we have 
    \begin{align*}
        \Vert \bbhat{\Gamma}_t\Vert_2
        & \leq \Vert \bbhat{\theta}_t\Vert_2 + \left\Vert [\nabla_{\bb{\theta}}\bbhat{U}_t]^{-1} [\nabla_{\bb{m}}\bb{U}(X_t)] \bbhat{e}_t(X_t) \right\Vert\\
        & \leq \Vert \bbhat{\theta}_t\Vert_2 + \dfrac{1}{\lambda_{K-1}(\nabla_{\bb{\theta}}\bbhat{U}_t)} \left\Vert \nabla_{\bb{m}}\bb{U}(X_t) \right\Vert_F \left\Vert \bbhat{e}_t(X_t) \right\Vert_2 \\
        & \leq K^{3/2} \log\left( \delta^{-1} - 1 \right) + \frac{(\delta^K + (1-\delta)^K)^2}{\delta^{K+1}(1-\delta)^K} \cdot 1 \cdot \sqrt{2} \frac{K(K-1)}{8\delta(1-\delta)}(1 - \delta)\\
        & = K^{3/2} \log\left( \delta^{-1} - 1 \right) + \frac{K(K-1)}{4\sqrt{2}}\frac{(\delta^K + (1-\delta)^K)^2}{\delta^{K+2}(1-\delta)^{K}} < \infty
    \end{align*}
    \noindent Thus ensuring that each element of $\bbhat{\Gamma}_t$ is uniformly bounded.
\end{proof}

\subsection{Proof of Lemma~\ref{lem:consistency-bounds}}\label{appendix:proof-lem-consistency-bounds}

\begin{proof}
    Firstly, we know that since there are only finitely many choices of $a$ and $b$, any convergence for fixed $a$ and $b$ can be lifted to a uniform convergence over the choices of the pair $(a, b) \in [K]^2$. Therefore, we shall indicate only pointwise convergence, by fixing any $a, b \in [K]^2$. For notational convenience, let us define $\Vert A(X) \Vert_2^2 = \E( (\sup_x \vert A(x)\vert )^2)$, for a random variable $A$ depending on the context $x$.

    \noindent\paragraph{Convergence of $\widehat{v}^{ab}_t(x)$.} As $m^{ab}(x) \leq 1$ and $\widehat{m}_t^{ab}(x) \leq 1$, we have
    \begin{equation}
        \sup_{x \in \chi} \vert \widehat{m}_t^{ab}(x)(1 - \widehat{m}_t^{ab}(x)) -  m^{ab}(x)(1 - m^{ab}(x)) \vert
        \leq 3 \sup_{x \in \chi} \vert m^{ab}(x) - \widehat{m}_t^{ab}(x) \vert.
        \label{eqn:m-weight-convergence}
    \end{equation}
    \noindent Therefore,
    \begin{equation*}
        \Vert \widehat{v}_t^{ab}(X) - v^{ab}(X) \Vert_2^2
        \leq 9 \Vert \widehat{m}_t^{ab}(X) - m^{ab}(X) \Vert_2^2 = o(t^{-1/2}).
    \end{equation*}

    \paragraph{Convergence of $\widehat{w}_t^{ab}(x)$.} By lifting this convergence to uniform convergence over the choices of $(a, b) \in [K]_2$, we also have 
    \begin{multline*}
        \E\left[ \sup_{x \in \chi}\left\vert \sum_{(a, b) \in [K]_2} \widehat{v}_t^{ab}(x) - \sum_{(a, b) \in [K]_2} v^{ab}(x) \right\vert^2 \right]\\
        \leq K^4 \sup_{(a, b) \in [K]_2} \E\left[ \sup_{x \in \chi} \left\vert  \widehat{v}_t^{ab}(x) - v^{ab}(x) \right\vert^2 \right]
        = o(t^{-1/2}).
    \end{multline*}
    \noindent By using the positivity assumption~\ref{assum:positivity} and the boundedness of $w_t^{ab}(x)$ (see the proof of Lemma~\ref{lem:gamma-est-bound}), we get 
    \begin{align*}
        & \sup_{x \in \chi} \vert \widehat{w}_t^{ab}(x) - w^{ab}(x) \vert\\
        \leq{} & \sup_{x \in \chi}\left\vert \widehat{v}_t^{ab}(x) \left( \frac{1}{\sum_{(a, b) \in [K]_2} \widehat{v}_t^{ab}(x) } - \frac{1}{\sum_{(a,b) \in [K]_2} v^{ab}(x) } \right) \right\vert + \sup_{x\in \chi}\left\vert \frac{\widehat{v}^{ab}_t(x) - v^{ab}(x)}{\sum_{(a, b) \in [K]_2} v^{ab}(x) } \right\vert\\
        \leq{} & \frac{K(K-1)}{8\delta^2(1-\delta)^2} \sup_{(a, b) \in [K]_2} \sup_{x \in \chi} \vert \widehat{v}_t^{ab}(x) - v^{ab}(x)\vert + \frac{2}{K(K-1)\delta (1-\delta)} \sup_{x \in \chi} \vert \widehat{v}_t^{ab}(x) - v^{ab}(x)\vert\\
        \leq{} & \frac{K^2(K-1)^2 + 16\delta(1-\delta)}{8\delta^2(1-\delta)^2 K(K-1)} \sup_{(a, b) \in [K]_2} \sup_{x \in \chi} \vert \widehat{v}_t^{ab}(x) - v^{ab}(x)\vert
    \end{align*}
    \noindent Taking squares and then expectation on both sides yield
    \begin{equation*}
        \E\left[ \sup_{x \in \chi} \vert \widehat{w}_t^{ab}(x) - w^{ab}(x) \vert^2 \right] = o(t^{-1/2}).
    \end{equation*}

    \paragraph{Convergence of $\bbhat{\theta}_t(x)$.} We first note that the lower bound of the minimum eigenvalue of $\nabla_{\bb{\theta}}\bb{U}$ as in~\eqref{eqn:Utheta-diff-eigen-lower-bound} holds for any $\theta$ satisfying the condition $\vert \theta^a(x)\vert \leq K\log(\delta^{-1}-1)$. Fix any $x \in \chi$, and by mean value theorem, we obtain
    \begin{equation}
        0 
        = \bb{U}(\bbhat{\theta}_t, \bbhat{m}_t, \bb{\rho}, x)\\
        = \bb{U}(\bb{\theta}^\ast, \bbhat{m}_t, \bb{\rho}, x) + [\nabla_{\bb{\theta}}\bb{U}]\vert_{\bb{\theta} = \bb{\theta}_s} (\bbhat{\theta}_t - \bb{\theta}^\ast),
        \label{eqn:theta-hat-taylor}
    \end{equation}
    \noindent where $\bb{\theta}_s = s \bbhat{\theta}_t + (1-s)\bb{\theta}^\ast$ for some $s \in (0, 1)$. Since each coordinate of $\bbhat{\theta}_t$ and $\bb{\theta}^\ast$ are uniformly bounded by $K\log(\delta^{-1}-1)$, it follows that each coordinate of $\bb{\theta}_s$ is also uniformly bounded by the same constant. Consequently, we have 
    \begin{align*}
        & \Vert \bbhat{\theta}_t - \bb{\theta}^\ast \Vert_2\\
        \leq{} & \frac{(\delta^K + (1-\delta)^K)^2}{\delta^{K+1}(1-\delta)^K} \Vert \bb{U}(\bb{\theta}^\ast, \bbhat{m}_t, \bb{\rho}, x)\Vert_2\\
        \leq{} & \frac{(\delta^K + (1-\delta)^K)^2}{\delta^{K+1}(1-\delta)^K} \left[ \Vert \bb{U}(\bb{\theta}^\ast, \bb{m}^\ast, \bb{\rho}, x)\Vert_2 + \sup_{\bb{\theta}} \Vert  \bb{U}(\bb{\theta}, \bb{m}, \bb{\rho}, x) - \bb{U}(\bb{\theta}, \bbhat{m}_t, \bb{\rho}, x) \Vert \right]\\
        \leq{} & \frac{(\delta^K + (1-\delta)^K)^2}{\delta^{K+1}(1-\delta)^K} \left(\sum_{a \in [K]} \sum_{b \in [K]\setminus \{a\} } (\rho^{ab}(x))^2 \right)^{1/2} \sup_{(a, b)\in [K]_2}\sup_{x \in \chi} \vert m^{ab}(x) - \hat{m}_t^{ab}(x) \vert\\
        \leq{} & \frac{(\delta^K + (1-\delta)^K)^2}{\delta^{K+1}(1-\delta)^K} \sup_{(a, b)\in [K]_2} \sup_{x \in \chi} \vert m^{ab}(x) - \hat{m}_t^{ab}(x) \vert.
    \end{align*}
    \noindent Here, we use the fact that $\bb{U}(\bb{\theta}^\ast, \bb{m}^\ast, \bb{\rho}, x) = \bb{0}$, by definition. Note that, the bound is uniform over $x \in \chi$. Thus, taking supremum over $x \in \chi$, squaring it and taking an expectation recovers that the uniform $L_2$-error convergence for $\bbhat{\theta}_t - \bb{\theta}$. 
    
    \paragraph{Convergence of gradients of $\bb{U}$.} As the entries of the gradient $\nabla_{\bb{m}}\bb{U}(x)$ depend only on the exogenous weights $\rho^{ab}(x)$, the bound is obvious.

    For the gradients of score-function $\bb{U}$ with respect to preference parameters $\bb{\theta}$, consider the chain of inequalities, 
    \begin{align*}
        & \sup_{x \in \chi} \Vert [\nabla_{\bb{\theta}}\bb{U}(\bbhat{\theta}_t, x)] - [\nabla_{\bb{\theta}}\bb{U}(\bb{\theta}^\ast, x)] \Vert_F \\
        \leq{} & (K-1) \sup_{(a, b) \in [K]_2} \sup_{x \in \chi} \left\vert [\nabla_{\bb{\theta}} \bb{U}(\bbhat{\theta}_t, x)]_{ab} - [\nabla_{\bb{\theta}} \bb{U}(\bb{\theta}^\ast, x)]_{ab} \right\vert \\
        ={} & (K-1)\sup_{a \in [K]} \sum_{b \in [K]} \sup_{x \in \chi} \left\vert \rho^{ab}(x) (\sigma(\widehat{\theta}_t^a(x) - \widehat{\theta}_t^b(x))) (1 - \sigma(\widehat{\theta}_t^a(x) - \widehat{\theta}_t^b(x))) \right. \\
        & \qquad \qquad \left. - \rho^{ab}(x) (\sigma(\theta^{a,\ast}(x) - {\theta}^{a,\ast}(x))) (1 - \sigma({\theta}^{a,\ast}(x) - {\theta}^{b,\ast}(x))) \right\vert\\
        \leq{} & 3(K-1)\sup_{a \in [K]} \sum_{b\in [K]} \sup_{x \in \chi} \vert \rho^{ab}(x)(\sigma(\widehat{\theta}_t^a(x) - \widehat{\theta}_t^b(x)) - \sigma({\theta}_t^{a,\ast}(x) - {\theta}_t^{b,\ast}(x))) \vert\\
        \leq{} & \frac{3}{4}(K-1) \sup_{a \in [K]} \sup_{x \in \chi} \vert  \widehat{\theta}_t^a(x) - \theta^{a,\ast}(x) \vert,
    \end{align*}
    \noindent where the second last inequality follows from the $1/4$-Lipschitz property of the sigmoid function, and since $\sum_{a,b} \rho^{ab}(x) = 1$.
\end{proof}

\subsection{Proof of Lemma~\ref{lem:KL-div-gap}}\label{proof-lem:KL-div-gap}

Let $d_N(\Pi) = m$. The KL-divergence between the distributions $P_{\bb{w}, \Delta}$ and $P_{\bb{w}^{(i)}, \Delta}$ can be calculated as follows:
\begin{align*}
    \text{KL}(P_{\bb{w},\Delta} \Vert P_{\bb{w}^{(i)}, \Delta})
    & = \E_{P_{\bb{w},\Delta}}\left[ \log\left( \frac{f_{P_{\bb{w},\Delta}}(\Hcal_T \mid \{ X_t \}_{t=1}^T )}{f_{P_{\bb{w}^{(i)},\Delta}}(\Hcal_T \mid \{ X_t \}_{t=1}^T )} \right) \right]\\
    & = \E_{P_{\bb{w},\Delta}}\left[ \sum_{t = 1}^T \log\left( \frac{f_{P_{\bb{w},\Delta}}( Y_t, a_t, b_t \mid X_t, \Hcal_{t-1} )}{f_{P_{\bb{w}^{(i)},\Delta}}( Y_t, a_t, b_t \mid X_t, \Hcal_{t-1} )} \right) \right]\\
    & = \E_{P_{\bb{w},\Delta}}\left[ \sum_{t = 1}^T \log\left( \frac{f_{P_{\bb{w},\Delta}}( Y_t \mid  a_t, b_t, X_t, \Hcal_{t-1} ) f_{P_{\bb{w},\Delta}}( a_t, b_t \mid X_t, \Hcal_{t-1} ) }{f_{P_{\bb{w}^{(i)},\Delta}}( Y_t \mid a_t, b_t, X_t, \Hcal_{t-1} ) f_{P_{\bb{w}^{(i)},\Delta}}( a_t, b_t \mid X_t, \Hcal_{t-1} ) } \right) \right]\\
    & = \sum_{t=1}^T \E_{P_{\bb{w},\Delta}}\left[  \log\left( \frac{f_{P_{\bb{w},\Delta}}( Y_t \mid  a_t, b_t, X_t, \Hcal_{t-1} )}{f_{P_{\bb{w}^{(i)},\Delta}}( Y_t \mid a_t, b_t, X_t, \Hcal_{t-1} )} \right) \right].
\end{align*}
\noindent In the last line, we use the fact that the sampling weights $w_t(\cdot)$ are $\sigma(\{\Hcal_{t-1}, X_t\})$-measurable functions, and hence for any underlying model, $f(a_t, b_t \mid X_t, \Hcal_{t-1}) = w_t(\cdot)$. When $X_t \neq x_i$, the models behave exactly same and hence the log-likelihood has no contribution. When $X_t = x_i$, which happens with probability $1/m$, the models swap the maximizing index of the arms. For notational convenience let $u = \alpha_{\bb{w}, i}$ and $v = \alpha_{\bb{w}^{(i)}, i}$, and $m_{\bb{w},\Delta}^{ab}(x_i)$ be the preference probability under the model $P_{\bb{w}, \Delta}$. Therefore, the KL-divergence simplifies as 
\begin{equation}
    \text{KL}(P_{\bb{w},\Delta}, P_{\bb{w}^{(i)}, \Delta})
    = \frac{1}{m}\sum_{t=1}^T \E_{P_{\bb{w},\Delta}}\left[ \sum_{(a, b) \in [K]_2} w_t((a, b) \mid \Hcal_{t-1}, X_t = x_i) \delta_{ab} \right],
    \label{eqn:kl-div-as-weighted-sum}
\end{equation}
\noindent where
\begin{equation*}
    \delta_{ab} = m_{\bb{w},\Delta}^{ab}(x_i) \log\left( \frac{m_{\bb{w},\Delta}^{ab}(x_i)}{m_{\bb{w}^{(i)},\Delta}^{ab}(x_i)} \right) + (1- m_{\bb{w},\Delta}^{ab}(x_i))\log\left( \frac{1 - m_{\bb{w},\Delta}^{ab}(x_i)}{1 - m_{\bb{w}^{(i)},\Delta}^{ab}(x_i)} \right).
\end{equation*}
\noindent Now, when $X_t = x_i$, there can be four scenarios:
\begin{enumerate}
    \item If $\{a, b\} \cap \{u, v\} = \emptyset$, then the log-likelihood ratio is zero, i.e., $\delta_{ab} = 0$.
    \item If $a = u, b = v$, then 
    \begin{equation*}
        \delta_{uv} = \sigma(\Delta) \log\left(\frac{\sigma(\Delta)}{\sigma(-\Delta)}\right) + \bar{\sigma}(\Delta) \log\left(\frac{\bar{\sigma}(\Delta) }{\bar{\sigma}(-\Delta)}\right)
        = \Delta \tanh(\Delta/2).
    \end{equation*}
    \noindent Since $\tanh z \leq z$, it follows that $\delta_{uv} \leq \Delta^2/2$.
    \item If $a \neq u, b = v$, then
    \begin{equation*}
        \delta_{av} = \frac{1}{2}\log(1/2\sigma(\Delta)) + \frac{1}{2}\log(1/\bar{\sigma}(\Delta)) = \log \cosh(\Delta/2).
    \end{equation*}
    \noindent We know that $\log \cosh z \leq z^2/2$, and therefore, $\delta_{av} \leq \Delta^2/8$.
    \item If $a = u, b \neq v$, then 
    \begin{equation*}
        \delta_{ub} = \sigma(\Delta) \log(\sigma(\Delta)/2) + \bar{\sigma}(\Delta) \log(\bar{\sigma}(\Delta)/2) = \frac{\Delta}{2}\tanh(\Delta/2) - \log \cosh(\Delta/2).
    \end{equation*}
    \noindent Since $\log \cosh(\Delta/2) \geq 0$, we also have $\delta_{ab} \leq \Delta^2/4$.
\end{enumerate}

Putting everything into Eq.~\eqref{eqn:kl-div-as-weighted-sum} yields,
\begin{equation*}
    \text{KL}(P_{\bb{w},\Delta}, P_{\bb{w}^{(i)}, \Delta})
    \leq \frac{1}{m}\sum_{t=1}^T \E_{P_{\bb{w},\Delta}}\left[ \sum_{(a, b) \in [K]_2} w_t((a, b) \mid \Hcal_{t-1}, X_t = x_i) \frac{\Delta^2}{2} \right]
    = \frac{T\Delta^2}{2m},
\end{equation*}
\noindent since, for any choice of sampling weights, they must add up to $1$ when summed over every possible pairs.

\subsection{Proof of Regret Lower Bound (Theorem~\ref{thm:regret-lower-bound})}\label{appendix:proof-regret-lower}

\begin{proof}
    Under $P_{\bb{w},\Delta}$, the oracle policy achieving the zero regret is given by 
    \begin{equation*}
        \pi_{\bb{w}}(x_i) = \alpha_{\bb{w}, i},
    \end{equation*}
    \noindent which, by definition of Natarajan dimension, belongs to the policy class $\Pi$. Now, consider any policy $\hat{\pi}_{\Acal, T} \in \Pi$ produced by the algorithm $\Acal$, after $T$ rounds. Since, $X$ is uniformly distributed, the expected simple regret of this policy under the model $P_{\bb{w}, \Delta}$ is given by
    \begin{align*}
        \E(R(\hat{\pi}_{\Acal, T}; P_{\bb{w}, \Delta}))
        & = \E_{P_{\bb{w}, \Delta}}\left[ \frac{1}{m}\sum_{j=1}^m \sum_{a \in [K]} (\max_{b \in [K]} \theta^{b,\ast}(x_j) - \theta^a(x_j)) \ind{\{ \hat{\pi}_{\Acal, T}(x_j) = a\} }  \right]\\
        & = \frac{\Delta}{m}\sum_{j=1}^m \prob_{P_{\bb{w}, \Delta}}(\hat{\pi}_{\Acal, T}(x_j) \neq \alpha_{\bb{w}, j})
        = \frac{\Delta}{m}\sum_{j=1}^m e_{\bb{w}, j},
    \end{align*} 
    \noindent where $e_{\bb{w}, j} = \prob_{P_{\bb{w}, \Delta}}(\hat{\pi}_{\Acal, T}(x_j) \neq \alpha_{\bb{w}, j})$ for $j \in [m]$. Let $\bb{w}^{(i)}$ be the binary vector obtained from $\bb{w}$ by flipping the $i$-th bit. Then, the models $P_{\bb{w}^{(i)}, \Delta}$ and $P_{\bb{w}, \Delta}$ behave similarly when the context is $x \neq x_i$, except when $x = x_i$, the best performing arm switches from $u_i$ to $v_i$. 
    
    Consider the chain of inequalities,
    \begin{align*}
        1 - e_{\bb{w}, i} - e_{\bb{w}^{(i)}, i}
        & = 1 - \prob_{P_{\bb{w}, \Delta}}( \hat{\pi}_{\Acal, T}(x_i) \neq \alpha_{\bb{w}, i}) - \prob_{P_{\bb{w}^{(i)}, \Delta}}( \hat{\pi}_{\Acal, T}(x_i) \neq \alpha_{\bb{w}^{(i)}, i})\\
        & = \prob_{P_{\bb{w}, \Delta}}( \hat{\pi}_{\Acal, T}(x_i) = \alpha_{\bb{w}, i}) - \prob_{P_{\bb{w}^{(i)}, \Delta}}( \hat{\pi}_{\Acal, T}(x_i) \neq \alpha_{\bb{w}^{(i)}, i})\\
        & \leq \prob_{P_{\bb{w}, \Delta}}( \hat{\pi}_{\Acal, T}(x_i) \neq \alpha_{\bb{w}^{(i)}, i}) - \prob_{P_{\bb{w}^{(i)}, \Delta}}( \hat{\pi}_{\Acal, T}(x_i) \neq \alpha_{\bb{w}^{(i)}, i})\\
        & \leq \mathrm{TV}(P_{\bb{w},\Delta}, P_{\bb{w}^{(i)}, \Delta})\\
        & \leq \frac{1}{\sqrt{2}}\sqrt{d_{\mathrm{KL}}(P_{\bb{w},\Delta} \Vert P_{\bb{w}^{(i)}, \Delta})}
        \leq \frac{\sqrt{T}}{2\sqrt{m}}\Delta 
    \end{align*}
    \noindent where, in the second last line, we use Pinsker's inequality, and the last line follows from Lemma~\ref{lem:KL-div-gap}. Choose $\Delta = \sqrt{m/T}$, we obtain that the KL divergence is less than or equal to $1/2$, and hence
    \begin{equation*}
        e_{\bb{w},i} + e_{\bb{w}^{(i)}, i} \geq 1 - 1/2 = 1/2.
    \end{equation*}
   \noindent Summing over all $x_i$'s and averaging all binary vectors $\bb{w} \in \{0, 1\}^m$ now produces the inequality that 
   \begin{equation*}
        \frac{1}{2^m}\sum_{\bb{w} \in \{0, 1\}^m} \sum_{i = 1}^m e_{\bb{w},i} \geq \frac{m}{4}.
   \end{equation*}
   \noindent Therefore, there exists a choice of binary vector $\bb{w}^\ast$ such that $\sum_{i = 1}^m e_{\bb{w}^\ast,i} \geq m/4$. In other words, the expected simple regret of $\hat{\pi}_{\Acal, T}$ satisfies a lower bound
   \begin{equation}
        \E(R(\hat{\pi}_{\Acal, T}; P_{\bb{w}^\ast,\Delta}))
        \geq \frac{\sqrt{m/T}}{m} \times \frac{m}{4} =  \frac{1}{4}\sqrt{\frac{d_N(\Pi)}{T}}.
        \label{eqn:regret-lower-bound-num}
   \end{equation}

   On the other hand, for any model $P_{\bb{w},\Delta}$, we have for any $x_i$, 
   \begin{align}
        \sum_{a \in [K]} \left( {\theta}^a(x_i) - \bar{\theta}(x_i)\right)^2 
        & = \sum_{a \in [K]\setminus \alpha_{\bb{w}, i}} \left(0 - \frac{\Delta}{K} \right)^2 + \left(\Delta - \frac{\Delta}{K} \right)^2 \nonumber\\
        & = \dfrac{(K-1)\Delta^2}{K^2} + \dfrac{\Delta^2 (K-1)^2}{K^2}
        = \dfrac{K-1}{K}\Delta^2  =: \xi_1 \Delta^2 \label{eqn:regret-lb-thetabias}\\
        \text{ and, }
        \textstyle\sum_{(a, b) \in [K]_2} v^{ab}(x_i)
        & = \frac{(K-1)(K-2)}{2} \sigma(0)\bar{\sigma}(0) + (K-1) \sigma(\Delta)\bar{\sigma}(\Delta)\nonumber\\
        & =: \xi_2 + \xi_3 \sigma(\Delta)\bar{\sigma}(\Delta)\label{eqn:regret-lb-sumv}
    \end{align}
    \noindent Here, $\xi_1, \xi_2$ and $\xi_3$ are appropriate constants that depend only on $K$. Let $M = K\log(1/\delta - 1)$ be the upper bound on $\vert \widehat{\theta}^a_t(x)\vert$ as assured by Lemma~\ref{lem:gamma-est-bound}. Then, 
    \begin{equation*}
        \mathrm{trace}(\Sigma_{\bb{A}}) = \Vert \bb{A}(x)\Vert_F^2
        \leq \frac{(1 + e^M)^4}{\delta^2 e^{2M}} \Vert \nabla_{\bb{m}}\bb{U}\Vert_F^2 \leq \frac{K(1 + e^M)^4}{\delta^2 e^{2M}}.
    \end{equation*}
    \noindent Combining Eq.~\eqref{eqn:regret-lb-thetabias} and~\eqref{eqn:regret-lb-sumv} with this, we get an upper bound on the denominator that
    \begin{multline*}
        \E_{P} \left[ \sum_{a \in [K]} (\theta^a(X) - \bar{\theta}(X))^2 + \trace{\bb{\Sigma}_A(X)} \sum_{(a, b) \in [K]_2} v^{ab}(X) \right]\\
        \leq \xi_1 \Delta^2 + \frac{K(1 + e^M)^4}{\delta^2 e^{2M}}\left( \xi_2 + \xi_3 \sigma(\Delta)\bar{\sigma}(\Delta) \right).
    \end{multline*}
    \noindent Since we choose $\Delta = \sqrt{m/T}$, for all sufficiently large $T$, the square of the denominator is bounded above by
    \begin{equation*}
        \frac{K(1+e^M)^4}{\delta^2 e^{2M}} \left[ \frac{(K-1)(K-2)}{4} + \frac{K-1}{2} \right] = \frac{K^2(K-1)(1+e^M)^4}{4\delta^2 e^{2M}}.
    \end{equation*}
    \noindent Therefore, for the specific choice $\bb{w}^\ast$, 
    \begin{equation*}
        \dfrac{\E(R( \hat{\pi}_{\Acal, T}, P_{\bb{w}^\ast,\Delta} )) }{\sqrt{\E_{P} \left[ \sum_{a \in [K]} (\theta^a(X) - \bar{\theta}(X))^2 + \trace{\bb{\Sigma}_A(X)} \sum_{(a, b) \in [K]_2} v^{ab}(X) \right]}}
        \geq 
        c \frac{\sqrt{d_N(\Pi)}}{\sqrt{T}}.
    \end{equation*}
    \noindent Here, 
    \begin{equation*}
        c = \frac{\delta e^{M}}{2K\sqrt{K-1}(1+e^M)^2}.
    \end{equation*}
    \noindent Since taking supremum over the choices of model can only increase the regret, and for every algorithm $\Acal$, we can produce such $w^\ast$, the result follows.
\end{proof}

\subsection{Proof of Lemma~\ref{lem:Q-function-diff-step1}}\label{appendix:proof-lem-Q-function-diff-step1}

\begin{proof}
    Before proceeding with the proof, similar to Eq.~\eqref{eqn:A-defn}, let us define the notations
    \begin{align}
        \bb{A}^\ast(x) & = [\nabla_{\bb{\theta}} \bb{U}(\bb{\theta}^\ast, x) ]^{-1} [\nabla_{\bb{m}} \bb{U}(x)], \label{eqn:A-star-defn}\\
        \bbhat{A}_t(x) & = [\nabla_{\bb{\theta}} \bb{U}(\bbhat{\theta}_t, x) ]^{-1} [\nabla_{\bb{m}} \bb{U}(x)], \label{eqn:A-hat-defn}
    \end{align}
    \noindent so that $\bb{\Sigma}_A(x) = \bb{A}^\ast(x) (\bb{A}^\ast(x))\tr$ and $\bbhat{\Sigma}_{A,t}(x) = \bbhat{A}_t(x) (\bbhat{A}_t(x))\tr$.

    Note that, under the construction of $Q_T^\ast(\pi)$, $\{ X_t, \Gamma_t^{a,\ast}\}$ are i.i.d. observations. For any policy $\pi \in \Pi$, since $\E(\bb{1}_t^{ab} Y_t \mid X_t = x) = m^{ab}(x)$ for any $x \in \chi$, the expectation of each summand of the proxy value function leads to the true value of the policy. Formally, we have 
    \begin{align*}
        & \E\left[ \sum_{a \in [K]} \bb{1}_{\{ \pi(X_t) = a\}} \Gamma_t^{a,\ast} \mid X_t = x \right]\\
        ={} & \sum_{a \in [K]} \bb{1}_{\{ \pi(x) = a \}} \E(\Gamma_t^{a,\ast} \mid X_t = x) \\
        ={} & \sum_{a \in [K]} \bb{1}_{\{ \pi(x) = a \}} \left( \theta^{a,\ast}(x)  - \left[\bb{A}^\ast(x) \text{vec}\left( \left\{ \frac{1}{w^{ab}(x)} \E\left[ \bb{1}_t^{ab} (Y_t - m^{ab}(x))  \mid X_t = x  \right] \right\} \right)\right]_{a} \right)\\
        ={} & \sum_{a \in [K]} \bb{1}_{\{ \pi(x) = a \}} \bb{\theta}^{a,\ast}(x).
    \end{align*}
    \noindent Let us define $\delta_t^\ast(\pi_1, \pi_2)$ as given in Eq.~\eqref{eqn:small-delta-t-star-defn}. As a result, $\delta_t^\ast(\pi_1, \pi_2)$ for $t = 1, 2, \dots, T$ are i.i.d. random variables each with expectation $\Delta(\pi_1, \pi_2)$, and $\Delta_T^\ast(\pi_1, \pi_2)$ is the empirical average of them. Additionally, due to Lemma~\ref{lem:gamma-est-bound}, each $\delta_t^\ast(\pi_1, \pi_2)$ is uniformly bounded, i.e.,
    \begin{multline*}
        \vert \delta_t^\ast(\pi_1, \pi_2)\vert
        = \left\vert \sum_{a \in [K]} (\pi_1(a \mid X) - \pi_2(a \mid X)) \Gamma_t^{a,\ast} \right\vert
        \leq 2 \sup_{a \in [K]} \vert \Gamma_t^{a,\ast}\vert\\
        \leq 2 \left[ K^{3/2} \log\left( \delta^{-1} - 1 \right) + \frac{K(K-1)(\delta^K + (1-\delta)^K)^2}{4\sqrt{2}\delta^{K+2}(1-\delta)^{K}}\right] = C_1.
    \end{multline*}
    Therefore, by an application of Theorem 4.10 of~\cite{wainwright2019high}, we obtain that for any $T \geq 1$ and $\epsilon > 0$, 
    \begin{align*}
        \sup_{\pi_1, \pi_2 \in \Pi} \vert \Delta(\pi_1, \pi_2) - {\Delta}^\ast_T(\pi_1, \pi_2) \vert 
        = \sup_{\pi_1, \pi_2 \in \Pi} \left\vert \dfrac{1}{T}\sum_{t=1}^T \delta_t^\ast(\pi_1, \pi_2) - \E(\delta_t^\ast(\pi_1, \pi_2)) \right\vert 
        \leq 2\Rcal_T(\Pi^D) + \epsilon,
    \end{align*}
    \noindent with probability $1 - \exp(-T\epsilon^2/2C_1^2)$. Let us call the left-hand side as $D_1$. Since the above is valid for any $T \geq 1$, we can obtain the bound on the expectation as 
    \begin{align*}
        \E(D_1)
        & \leq 2\Rcal_T(\Pi^D) + \E\left( (D_1 - 2\Rcal_T(\Pi^D))_+ \right)\\
        & = 2\Rcal_T(\Pi^D) + \int_0^\infty \prob((D_1 - 2\Rcal_T(\Pi^D))_+ > \epsilon ) d\epsilon\\
        & \leq 2\Rcal_T(\Pi^D) + \int_0^\infty \exp(-T\epsilon^2/2C_1^2) d\epsilon = 2\Rcal_T(\Pi^D) + \sqrt{\frac{\pi}{2T}}C_1
    \end{align*}
    \noindent Here, $X_+ = \max\{ X, 0\}$.
    
    However, consider the inequality (40) (in the proof of Theorem 1) of~\cite{zhou2023offline} which bounds the conditional Rademacher complexity of the policy class. It ensures that
    \begin{equation}
        \Rcal_T(\Pi^D) \leq \dfrac{27.2\sqrt{2}(\kappa(\Pi) + 8)}{T}\sqrt{ \E\left[ \sup_{\pi_1, \pi_2 \in \Pi} \sum_{t = 1}^T \left( \delta_t^\ast(\pi_1, \pi_2) \right)^2 \right] } + o(1/\sqrt{T})
        \label{eqn:radmacher-bound}
    \end{equation}
    \noindent where the expectation on the right-hand-side is taken over the i.i.d. observations $\{ X_t, \Gamma_t^{a,\ast}\}$. Here, 
    \begin{align}
        B_1
        & = \E\left[ \sup_{\pi_1, \pi_2 \in \Pi} \sum_{t=1}^T (\delta_t^\ast(\pi_1, \pi_2))^2 \right] \nonumber \\
        & = \E\left[ \sup_{\pi_1, \pi_2 \in \Pi} \sum_{t = 1}^T \left( \sum_{a \in [K]} ( \bb{1}_{\{ \pi_1(X_t) = a \}} - \bb{1}_{\{ \pi_2(X_t) = a \}})\Gamma^{a,\ast}_t \right)^2 \right] \nonumber \\
        & = \E\left[ \sup_{\pi_1, \pi_2 \in \Pi} \sum_{t = 1}^T \left( \sum_{a \in [K]} (\bb{1}_{\{ \pi_1(X_t) = a \}} - \bb{1}_{\{ \pi_2(X_t) = a \}} )(\Gamma^{a,\ast}_t - c) \right)^2 \right], \text{ for any constant } c > 0, \nonumber \\
        & \leq \E\left[ \sup_{\pi_1, \pi_2 \in \Pi} \sum_{t = 1}^T \left( \sum_{a \in [K]}( \bb{1}_{\{ \pi_1(X_t) = a \}} -  \bb{1}_{\{ \pi_2(X_t) = a \}} )^2 \right) \left( \sum_{a \in [K]} (\Gamma_t^{a,\ast} - c)^2 \right) \right], \label{eqn:B1-bound-1}\\
        & \leq 2\E\left[ \sum_{t=1}^T \sum_{a \in [K]} (\Gamma_t^{a,\ast} - c)^2  \right]. \label{eqn:B1-bound-2}
    \end{align}
    \noindent Inequality~\eqref{eqn:B1-bound-1} follows from an application of the Cauchy-Schwartz inequality and inequality~\eqref{eqn:B1-bound-2} is a consequence of the fact that both $\pi_1(a \mid X_t)$ and $\pi_2(a \mid X_t)$ are probability distributions. Continuing, 
    \begin{align*}
        B_1 
        & \leq 2 \sum_{t=1}^T \E\left[ (\bb{\Gamma}_t^\ast - c \bb{1}) \tr (\bb{\Gamma}_t^\ast - c\bb{1}) \right]\\
        & = 2\sum_{t=1}^T \E\left[ \Vert \bb{\theta}^\ast(X_t) - c \bb{1} \Vert^2 + \trace{\bb{A}^\ast(X) \var\left( \dfrac{\bb{1}_t^{ab}(Y_t - m^{ab}(X_t))}{w^{ab}(X_t)} \right) \bb{A}^\ast(X)\tr } \right]
    \end{align*}
    \noindent Now note that, the variance can be calculated as
    \begin{align*}
        \E\left( \dfrac{\bb{1}_t^{ab}(Y_t - m^{ab}(X_t))^2}{(w^{ab}(X_t))^2} \right)
        ={} & \E\left[ \dfrac{\bb{1}_t^{ab}}{(w^{ab}(X_t))^2} \E\left[ (Y_t - m^{ab}(X_t))^2 \mid \bb{1}_t^{ab} \right] \right]\\
        ={}& \dfrac{m^{ab}(X_t) (1 - m^{ab}(X_t))}{w^{ab}(X_t)}
        ={}  \sum_{(a, b) \in [K]_2} v^{ab}(X_t).
    \end{align*}
    \noindent For any $(a, b) \in [K]_2$ with $\{ a, b\} \neq \{ a', b'\}$, we have the covariance term
    \begin{equation*}
        \E\left[ \dfrac{\bb{1}_t^{ab} \bb{1}_t^{a'b'}}{w^{ab}(X_t) w^{a'b'}(X_t)} (Y_t - m^{ab}(X_t))(Y_t - m^{a'b'}(X_t)) \right]
        = 0.
    \end{equation*}
    \noindent as at most one of $\bb{1}_t^{ab}$ and $\bb{1}_t^{a'b'}$ could be equal to $1$ when $\{ a, b\} \neq \{ a', b'\}$. Therefore, 
    \begin{equation*}
        B_1 
        \leq 2T \E\left[ \sum_{a \in [K]}  (\theta^{a,\ast}(X) - c)^2 + \trace{\bb{A}^\ast(X)\bb{A}^\ast(X)\tr} \sum_{(a, b) \in [K]_2} v^{ab}(X) \right].
    \end{equation*}
    \noindent We choose $c = \bar{\theta}^\ast(X)$. Putting it back into~\eqref{eqn:radmacher-bound}, and then applying to the bound on $\E(D_1)$ completes the proof.
\end{proof}

\subsection{Empirical Process Bound for Martingale Differences}

\begin{lemma}\label{lem:ulln-martingale}
    Let for each $a \in [K]$, $\{ Z_t^a \}_{t \geq 1}$ be a martingale difference sequence with respect to the filtration $\Fcal_t$ such that $\E(Z_t^a \mid \Fcal_{t-1}) = 0$ and $\vert Z_t^a\vert \leq M$ for some constant $M > 0$. Moreover, let $X_1, \dots, X_T$ be the i.i.d. context such that $X_t$ is $\Fcal_{t-1}$-measurable. Then,
    \begin{equation*}
        \E\left[ \sup_{\pi_1, \pi_2 \in \Pi} \left\vert \frac{1}{T}\sum_{t=1}^T \sum_{a \in [K]} (\bb{1}_{\{ \pi_1(X_t) = a \}} - \ind{\{ \pi_2(X_t) = a \}} ) Z_t^a \right\vert  \right] \leq \frac{96\sqrt{2}M}{\sqrt{T}} \kappa(\Pi),
    \end{equation*}
    \noindent where $\kappa(\Pi)$ is the entropy integral as in~\eqref{eqn:entropy-integral}.
\end{lemma}

\begin{proof}
    For any policy $\pi \in \Pi$, define
    \begin{equation*}
        M_T(\pi) = \frac{1}{T}\sum_{t=1}^T Z_t^{\pi(X_t)}.
    \end{equation*}
    \noindent Let us define a family of functions indexed by $\pi \in \Pi$ as 
    \begin{equation*}
        \pi \mapsto \frac{1}{M} Z_t^{\pi(X_t)} = \frac{1}{M}\sum_{a \in [K]} \ind{\{ \pi(X_t) = a\}} Z_t^a , 
    \end{equation*}
    \noindent for each $t \geq 1$. Clearly, $\E(\bb{1}_{\{ \pi(X_t) = a\} } Z_t^a \mid \Fcal_{t-1}) = 0$. Each such function is $[-1, 1]$-valued. By Theorem 2 of~\cite{Rakhlin_Sridharan_Tewari_2015}, we obtain that
    \begin{equation}
        \E\left[ \sup_{\pi_1, \pi_2 \in \Pi} \left\vert M_T(\pi_1) - M_T(\pi_2) \right\vert \right]
        \leq 4M \Rcal_{\mathrm{seq}, T}(\Pi^D),
        \label{eqn:lem:ulln-martingale-1}
    \end{equation}
    \noindent where $\Rcal_{\mathrm{seq}, T}(\Pi^D)$ is the sequential Rademacher complexity of the policy difference class $\Pi^D$ as in Definition~\ref{appendix:policy-diff-Rademacher}, but with $Z_t^a$ substituting for $\Gamma_t^a$. Theorem 3 of~\cite{Rakhlin_Sridharan_Tewari_2015} provides that
    \begin{equation}
        \Rcal_{\mathrm{seq}, T}(\Pi^D)
        \leq \sup_{\bb{z}} \inf_{\alpha > 0}\left[ 4\alpha + \frac{12}{\sqrt{T}}\int_\alpha^1 \sqrt{\log N_2(\delta, \Pi^D; \bb{z}, \bb{x})} d\delta \right].
        \label{eqn:lem:ulln-martingale-2}
    \end{equation}
    \noindent Here, $N_2(\delta, \Pi^D; \bb{z}, \bb{x})$ is the $L_2$-norm covering number of $\Pi^D$ on a tree $\bb{z}$ of depth $T$ with covariates $\bb{x}$; see Section~\ref{appendix:z-tree} for related definitions.

    Now, we apply a variation of Lemma 4 of~\cite{zhan2024policy} for the policy difference class $\Pi^D$. Fix any Radamacher tree $\bb{z}$ and corresponding context sequence $\bb{x}$. Consider any $\pi_1, \pi_2 \in \Pi$, and any family of functions $f_\pi$ indexed by $\pi \in \Pi$ and bounded between $-1$ and $1$. Then, for any path through $\bb{z}$, the two policies contribute the same score whenever they agree, while if they disagree, the difference is at most $2$. As a result, the pathwise empirical $L_2$-norm $\Vert f_\pi - f_{\pi'}\Vert_{2,\mathrm{path}}$ is bounded above by $2\sqrt{d_H(\pi, \pi'; \bb{x})}$, the hamming distance corresponding to the fixed context. Now, let $\mathcal{C}$ be a Hamming cover of $\Pi$ at radius $u^2/4$. For each $\pi \in \Pi$, let $\bar{\pi} \in \mathcal{C}$ be its representative in the cover. Then, by triangle inequality
    \begin{equation*}
        \left\Vert \frac{f_{\pi_1} - f_{\pi_2}}{2} - \frac{f_{\bar{\pi}_1} - f_{\bar{\pi}_2}}{2} \right\Vert_{2,\mathrm{path}}
        \leq \frac{u}{2} + \frac{u}{2} = u.
    \end{equation*}
    \noindent For the policy class $\Pi^D$, we have at most $\vert \mathcal{C}\vert^2$ ordered pair of representatives. Hence, a similar line of reasoning as in Lemma 4 of~\cite{zhan2024policy} yields
    \begin{equation*}
        N_2(\delta, \Pi^D; \bb{z}, \bb{x}) \leq \left( N_H(\delta^2/4, \Pi) \right)^2.
    \end{equation*}
    \noindent Consequently, taking $\alpha \to 0$ in inequality~\eqref{eqn:lem:ulln-martingale-2} recovers
    \begin{equation*}
        \Rcal_{\mathrm{seq}, T}(\Pi) \leq \frac{12\sqrt{2}}{\sqrt{T}}\int_0^1 \sqrt{\log N_H(\delta^2/4, \Pi)}d\delta \leq \frac{24\sqrt{2}}{\sqrt{T}}\kappa(\Pi).
    \end{equation*}
    \noindent Substituting this upper bound into inequality~\eqref{eqn:lem:ulln-martingale-1}, we obtain that
    \begin{equation*}
        \E\left[ \sup_{\pi_1, \pi_2 \in \Pi} \left\vert \frac{1}{T}\sum_{t=1}^T \sum_{a \in [K]} (\bb{1}_{\{ \pi_1(X_t) = a \}} - \bb{1}_{\{ \pi_2(X_t) = a \}})Z_t^a \right\vert \right] \leq \frac{96\sqrt{2}M}{\sqrt{T}}\kappa(\Pi).
    \end{equation*}
\end{proof}

\subsection{Proof of Lemma~\ref{lem:value-function-diff}}\label{appendix:proof-value-function-diff}

\begin{proof}
Let us consider the notation $\bb{Z}_{1t}$ and $\bb{Z}_{2t}$ as given in Eq.~\eqref{eqn:Z1t-defn} and~\eqref{eqn:Z2t-defn}, and their $a$-th element as $Z_{1t}^a$ and $Z_{2t}^a$. We have the chain of inequalities,
\begin{align}
    \left\vert D_T(\pi_1) - D_T(\pi_2) \right\vert 
    ={} & \left\vert \dfrac{1}{T}\sum_{t=1}^T \sum_{a \in [K]} (\bb{1}_{\{ \pi_1(X_t) = a \}} - \ind{\{ \pi_2(X_t) = a \}} ) (\widehat{\Gamma}_t^a - \Gamma_t^{a,\ast})\right\vert \nonumber\\
    \leq{} & \left\vert \dfrac{1}{T}\sum_{t=1}^T \sum_{a \in [K]} (\bb{1}_{\{ \pi_1(X_t) = a \}} - \ind{\{ \pi_2(X_t) = a \}} ) ({Z}_{1t}^a - {Z}_{2t}^a) \right\vert \nonumber \\
    & \qquad + 2 \sup_{\pi \in \Pi} \left\vert \dfrac{1}{T}\sum_{t=1}^T \sum_{a \in [K]} \bb{1}_{\{ \pi(X_t) = a \}} ( \E(\widehat{\Gamma}_t^a \mid \Fcal_t) - \E(\Gamma_t^{a,\ast} \mid \Fcal_t) )  \right\vert \label{eqn:value-function-diff-triangle}
\end{align}
\noindent Note that, for $j = 1, 2$, as seen from Lemma~\ref{lem:gamma-est-bound},
\begin{equation*}
    \vert Z_{jt}^a\vert 
    \leq 2 \sup_{a \in [K]}\vert \hat{\Gamma}_t^a\vert 
    \leq 2 \left[ K^{3/2} \log(\delta^{-1}-1) + \frac{K(K-1)(\delta^K + (1-\delta)^K)^2}{4\sqrt{2}\delta^{K+2} (1-\delta)^{K}} \right] = C_1.
\end{equation*}
\noindent Applying Lemma~\ref{lem:ulln-martingale} to the expectation of the maximum of the first summand, where the maximum is taken over $\pi_1, \pi_2 \in \Pi$, we obtain
\begin{equation*}
    \E\left[ \sup_{\pi_1, \pi_2 \in \Pi} \left\vert \dfrac{1}{T}\sum_{t=1}^T (\sum_{a \in [K]} (\bb{1}_{\{ \pi_1(X_t) = a \}} - \bb{1}_{\{ \pi_2(X_t) = a \}}) ({Z}_{1t}^a - {Z}_{2t}^a)) \right\vert \right]
    \leq \frac{192\sqrt{2}}{\sqrt{T}}\kappa(\Pi) C_1.
\end{equation*} 

\noindent To deal with the second summand in~\eqref{eqn:value-function-diff-triangle}, starting from the decomposition~\eqref{eqn:theta-hat-taylor} and the linearity of $\bb{m}$ in the $\bb{U}$ function, we obtain
\begin{align*}
    \bbhat{\theta}_t - \bb{\theta}^\ast
    & = -[\nabla_{\bb{\theta}}\bb{U}\vert_{\bb{\theta} = -\bb{\theta}^\ast} ]^{-1} (\bb{U}(\bbhat{\theta}^\ast, \bbhat{m}_t, \bb{\rho}, x)) + \bb{R}_t\\
    & = -[\nabla_{\bb{\theta}}\bb{U}\vert_{\bb{\theta} = -\bb{\theta}^\ast} ]^{-1} [\nabla_{\bb{m}}\bb{U}] (\bbhat{m}_t - \bb{m}) + \bb{R}_t
    = -\bb{A}_t (\bbhat{m}_t - \bb{m}) + \bb{R}_t.
\end{align*}
\noindent Here, $\bb{R}_t$ is a second-order remainder term satisfying a boundedness condition
\begin{equation*}
    \Vert \bb{R}_t\Vert_2 \leq \Vert [\nabla^2_{\bb{\theta}}\bb{U}] \Vert \Vert \bbhat{\theta}_t - \bb{\theta}^\ast\Vert_2^2.
\end{equation*}
\noindent Since the sigmoid function is Lipschitz, and so is $\sigma(x)(1-\sigma(x))$ and $\Vert \bb{\theta}\Vert$ is bounded due to Lemma~\ref{lem:gamma-est-bound}, the remainder term is bounded by $C \sup_{(a, b)\in [K]_2} \sup_{x \in \chi} \vert m^{ab}(x) - \widehat{m}_t^{ab}(x) \vert^2$, where $C$ is a constant depending on $\delta$ and $K$; see Lemma~\ref{lem:consistency-bounds}. Consequently, we now have the chain of inequalities 
\begin{align*}
    \Vert \E(\bbhat{\Gamma}_t \mid \Fcal_t) - \E(\bb{\Gamma}_t^{\ast} \mid \Fcal_t)\Vert 
    & = \Vert \bbhat{\theta}_t(X_t) - \bb{\theta}^\ast_t(X_t) - \bbhat{A}_t (\bb{m}(X_t) - \bbhat{m}_t(X_t)) \Vert \nonumber \\
    & \leq \Vert (\bb{A}_t - \bbhat{A}_t) (\bbhat{m}_t - \bb{m})\Vert + \Vert R_t\Vert\\
    & \leq C' \sup_{(a, b)\in [K]_2}\sup_{x \in \chi} \vert m^{ab}(x) - \widehat{m}_t^{ab}(x) \vert^2,
\end{align*}
\noindent where $C'$ is another constant depending on $\delta$ and $K$. Here, we also use the fact that $\Vert \bbhat{A}_t - \bb{A}_t\Vert^2 \leq C \Vert \bbhat{m}_t - \bb{m}\Vert$ as a consequence of Lemma~\ref{lem:consistency-bounds}. Therefore,
\begin{align*}
    & \E\left[ \left\vert \dfrac{1}{T}\sum_{t=1}^T \sum_{a \in [K]} \bb{1}_{\{ \pi(X_t) = a \}} ( \E(\hat{\Gamma}_t^a \mid \Fcal_t) - \E(\Gamma_t^{a,\ast} \mid \Fcal_t) )  \right\vert \right]\\
    \leq{} & \frac{1}{T}\sum_{t=1}^T \E\left[ \left\vert \sum_{a \in [K]} \bb{1}_{\{ \pi(X_t) = a \}} ( \E(\hat{\Gamma}_t^a \mid \Fcal_t) - \E(\Gamma_t^{a,\ast} \mid \Fcal_t) ) \right\vert  \right]\\
    \leq{} & \frac{1}{T}\sum_{t=1}^T \E\left[ \Vert \E(\bbhat{\Gamma}_t \mid \Fcal_t) - \E(\bb{\Gamma}_t^\ast \mid \Fcal_t) \Vert \right]\\
    \leq{} & \frac{C'}{T}\sum_{t=1}^T \E\left[  \sup_{(a, b)\in [K]_2}\sup_{x \in \chi} \vert m^{ab}(x) - \hat{m}_t^{ab}(x) \vert^2 \right]
    = T^{-1}\sum_{t=1}^T o(t^{-1/2}) = o(T^{-1/2}).
\end{align*}
\noindent The last line uses Assumption~\ref{assum:estimate-consistency} and that the term $\sum_{t=1}^T t^{-1/2}$ is a lower Riemann sum for the integral $\int_1^T t^{-1/2} dt = 2\sqrt{T}$.

Putting it back into~\eqref{eqn:value-function-diff-triangle}, we obtain
\begin{equation}
    \E\left[ \sqrt{T}\sup_{\pi_1, \pi_2 \in \Pi}\vert D_T(\pi_1) - D_T(\pi_2)\vert\right]
    \leq 192\sqrt{2}C_1 \kappa(\Pi) + o(1).
    \label{eqn:proof-value-function-diff}
\end{equation}
\end{proof}

\subsection{Proof of Regret Upper Bound (Theorem~\ref{thm:regret-upper-bound})}\label{appendix:proof-regret-ub}

Starting with the decomposition given in Eq.~\eqref{eqn:risk-bound-decomp}, the risk is bounded by
\begin{align}
    R_T(P; \widehat{\pi}_T)
    \leq{} & \sup_{\pi_1, \pi_2 \in \Pi} \left\vert \Delta(\pi_1, \pi_2) - {\Delta}^\ast_T(\pi_1, \pi_2) \right\vert + \sup_{\pi_1, \pi_2 \in \Pi} \left\vert \Delta_T^\ast(\pi_1, \pi_2) - \widehat{\Delta}_T(\pi_1, \pi_2) \right\vert \nonumber \\
    \leq{} &  \sup_{\pi_1, \pi_2 \in \Pi} \left\vert \Delta(\pi_1, \pi_2) - {\Delta}^\ast_T(\pi_1, \pi_2) \right\vert + \sup_{\pi_1, \pi_2 \in \Pi} \left\vert D_T(\pi_1) - D_T(\pi_2) \right\vert
    \label{eqn:risk-upper-bound-decomp}
\end{align}
For the first term, applying Lemma~\ref{lem:Q-function-diff-step1} yields
\begin{equation*}
    \E\left[ \sqrt{T}\sup_{\pi_1, \pi_2 \in \Pi} \vert \Delta(\pi_1, \pi_2) - \Delta_T^\ast(\pi_1, \pi_2)\vert \right]
    \leq 108.8(\kappa(\Pi) + 8)\sqrt{V^\ast} + \sqrt{\frac{\pi}{2}}C_1 + o(1).
\end{equation*}
For the second term, we apply Lemma~\ref{lem:value-function-diff}. This yields the upper bound of the expected risk,
\begin{equation*}
    \E(\sqrt{T}R_T(P; \hat{\pi}_T))
    \leq 108.8(\kappa(\Pi) + 8)\sqrt{V^\ast} + C_1\left(192\sqrt{2}\kappa(\Pi) +  \sqrt{\frac{\pi}{2}}\right) + o(1).
\end{equation*}

\end{document}